\documentclass[reprint,nofootinbib,superscriptaddress,amsmath,amssymb,aps,]{revtex4-2}

\usepackage{graphicx} 
\usepackage{dcolumn}
\usepackage{bm}
\usepackage{lipsum}
\usepackage{wrapfig,lipsum,booktabs}
\usepackage[utf8]{inputenc} 
\usepackage[T1]{fontenc}    
\usepackage{hyperref}       
\usepackage{url}            
\usepackage{booktabs}       
\usepackage{amsmath}
\usepackage{amsfonts} 
\usepackage{amssymb}
\usepackage{nicefrac}       
\usepackage{microtype}      
\usepackage{xcolor}         
\usepackage{bm}
\usepackage{wrapfig}
\usepackage{comment}
\usepackage{algorithm}
\usepackage[noend]{algpseudocode}
\usepackage{mathtools}
\usepackage{amsthm}
\usepackage{autonum}
\newcommand{\app}[2]{#1~\ref{#2}}
\labelformat{subsection}{\thesection#1}
\newtheorem{theorem}{Theorem}[section]
\newtheorem*{theorem*}{Theorem}
\newtheorem*{proposition*}{Proposition}
\newtheorem*{corollary*}{Corollary}

\DeclareMathOperator*{\argmin}{arg\,min}
\newcommand{\rev}[1]{\textcolor{black}{#1}}
\begin{document}

\title{Inferring biological processes with intrinsic noise from cross-sectional data}
\author{Suryanarayana Maddu}
\altaffiliation{These authors contributed equally to this work}
\affiliation{Center for Computational Biology, Flatiron Institute, New York, NY, USA, 10010}
\author{Victor Chard\`es}
\altaffiliation{These authors contributed equally to this work}
\affiliation{Center for Computational Biology, Flatiron Institute, New York, NY, USA, 10010}
\author{Michael. J. Shelley}
\affiliation{Center for Computational Biology, Flatiron Institute, New York, NY, USA, 10010}
\affiliation{Courant Institute of Mathematical Sciences, New York University, New York, NY, USA, 10012}

\begin{abstract}
Inferring dynamical models from data continues to be a significant challenge in computational biology, especially given the stochastic nature of many biological processes. We explore a common scenario in omics, where statistically independent cross-sectional samples are available at a few time points, and the goal is to infer the underlying diffusion process that generated the data. Existing inference approaches often simplify or ignore noise intrinsic to the system, compromising accuracy for the sake of optimization ease. We circumvent this compromise by inferring the phase-space probability flow that shares the same time-dependent marginal distributions as the underlying stochastic process. Our approach, probability flow inference (PFI), disentangles force from intrinsic stochasticity while retaining the algorithmic ease of ODE inference. Analytically, we prove that for Ornstein-Uhlenbeck processes the regularized PFI formalism yields a unique solution in the limit of well-sampled distributions. In practical applications, we show that PFI enables accurate parameter and force estimation in high-dimensional stochastic reaction networks, and that it allows inference of cell differentiation dynamics with molecular noise, outperforming state-of-the-art approaches.
\end{abstract}
\maketitle

\section{Introduction}
From gene expression \cite{thattai2001intrinsic, losick2008stochasticity}, collective motion in animal groups \cite{bialek2012statistical, jhawar2020noise}, to growth in ecological communities \cite{grilli2020macroecological}, the behavior of biological processes is driven by a dynamic interplay between deterministic mechanisms and intrinsic noise. In these systems, stochasticity plays a pivotal role, often leading to outcomes that diverge significantly from those predicted by deterministic frameworks \cite{coomer2022noise, maheshri2007living, jhawar2020noise}. To elucidate the respective roles of deterministic forces and intrinsic noise, many inference approaches exploit time-correlations in high-resolution stochastic trajectories \cite{frishman2020learning, bruckner2020inferring}. Comparatively, inferring biological processes at a single-cell level presents a unique challenge: due to the destructive nature of the single-cell omics measurements, we can only observe statistically independent cross-sectional samples from the latent stochastic process \cite{saelens2019comparison, schiebinger2019optimaltransport}. This limitation imposes fundamental constraints on the identifiable dynamics and makes it difficult to disentangle deterministic forces from intrinsic noise \cite{brackston2018transition, weinreb2018fundamental}.

In this study, we investigate this issue for systems that can be modeled as diffusion processes \cite{frishman2020learning}, for which the continuous-time evolution of the degrees of freedom $\mathbf{x} \in \mathbb{R}^d$ follows a stochastic differential equation (SDE) \cite{gardiner2009stochastic}. Under the assumption that there are no exogenous factor or unobserved variable driving the evolution of the system, we consider autonomous It\^o processes of the form
\begin{equation}
d\mathbf{x} = \mathbf{f}(\mathbf{x})dt + \sqrt{2} \mathbf{G}(\mathbf{x})d\mathbf{W},  \label{eq:diffusion}
\end{equation}
where $\mathbf{W}$ is a standard $d$-dimensional Wiener process, $\mathbf{f} : \mathbb{R}^{d} \rightarrow \mathbb{R}^d$ is a deterministic force and $\mathbf{G} : \mathbb{R}^{d} \rightarrow \mathbb{R}^{d \times d}$ the intrinsic noise model. Many molecular processes in cells have been shown to follow negative binomial statistics \cite{ahlmann-eltze2023comparison, choudhary2022comparison}, revealing a strong correlation between the amplitude of regulation and intrinsic noise. As a result, a biologically plausible intrinsic noise model $\mathbf{G}(\mathbf{x})$ can be force and state-dependent.

The formulation via Eq.~(\ref{eq:diffusion}) in terms of stochastic trajectories $\{\mathbf{x}(t), t \geq 0\}$ is equivalent to a formulation in terms of the transition probability $p(\mathbf{x},t| \mathbf{y},s)$, which describes the probability to reach the state $\mathbf{x}$ at time $t$, having started at $\mathbf{y}$ at time $s$ \cite{gardiner2009stochastic}. Its evolution obeys the Kolmogorov forward equation, 
\begin{align}
\partial_t p(\mathbf{x},t | \mathbf{y},s) =& - \nabla \cdot  \bigg[ \mathbf{f} (\mathbf{x}) p(\mathbf{x},t | \mathbf{y},s) \notag \\
&- \nabla \cdot \left( \mathbf{D}(\mathbf{x})  p(\mathbf{x},t | \mathbf{y},s) \right) \bigg], \label{eq:kolmogorov}
\end{align}
for all $\mathbf{x}, \mathbf{y} \in \mathbb{R}^{d}$, $t, s \geq 0$, with $\mathbf{D} = \mathbf{G} \mathbf{G}^T \in \mathbb{S}_{+}^d$. When time-resolved trajectories are available, both the force field and the diffusion can be inferred simultaneously by fitting either of these equations to the data \cite{frishman2020learning}. To simplify computations, most approaches rely on discretizing Eq.~(\ref{eq:diffusion}) rather than fitting transition probabilities with Eq.~(\ref{eq:kolmogorov}), with successful applications in fields such as soft matter and finance \cite{perezgarcia2018highperformance, frishman2020learning, ferretti2020building, kutoyants2004statistical}.

However, with single-cell omics data, the lack of trajectory information makes it impossible to reconstruct the transition probabilities $ p(\mathbf{x},t|\mathbf{y},s)$. Within this setting, it is more appropriate to model the evolution of marginal distributions over time with the Fokker-Planck equation
\begin{equation}
\partial_t p_{t}(\mathbf{x}) = - \nabla \cdot  \left[ \mathbf{f} (\mathbf{x}) p_{t}(\mathbf{x})  - \nabla \cdot \left( \mathbf{D}(\mathbf{x})  p_{t}(\mathbf{x}) \right) \right], \label{eq:fokker}
\end{equation}
which is obtained by marginalizing Eq.~(\ref{eq:kolmogorov}) over an initial condition $p_0(\mathbf{y})$. The inverse problem now reduces to learning how the probability mass is moved between empirical distributions at successive time points rather than how one trajectory evolves. 
Unlike trajectory-based methods, it is no longer possible to infer both the force and noise models simultaneously, requiring a prior on one to infer the other.

Approaches based on optimal transport have been used to tackle this question, first in static settings by learning pairwise couplings between successive empirical distributions, and subsequently in dynamical settings by learning a time-continuous model connecting distributions at all times. While static methods cannot model time-continuous and non-linear dynamics \cite{yang2018scalable, schiebinger2019optimaltransport, schiebinger2021reconstructing, zhang2021optimal, bunne2023learning}, their dynamical counterparts lift these constraints, but all methods remain limited to additive priors on the noise model \cite{tong2020trajectorynet, chizat2022trajectory, lavenant2023mathematical, bunne2023schrodinger}. Other approaches that integrate trajectory data with mechanistic differential equation models also use deterministic frameworks \cite{ocone2015reconstructing} or assume additive noise \cite{matsumoto2017scode, sanchez2018bayesian}. Even popular methods that infer cell-fate directionality from messenger RNA splicing and spatial transcriptomics data also adopt additive noise models for parameter estimation \cite{zhou2024spatial}. However, force and state-dependent noise models not only better capture biological variability, but they also have the capacity to shift, create, or eliminate fixed points in the energy landscape, which is of paramount importance to model processes like cell differentiation \cite{coomer2022noise, losick2008stochasticity, thattai2001intrinsic}. This underscores the need for inference methods that accommodate molecular noise while retaining algorithmic simplicity.

\begin{figure*}[hbt!]
  \includegraphics[width=7in]{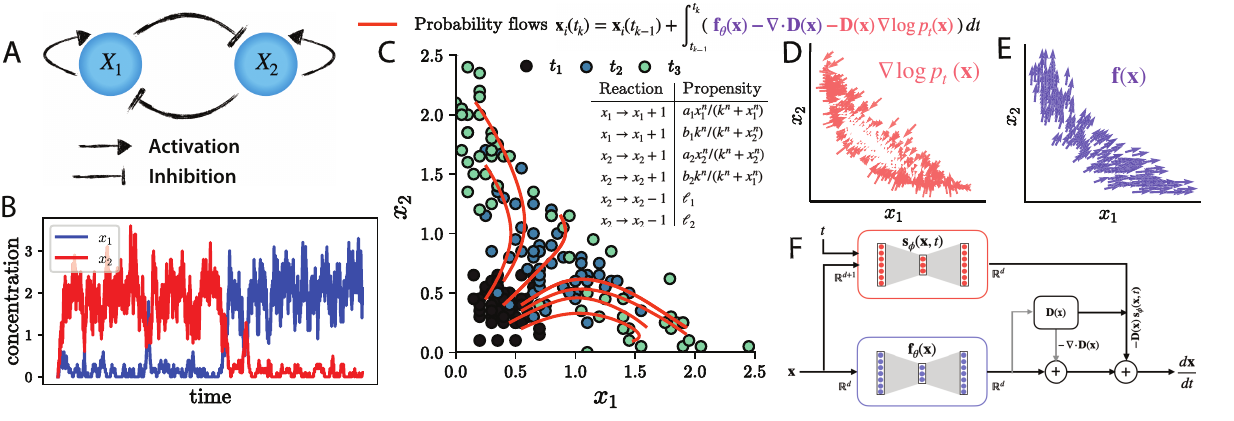}
  \caption{ \textbf{Probability flow inference}. {\bf{A}.} Dynamics of the canonical fate decision regulatory circuit of two mutually opposing transcription factors $(x_1,x_2)$ that positively self-regulate themselves. {\bf{B}.} Time series generated from direct stochastic simulation (Gillespie algorithm) of the toggle-switch model showing the concentration levels of $x_1$ (blue) and $x_2$(red), illustrating their dynamics over time. The Gillespie simulation was run with parameters $a_1=a_2=b_1=b_2=k=1.0$, and $n=4$. The degradation rates are $\ell_1 = \ell_2 = 1$. {\bf{C}.} Cross-sectional snapshots generated from the discrete Gillespie simulations at time points $\{t_1, t_2, t_3\}$ represent the marginals' evolution through the stochastic process. Characteristics lines (in red) show the probability flows deforming the initial state $t_1$ to a future state $t_3$ in the state space, illustrating the dynamic transition of the system's state. The inset describes the reactions and associated propensities of the two-gene fate decision circuit shown in \textbf{A}. Score ({\bf{D}}) and force ({\bf{E}}) approximation at time $t_1$. {\bf{F}.} Network architecture of our probability flow inference (PFI) approach. Both the autonomous force field $\mathbf{f}_\theta(\mathbf{x}): \mathbf{R}^d \rightarrow \mathbb{R}^d$ and the time-dependent score model $\mathbf{s}_\phi (\mathbf{x},t): \mathbb{R}^{d+1} \rightarrow \mathbb{R}^d$ are parameterized using feedforward networks, with parameters $\theta$ and $\phi$, respectively.}
  \label{fig:PF}
\end{figure*}

We introduce Probability Flow Inference (PFI), a method that separates the force field from any intrinsic noise model while retaining the computational efficiency of ODE optimization. First, by analytically solving the PFI approach for Ornstein-Uhlenbeck (OU) processes, we prove that the inverse problem admits a unique solution when the force field is regularized, and that without regularization it can lead to infinitely many solutions. By focusing on a subclass of isotropic OU processes, we show that all non-conservative force contributions can be recovered in the limit of vanishingly small regularization. We contrast this observation by demonstrating the existence of a bias-variance trade-off coming from finite sampling of the cross-sectional data, thereby advocating for a non-zero regularization. Finally, with the same analytical solution we show that a good prior on intrinsic noise is necessary to accurately infer the underlying force field.

Building on these theoretical insights, we apply the PFI approach to stochastic reaction networks, widely used to model cellular processes. We show that PFI reliably infers the continuous diffusion approximation of these discrete-state stochastic processes, with a particular focus on estimating rate parameters and inferring gene regulatory networks. In agreement with our theoretical results for the OU processes, we underscore the importance of an informed biophysical prior on the noise model to achieve accurate force inference. Finally, using a curated hematopoietic stem cell model, we demonstrate that PFI outperforms state-of-the-art generative models in inferring cell differentiation dynamics and predicting \textit{in silico} gene knockdown perturbations.

\section*{Methodology}

\subsection*{Problem statement} 
We assume that the data is given as $K+1$ statistically independent cross-sectional snapshots, each composed of \rev{$n_k$} samples, taken from the true process at successive times $t_0 = 0 < ... < t_k < ... < t_{K} = T$, with uniform spacing $\Delta t$. We further assume that the \rev{$n_k$} i.i.d samples \rev{$\{\mathbf{x}_{i,t_k}, 1 \leq i \leq n_k\}$} are measured at each time \rev{$t_k$} from the true marginals, giving access to an empirical estimator of \rev{$p_{t_k}$}, 
\begin{equation}
{p}_{t_k} (\mathbf{x}) \approx \frac{1}{n_k} \sum_{i = 1}^{n_k} \delta(\mathbf{x} - \mathbf{x}_{i,t_k}). \label{density}
\end{equation}
The objective is to infer the force field $\mathbf{f(x)}$ associated with a latent stochastic process interpolating the observed marginals. Since it is not possible to simultaneously infer both the force field and diffusion from statistically independent cross-sectional samples, we impose strong priors on the noise model. For example, in the case of molecular noise arising from chemical reactions, the diffusion tensor $\mathbf{D(x)}$ is force and state dependent and follows a known functional form \cite{gillespie2000chemical}. We leverage such intrinsic noise priors to accurately infer the force fields and reconstruct the underlying stochastic dynamics.

\subsection*{Probability flow inference (PFI)}
A common approach to inferring the force field $\mathbf{f(x)}$ is to fit the SDE by minimizing a distance metric between the observed empirical marginals and the generated samples, in a \textit{predict and correct} manner \cite{hashimoto2016learning, li2020scalable}. However, this optimization is both memory- and compute-intensive, and is limited to additive or diagonal noise models \cite{li2020scalable}. As previously noted, we can improve on this by observing that the SDE formulation Eq.~(\ref{eq:diffusion}) contains redundant information for modeling the evolution of marginals, and that Eq.~(\ref{eq:fokker}) is sufficient. 

Rather than fitting the SDE, we \rev{instead fit the Fokker–Planck equation directly. This can be accomplished using its Lagrangian-frame formulation, known as the probability flow ODE \cite{song2020scorebased}, which reads:
\begin{equation}
\frac{d \mathbf{x}}{dt}= \mathbf{f(x)} - \nabla \cdot \mathbf{D(\mathbf{x})} - \mathbf{D}(\mathbf{x}) \nabla \log p_t(\mathbf{x}). \label{eq:Probflow}
\end{equation}
The term $\nabla \log p_t(\mathbf{x})$, or the gradient of the log-probability of the marginals, is known as the score \cite{hyvarinen2005estimation}. Solving this ODE from the initial condition $p_0(\mathbf{x})$ generates samples from the same marginal distributions $p_t(\mathbf{x})$ as the underlying SDE. Notably, this property has been used to simulate the Fokker-Planck equation \cite{maoutsa2020interacting, boffi2023probability}, as well as to estimate entropy production in active matter sytems \cite{boffi2024deep}.} The PFI approach consists of two steps: (i) estimating the score function, and (ii) fitting the probability flow ODE Eq.~(\ref{eq:Probflow}) to the observed marginals. We now outline these two steps: 
\paragraph*{Score estimation:}The first step of the PFI approach requires estimating the time-dependent score function from empirical samples at various time points. To efficiently solve this task, we leverage recent advancements in generative modeling that allow fast and accurate score estimation in high dimensions \cite{song2020scorebased, song2020sliced}. Specifically, we use sliced score matching (Materials and Methods) to train a score network $\mathbf{s}_\phi(\mathbf{x}, t)$ that approximates $\nabla \log p_t(\mathbf{x})$.
\paragraph*{Force inference:} \rev{Once an accurate score model is available, we seek to fit the force via Eq.~(\ref{eq:Probflow}), following the \rev{\textit{predict and correct}} strategy outlined in Algorithm 1 (see Appendix~\ref{sec:algorithm}).  Using the estimated score function and an initial guess of the force field, we push the observed samples from time $t_k$ to $t_{k+1}$ using the probability flow ODE in Eq.~(\ref{eq:Probflow}), generating a predicted distribution $\hat{p}_{t_{k+1}}(\mathbf{x})$ as defined in Eq.~(\ref{density}). The force field is then optimized to minimize a distance $\mathcal{D}$ between the predicted distribution and the observed distributions. In practice, we minimize the total discrepancy across all cross-sectional time points, i.e.,
$\hat{\mathbf{f}} = \argmin_\mathbf{f}\sum_{i=1}^K \mathcal{D} (\hat{p}_{t_i}, p_{t_i})$.}\\

In Fig.~\ref{fig:PF}, we illustrate the two step PFI approach applied to a bistable genetic switch system (see Fig.~\ref{fig:PF}A-B). The probability flows in transcription factor concentration space (Fig.\ref{fig:PF}C) have two components: one derived from the estimated score (Fig.~\ref{fig:PF}D), and the other from the force field (Fig.~\ref{fig:PF}E). Finally, as shown in Fig.~\ref{fig:PF}F, both the score and the force field can be parameterized using feedforward neural networks.

\rev{In many cases it is impossible to uniquely identify a force field matching the observed marginals. To see this, consider a force field $\hat{\mathbf{f}}(\mathbf{x}) = {\mathbf{f}}(\mathbf{x})  + {\mathbf{h}}(\mathbf{x})$. Using Eq.~(\ref{eq:fokker}), we see that $\mathbf{f}(\mathbf{x})$ and $\hat{\mathbf{f}}(\mathbf{x})$ generate the same marginals $p_t(\mathbf{x})$ if $\mathbf{h}(\mathbf{x})$ satisfies 
\begin{equation}
\nabla \cdot (\mathbf{h}(\mathbf{x}) p_t(\mathbf{x})) = 0, \; \forall \; t \geq 0, \forall \mathbf{x}.\label{identify}
\end{equation}
This equation can have multiple non-gradient solutions (see Appendix \ref{sec:recovery}). For instance, if $p_t(\mathbf{x})$ has radial symmetry in $\mathbf{x}$, any force field $\mathbf{h}(\mathbf{x}) = \mathbf{K} \mathbf{x}$, with $\mathbf{K}$ a skew-symmetric matrix, is a solution of Eq.~(\ref{identify}). This lack of uniqueness, referred to as the identifiability issue, has been of long standing concern in the analysis of single-cell RNA-seq data \cite{weinreb2018fundamental, hashimoto2016learning, coomer2022noise}.}

In statistical physics, the component of the force $\mathbf{f}(\mathbf{x})$ that satisfies equation Eq.~(\ref{identify}) in the limit $t \rightarrow \infty$ is referred to as the non-conservative force as it induces
non-zero phase-space probability currents at steady-state \cite{gardiner2009stochastic}. The remaining part of the force is termed the conservative force and it alone balances the effect of diffusion. While the conservative force can always be inferred, even with data at steady state, the identifiability issue tells us that non-conservative forces may not be uniquely identifiable from the temporal evolution of the marginals. \rev{In fact, if the marginals do not evolve in time, the underlying non-conservative forces simply can't be recovered. This has consequences, because the knowledge of such non-conservative forces is instrumental to accurately predict cell differentiation and cell reprogramming pathways \cite{wang2008potential, wang2011quantifying}. }

To circumvent this ill-posedness, we choose to introduce an $\ell_2$-regularization on the Jacobian matrix of the inferred force field to the loss function which reads
\begin{align}
\mathcal{L}_{\Delta t, K} &=  \sum_{i=1}^{K} \Bigg[\mathcal{W}_2^2 (\hat{p}_{t_i}( \mathbf{x}), p_{t_i}( \mathbf{x}) ) \notag \\
&+ \lambda \Delta t \int_{t_{i-1}}^{t_i} \int \!\!\| \nabla \hat{\mathbf{f}} (\mathbf{x}) \|_F^2 {p}_{t_i}(\mathbf{x}) d\mathbf{x} dt \Bigg]. \label{eq:total_loss}
\end{align}
Here $\lambda \geq 0$ is a tunable parameter that controls the strength of regularization, and we use the Wasserstein distance $\mathcal{W}_2$ to quantify the discrepancy between distributions. Though the above regularization penalizes the gradient of the force field, a regularization minimizing the kinetic energy is also appropriate \cite{tong2020trajectorynet}.

Computing the exact Wasserstein distance $\mathcal{W}_2$ requires $\mathcal{O}(n^3 \log n)$ operations, and its estimation in $d$ dimensions has a sample complexity of $\mathcal{O}(n^{-1/d})$. Its entropy-regularized version, known as the Sinkhorn divergence, reduces the computational cost to $\mathcal{O}(n^2)$ with a dimension-independent sample complexity $\mathcal{O}(n^{-1/2})$ for large entropic regularization \cite{genevay2018learning, genevay2019sample}. \rev{In all our numerical examples, we use the Sinkhorn divergences to approximate the Wasserstein distance. Overall, the PFI approach allows for the use of accurate forward solvers \cite{atkinson1991introduction} and constant-memory gradient computations \cite{chen2018neural} to fit the force field. More broadly, and relevant to PFI, reverse-mode automatic differentiation tools now allow the fitting of ODEs with millions of parameters to data \cite{chen2018neural}, enabling more flexible approaches to density estimation and time series modeling \cite{grathwohl2018ffjord, rubanova2019latent}.} Before stepping into numerical examples, in the next section we study analytically the identifiability issue for OU processes.

\color{black}

\begin{figure*}[hbt!]
  \includegraphics[width=7in]{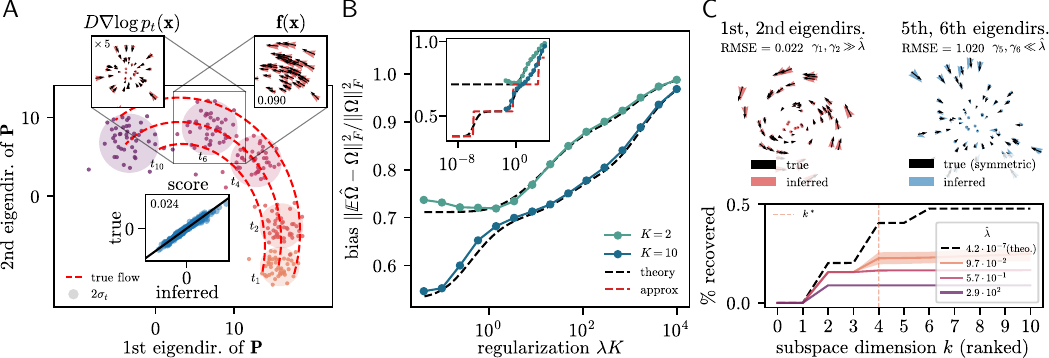}
  \caption{\textbf{Theoretical insight for the inference of a Ornstein-Uhlenbeck process.} {\bf A.} Illustration of the inference process: the inferred process samples are pushed from the true process samples at time $t_i$ to time $t_{i+1}$ using the PF ODE.  The inferred score and inferred force field are shown (red arrows) overlayed on the true score and the true force field (dark arrows) at time $t_6$. The inferred force has an error of $9\%$ (computed over the whole state-space). As shown in inset for $t_6$, the score inference is also accurate with an error of $2.4\%$ {\bf B.} Bias of the inferred interaction matrix as a function $\tilde{\lambda} = \lambda K$. In dashed lines are shown the theory (dark) and the dots correspond to numerical minimization of Eq.~(\ref{eq:total_loss}) using a pre-computed score model. In inset the bias is shown for a wider range of $\tilde{\lambda}$, and the dashed red line shows the approximate solution. {\bf{C.}} In the two upper panels are shown, at $\tilde{\lambda} \simeq 10^{-1}$, projections of the inferred force field (centered in $\mathbf{x} = 0$) in the 1st and 2nd eigendirections of $\mathbf{P}$ (red arrows), and in the 5th and 6th (blue arrows). In the leading eigendirections the force field inferred matches the true force field, while as the weaker eigendirections the inference recovers the symmetric part of the true force field. In the lower panel is shown the $\%$ of true skew-symmetric interactions recovered in the $k$-leading eigendirections of $\mathbf{P}$ . For $\tilde{\lambda} \simeq 10^{-1}$, beyond the effective dimension $k^*=4$ the inference only infers equilibrium (here symmetric) dynamics. The parameters used are $d = 10, n = 8000, \|m_0\| = 20, \Sigma_0 = 1, D = 8, \Omega_s = 2, \Delta t = 0.05$, $\mathbf{\Omega}_a = 3 \Omega_s \mathbf{A}$, with $\mathbf{A}$ a skew-symmetric matrix is chosen at random (Materials and Methods). The score is pre-computed with the same samples (\app{App.}{sec:SM_score})}
  \label{fig_ou_process}
\end{figure*}

\subsection*{Analytical case study: Ornstein-Uhlenbeck process}
Linear models are a very popular choice for gene regulatory network inference \cite{yuan2021cellbox}, and reconstructing cellular dynamics based on RNA velocity \cite{la2018rna, bergen2020generalizing}. Despite their frequent use for such inference tasks, the challenges related to the identifiability issue, the role of regularization, and various sources of error have not been addressed. In this section we tackle this issue in a continuous-time limit, $\Delta t \rightarrow 0$, with which we establish a uniqueness result for the inferred process in the presence of regularization.  For this purpose, we assume that the underlying latent process to be inferred is a $d$-dimensional OU process with an interaction matrix $\mathbf{\Omega}$. That is,
\begin{equation}
d\mathbf{x} = \mathbf{\Omega} \mathbf{x} dt + \sqrt{2\mathbf{D}} d\mathbf{W}, \: \text{with } \mathbf{x}_0 \sim p_0(\mathbf{x}),
\end{equation}
where $\mathbf{\Omega}$ has eigenvalues with strictly negative real part. When $p_0 = \mathcal{N}(\mathbf{m}_0, \mathbf{\Sigma}_0)$, the solution of the OU process is Gaussian at all times with $\mathbf{x}_t \sim \mathcal{N} ( \mathbf{m}_t, \mathbf{\Sigma}_t )$, with $\mathbf{\Sigma}_t$ and $\mathbf{m}_t$ being, respectively, the covariance and mean of the process at time $t$ \cite{sarkka2019applied}. We assume that $\mathbf{\Sigma}_0$ is full rank, so that $\mathbf{\Sigma}_t$ is positive definite at all later times \cite{sarkka2019applied}. The covariance matrix $\mathbf{\Sigma}_t$ can be decomposed as $\mathbf{\Sigma}_t = \sum_i \sigma_{i,t}^2 \mathbf{w}_{i,t} \mathbf{w}_{i,t}^T$, where  $\sigma_{i,t}$ are its eigenvalues and $\mathbf{w}_{i,t}$ are the corresponding eigenvectors. In this problem, we restrict the inferred force model to be linear, $\hat{\mathbf{f}}(\mathbf{x}) = \hat{\mathbf{\Omega}}\mathbf{x}$, and the diffusion tensor to be a given constant $\hat{\mathbf{D}}$, perhaps previously estimated. We present an analytical form of the loss function (Eq.~(\ref{eq:total_loss})) as a function of $\hat{\mathbf{\Omega}}$ in the continuous-time limit $(\Delta t \rightarrow 0)$ and large sample limit ($n \rightarrow \infty$).

\paragraph*{Continuous-time loss function.} With the following theorem, we prove in \app{App.}{sec:SM_ouprocess}, that the loss in Eq.~(\ref{eq:total_loss}) converges to a strongly convex loss for $\lambda > 0$. 
\begin{theorem} \label{theo_loss}
With $K = \lfloor T/ \Delta t \rfloor$, when $n\rightarrow \infty$ and $\Delta t \rightarrow 0$, the loss function $\mathcal{L}_{\Delta t, K} /\Delta t \rightarrow$  $\mathcal{L}$ with
\begin{align}
\mathcal{L}& = \mathrm{tr} \left( (\hat{\mathbf{\Omega}} - \mathbf{\Omega}) \mathbf{P} (\hat{\mathbf{\Omega}} - \mathbf{\Omega} )^T  + \lambda T \hat{\mathbf{\Omega}} \hat{\mathbf{\Omega}}^T\right) \notag \\
&+ \int_0^T  \sum_{i,p} \frac{\sigma_{i,t}^2}{\left( \sigma_{i,t}^2 + \sigma_{p,t}^2\right)^2} \bigg(\mathbf{w}_{i,t}^T \bigg(\sigma_{p,t}^2 (\hat{\mathbf{\Omega}} - \mathbf{\Omega})\notag \\&+ \sigma_{i,t}^2 (\hat{\mathbf{\Omega}}^T - \mathbf{\Omega}^T) + 2(\hat{\mathbf{D}} - \mathbf{D})\bigg) \mathbf{w}_{p,t} \bigg)^2 dt, \label{continuous_loss} \\
&\text{where } \mathbf{P} = \int_0^T \mathbf{m}_t \mathbf{m}_t^T \notag.
\end{align}
For $\lambda > 0$ this loss function is strongly convex, and thus has a unique minimum on $\mathbb{R}^d$.
\end{theorem}
\noindent In \app{App.}{sec:SM_sinkhorn}, we also provide a version of this theorem for the Sinkhorn divergence with entropic regularization $\epsilon$, of which Eq.~(\ref{continuous_loss}) is the particular case when $\epsilon \rightarrow 0$. 

In the discrete-time setting, the asymptotic loss function $\mathcal{L}$ of Eq.~(\ref{continuous_loss}) is achieved once $\Delta t$ is sufficiently small to resolve all the relevant timescales. More specifically, when $\Omega_{\mathrm{max}}  \Delta t, D \Delta t/\sigma_{\mathrm{min}} \ll 1$ where $\Omega_{\mathrm{max}}$ denotes \rev{the largest singular value of $\mathbf{\Omega}$} and $\sigma_{\mathrm{min}}$ is the smallest eigenvalue of $\mathbf{\Sigma}_0$. The $\lambda$ regularization ensures that the objective function is strongly convex, as without regularization we can face the identifiability issue. \rev{In particular, when $\lambda = 0$ and $\hat{\mathbf{D}} = \mathbf{D}$, the loss reaches $\mathcal{L} = 0$ when $\hat{\mathbf{\Omega}} = \mathbf{\Omega}$, and any matrix of the form $\hat{\mathbf{\Omega}} = \mathbf{\Omega} + \mathbf{K}$ which satisfies
\begin{equation}
\mathbf{K}\mathbf{m}_t = 0, \; \mathbf{K} \mathbf{\Sigma}_t + \mathbf{\Sigma}_t \mathbf{K}^T = 0, \forall t \in [0,T]. \label{ou_identify}
\end{equation}
is also a minimum. Because of this, when $\lambda = 0$, the matrix $\mathbf{\Omega}$ is uniquely identifiable when the unique solution to this system of equations is $\mathbf{K} = 0$. This recovers an identifiability criterion analogous to that presented in \cite{guan2024identifying}.}

\begin{figure*}[hbt!]
  \includegraphics[width=7in]{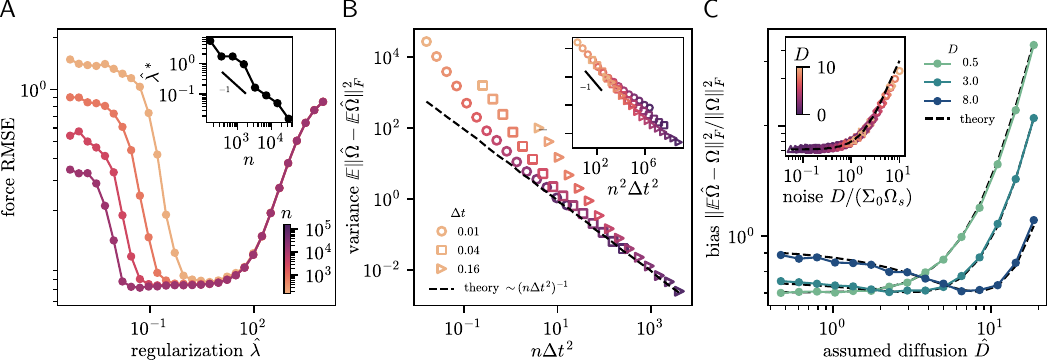}
  \caption{\textbf{Sources of error for the inference of an Ornstein-Uhlenbeck process.} {\bf A.} Bias-variance trade-off as a function of the regularization $\tilde{\lambda}$ in the relative mean square error of the force field. The inflection point of the trade-off is reached at smaller $\tilde{\lambda}$ for larger sample sizes $n$. {\bf B.} Variance on the inferred interaction matrix as a function of $n \Delta t^2$, with $\Omega_s \Delta t = 0.1, D/\Sigma_0 \Omega_s = 4$. For various $\Delta t$ the variance collapses at large $n$ on the prediction (black dashed line). As shown in inset, at smaller $n$ the scaling is found to be $n^2 \Delta t^2$. {\bf C.} Bias on the inferred interaction matrix as a function of the assumed diffusion tensor $\hat{\mathbf{D}} = \hat{D} \mathbf{S}$ with $\mathbf{S}$ a semi-definite positive symmetric matrix with unit maximum eigenvalue (Materials and Methods). The bias shows a minimum when $\hat{D} = D$. When $\hat{D} = 0$ (which corresponds to fitting a deterministic model), the bias increases with $D$. This is shown in inset as a function of the non-dimensional strength of noise $D/\Sigma_0 \Omega_s = 3\tau_{\mathrm{force}}/\tau_{\mathrm{diff}}$ for different values of $\Sigma_0$ and $\Omega_s$. With respectively increasing values for $\Sigma_0 \in \{0.5, 2, 3\}$, square, circle and right triangle markers correspond to $\Omega_s = 2$, while diamonds, left and upper triangle markers corresponds to $\Omega_s = 3$. The parameter $\|m_0\|^2/\Sigma_0$ is fixed to $400$ (s.t $\|m_0\| = 20$ when $\Sigma_0=1$). The color map for $n$ is shared between {\bf A.} and {\bf B.}. In {\bf B.} and {\bf C.} the regularization is $\tilde{\lambda} = 0.2$. In the three panels $K = 5$, $\mathbf{\Omega}_a = 3\Omega_s \mathbf{A}$ and all the remaining free parameters (notably $\mathbf{A}$) in each panel are the same as Fig.~\ref{fig_ou_process}. We use the analytical prediction for the score.}
  \label{fig_ou_error}
\end{figure*}
 
\paragraph*{An analytically solvable isotropic process} To gain more insight into the role of the regularization, we simplify our model by considering an isotropic process, where $\mathbf{\Sigma}_0 = \sigma_0^2 \mathbf{I}$, $\mathbf{D} = D \mathbf{I}$, and $\mathbf{\Omega} = \Omega_s \mathbf{I} + \mathbf{\Omega}_a$, with $\mathbf{\Omega}_a$ a skew-symmetric matrix. As a result, the covariance matrix is isotropic at all times, i.e. $\mathbf{\Sigma}_t = \sigma_t^2 \mathbf{I}$ \cite{sarkka2019applied}. For such an isotropic OU process the non-conservative and conservative forces correspond to the skew-symmetric and the symmetric parts of $\mathbf{\Omega}$, respectively. The non-conservative force $\mathbf{\Omega}_a \mathbf{x}$ generates rotations around the origin $\mathbf{x} = 0$, while the conservative part $\Omega_s \mathbf{x}$ induces inward flows.

We illustrate the PFI approach with a numerical example of an isotropic OU process in $d = 10$ dimensions, using $K = 10$ snapshots, $n = 8000$ samples and with $\lambda = 10^{-2}$. Fig.~\ref{fig_ou_process}A shows the evolution of the true OU process projected along the two leading eigendirections of $\mathbf{P}$, with the true probability flow (dashed red lines) interpolating the successive Gaussian marginal distributions (shaded discs). The probability flow lines spiral towards the origin under the combined effect of the non-conservative and conservative forces, while diffusion is reflected in the spreading over time of the distributions. In this example, the relative strength of the non-conservative and conservative forces is $3$ to $1$, while the time scales of diffusion and non-conservative forces are comparable (as measured by the ratio $\tau_{\mathrm{force}}/ \tau_{\mathrm{diff}} = D/(\Sigma_0 \Omega_{\max})$). The score estimated by sliced score-matching (Materials and Methods) is accurate (RMSE of $\sim 2.5\%$), as shown in the inset by comparing it to its analytical prediction $\nabla \log p_t(\mathbf{x}) \sim- \Sigma_t^{-1}(\mathbf{x} - \mathbf{m}_t)$. Using this score model, we infer the force using the Gaussian Wasserstein estimator to avoid the curse of dimensionality associated with the empirical Wasserstein distance \cite{genevay2018learning}. At time $t_6$, we overlay the inferred force field on the true force field, and find that we accurately infer the force with RMSE $\sim~9\%$.

\rev{Because the covariance is isotropic at all times, we can directly see from Eq.~(\ref{ou_identify}) that when the range of $\mathbf{P}$ does not span $d - 1$ dimensions, Eq.~(\ref{ou_identify}) has non-trivial skew-symmetric solutions and some non-conservative terms in the force field are simply inaccessible to the inference.} Decomposing $\mathbf{P} = \sum_{i} \gamma_i \mathbf{u}_i \mathbf{u}_i^T$ in terms of its eigenvalues and eigenvectors, we derive in \app{App.}{sec:SM_isotropic}, an analytical formula for $\hat{\mathbf{\Omega}}$, the minimum of the loss for $\lambda > 0$. This formula shows an excellent agreement with the PFI solution, as shown in Fig.~\ref{fig_ou_process}B by plotting the bias $\|\mathbb{E}\hat{\mathbf{\Omega}} - \mathbf{\Omega}\|_F^2/\|\mathbf{\Omega}\|_F^2$ as a function of $\tilde{\lambda} = \lambda K$ for $K \in \{ 2, 10 \}$. We can gain real insight into the role of the regularization using an approximation to $\hat{\mathbf{\Omega}}$:
\begin{equation}
\hat{\mathbf{\Omega}} \approx \mathbf{\Omega} - \mathbf{Q}_{\tilde{\lambda}} \mathbf{\Omega}_a \mathbf{Q}_{\tilde{\lambda}} \label{bias_lambda},
\end{equation}
where $\mathbf{Q}_{\tilde{\lambda}} = \sum_i \mathbf{\chi} (\gamma_i < \tilde{\lambda}) \mathbf{u}_i \mathbf{u}_i^T$ is the projector onto the eigenspace of $\mathbf{P}$ having eigenvalues smaller than the penalty $\tilde{\lambda}$, with $\mathbf{\chi}$ the indicator function. This approximation, valid for $\tilde{\lambda} \ll \sum_{j=0}^{K-1} \sigma_t^2$ and $|\tilde{\lambda}^{-1} \gamma_i - 1| \gg 1$ for all $i$, is shown in inset of Fig.~\ref{fig_ou_process}B (dashed red) in comparison with the analytical solution (dashed black). This solution suggests that the non-conservative force $\mathbf{\Omega}_a \mathbf{x}$ is inferred only in the subspace spanned by $\{\mathbf{u}_i \vert \gamma_i \gg \tilde{\lambda}\}$, and is set to zero in its orthogonal complement. In these remaining directions, only the conservative force is correctly estimated. This idea is further exemplified in Fig.~\ref{fig_ou_process}C (upper panel) by showing the projections of the inferred force field on eigendirections of $\mathbf{P}$ sorted by decreasing eigenvalues $\gamma_i$. We can observe that the force field is fully estimated in the leading eigendirections, for which $\gamma_i \gg \tilde{\lambda}$, but that only the symmetric contribution is inferred when $\gamma_i \ll \tilde{\lambda}$. This observation is rationalized in Fig.~\ref{fig_ou_process}C \rev{(lower panel)} where we plot the fraction of recovered skew-symmetric interactions in the subspace spanned by the $k$ leading eigenvectors of $\mathbf{P}$. This fraction is measured by $\|\mathbf{Q}_{k} \hat{\mathbf{\Omega}}_a \mathbf{Q}_{k} \|^2_F / \|\mathbf{\Omega}_a\|^2_F$, where $\mathbf{Q}_k$ is the projector on subspace spanned by the $k$ eigenvectors associated with the $k$ largest eigenvalues of $\mathbf{P}$. We see that beyond an effective dimension $k^*$ the fraction plateaus, suggesting that non-conservative forces are not recovered in the remaining eigendirections. In other words, from the standpoint of inference, the time-dependent deformations of the marginal distributions in the subspace associated with the $d - k^*$ smallest eigenvalues of $\mathbf{P}$ are indistinguishable from equilibrium dynamics.

These findings highlight the role of regularization as a recovery threshold for non-equilibrium dynamics, and taking $\lambda \rightarrow 0$ ensures exact recovery of all non-conservative forces available in the data. However, decreasing $\tilde{\lambda}$ also incurs an increased variance coming from the finite sample size $n$. As shown in Fig.~\ref{fig_ou_error}A, we observe a bias-variance trade-off in the expected relative mean square error for the interaction matrix $\mathbb{E} \| \hat{\mathbf{\Omega}} - \mathbf{\Omega}\|^2_F/\|\mathbf{\Omega}\|^2_F$. In \app{App.}{sec:SM_variance}, we derive the first-order, finite sample size correction to the continuous-time loss function, allowing us to estimate analytically the variance $\mathbb{E} \| \hat{\mathbf{\Omega}} - \mathbb{E} \hat{\mathbf{\Omega}} \|^2_F$. Given $K$, with all non-dimensional quantities being fixed and for $n$ large, we predict the variance to read as \rev{$C/(n \Delta t^2)$}, with $C$ a constant that captures the magnitude of deformation of the marginals. \rev{In \app{App.}{sec:SM_variance}, we derive $C$ analytically, and show that $C$ diverges as the marginals become indistinguishable from the steady state distribution. The scaling $(n \Delta t^2)^{-1}$} is in excellent agreement with the PFI solution, as shown by the collapse in Fig.~\ref{fig_ou_error}B as $\Delta t$ is varied. Additionally, in inset we observe that at smaller sample sizes the variance scales as $(n^2 \Delta t^2)^{-1}$, which we hypothesize comes from higher order terms in the finite sample size correction to the loss function. These results highlight the rather subtle balance between the regularization $\lambda$, the sample size $n$, and the time-step $\Delta t$, necessary to minimize the error. 

In practice, this interplay strongly depends on the parametrization of the force field. For example, to go beyond linear models, the force field can be expressed as a linear combination of function basis \cite{frishman2020learning} or parameterized using a neural network \cite{yeo2021generative, tong2020trajectorynet}. These choices introduce implicit regularization, such as biases arising from the smoothness of the selected basis (e.g., Fourier) or activation functions in neural networks \cite{mishra2022estimates}, and simplicity biases inherent to deep neural network models \cite{perez2019deep}. Such implicit regularizations provide additional ways to constrain the solution space beyond the explicit regularization discussed here, ensuring unique solutions. For this reason, for all practical purposes, the choice of the explicit regularization is problem-specific and usually determined in an ad-hoc manner \cite{bishop2006pattern, hastie2009elements}.

Finally, our analytical solution allows us to study the error coming from the misestimation of the diffusion tensor $\hat{\mathbf{D}}$. Assuming that $\hat{\mathbf{D}} = \hat{D}\mathbf{S}$, where $\mathbf{S}$ is an arbitrary matrix in $\mathbb{S}^d_{+}$ (Materials and Methods) with unit maximum eigenvalue, we plot in Fig.~\ref{fig_ou_error}C the bias $\| \mathbb{E} \hat{\mathbf{\Omega}} - \mathbf{\Omega}\|^2_F$ as a function of $\hat{D}$. These results show that the bias is minimum when $\hat{D} = D$, \rev{and that using a deterministic noise model, where $\hat{D} = 0$, results in a larger error as the intrinsic noise $D$ increases}. This observation is rationalized using the analytical solution (\app{App.}{sec:SM_isotropic}), which predicts that for $\lambda$ small, with $K$ and all other non-dimensional quantities fixed, the bias at $\hat{D} = 0$ (which corresponds to fitting a deterministic model) is a known function of $\tau_{\mathrm{force}}/ \tau_{\mathrm{diff}}$ the ratio of the timescales associated with the force and with the diffusion. This prediction agrees very well with the PFI solution, as shown in the inset of Fig.~\ref{fig_ou_error}C for various values of $\Sigma_0$ and $\Omega_s$.  This shows that estimating $\tau_{\mathrm{force}}/ \tau_{\mathrm{diff}}$ is an efficient way to decide whether fitting a deterministic model is sufficient ($\tau_{\mathrm{force}}/ \tau_{\mathrm{diff}} \ll 1$) or noise is necessary ($\tau_{\mathrm{force}}/ \tau_{\mathrm{diff}} \gtrsim 1$). 

In more general cases, it may not be possible to quantify the relative strengths of intrinsic noise and force. However, due to the Poisson nature of the chemical reactions underlying cellular processes, the intrinsic noise variance is often closely linked to the amplitude of the force \cite{ahlmann-eltze2023comparison, choudhary2022comparison}, resulting in comparable timescales for both intrinsic noise and force (i.e. $\tau_{\mathrm{force}} \sim \tau_{\mathrm{diff}}$ ). Therefore, accounting for intrinsic noise is crucial to accurately infer cellular processes from single-cell omics data. In the following, we perform numerical experiments to demonstrate how the PFI approach offers a flexible mechanism to incorporate stochasticities of arbitrary nature, and in particular molecular noise.

\begin{figure*}[hbt!]
  \includegraphics[width=7in]{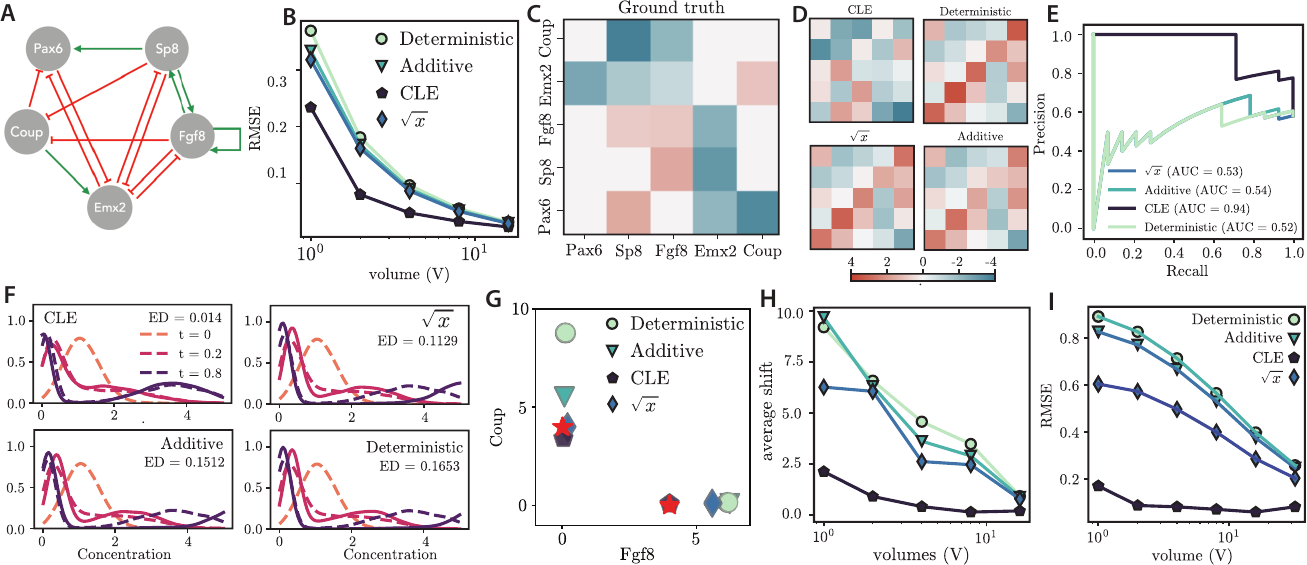}
  \caption{\textbf{Parameter and force estimation
for stochastic reaction networks.} {\bf{A.}} mCAD gene regulatory network. Red lines denote inhibition and green arrows activation. {\bf{B.}} The RMSE $ \Vert \mathbf{g(x)} - \mathbf{\hat{g}(\mathbf{x})}  \Vert_2^2/\Vert \mathbf{g(x)} \Vert_2^2 $ of the inferred force shown for different compartment volumes $V$. The Jacobian corresponding to the true force field $\mathbf{f(x)}$ ({\bf{C}}), with estimated Jacobians $\nabla \hat{\mathbf{g}}(\mathbf{x})$ obtained under different noise models shown in \textbf{D}. The associated precision-recall (PR) curves are displayed in \textbf{E}, with the area under the curve (AUC) values reported in the inset. {\bf{F.}} Comparison of empirical marginals from the inferred diffusion process (solid) and Gillespie simulations (dashed) for $V=4$, under various noise models (inset). The inset also reports the Energy Distance (ED) between predicted marginals and Gillespie simulations for each noise model. \textbf{G}-\textbf{H}: Visualization of fixed points, which are solutions to the equation $\mathbf{\hat{g}(\mathbf{x})} - \ell \mathbf{x} = 0$, where $\mathbf{\hat{g}(\mathbf{x})}$ represents the force fields inferred under different noise models. Different markers indicate the fixed points derived from the corresponding noise models, plotted against the \textit{Pax6} and \textit{Coup} genes. Results are shown for two reaction volumes: $V=4$ (\textbf{G}) and $V=16$ (\textbf{H}). The red star represents the true fixed point (Ground Truth, $\mathbf{x}^*$), while the marginal densities at the initial state $p_0(\mathbf{x})$ (orange) and final state $p_T(\mathbf{x})$ (blue) are shown as contour plots. {\bf{I.}} (\textbf{I}) The average shift $\Vert \mathbf{x} - \mathbf{x}^* \Vert_2$ as a function of reaction volume $V$, for different noise models.}
  \label{Fig_CRN}
\end{figure*}

\section*{Numerical results} \label{sec:numerical}

In this section, we exploit the flexibility of PFI approach to infer gene regulatory networks and and model cell differentiation dynamics, incorporating molecular noise. 

\subsection*{PFI allows accurate parameter and force estimation for stochastic reaction networks}

Cellular processes are driven by an intricate array of chemical reactions \cite{mcadams1997stochastic},\cite{raser2005noise}. While techniques like flow cytometry, microscopy, and high-throughput omics provide extensive data on cellular processes, interpreting this data, their variability, and estimating reaction rate constants from it requires mechanistic models. Under the assumption that the system is well-mixed, the chemical master equation (CME) offers a detailed probabilistic representation of these stochastic reaction networks (SRNs) \cite{gillespie1992rigorous, schnoerr2017approximation}. However, the use of CME is limited in both simulation and inference tasks due to the significant computational complexity involved in solving it \cite{frohlich2017scalable, zechner2014scalable, drovandi2011approximate}. Diffusion approximations, such as the Chemical Langevin Equation (CLE), offer a computationally tractable alternative by approximating the discrete CME with a continuous diffusion process. These methods accurately capture stochastic effects at moderate molecule counts \cite{mozgunov2018review, schnoerr2017approximation}, making them a practical and effective approximation to the CME.  

However, when dealing with single-cell omics data, the stoichiometry of gene regulatory networks is unknown. In addition, simultaneous measurements of both protein and mRNA counts are generally unavailable, complicating efforts to develop detailed descriptions of the underlying stochastic regulatory networks. To address this challenge, simple coarse-grained models have been introduced that leverage the separation of timescales between transcription factor binding to regulatory DNA sites and the processes of transcription and translation \cite{marbach2010revealing, bower2001computational}. One such model describes the stochastic evolution of mRNA counts, denoted by \( \mathbf{x} \in \mathbf{R}^d \), with $d$ the number of genes, assuming that the transcription rate of gene \( i \) is proportional to an activation function $g_i(\mathbf{x},V)$, \( 0 < g_i(\mathbf{x},V) < 1 \) with $V$ being the reaction volume, typically that of the nucleus. Under well-mixed assumptions, the stochastic mRNA dynamics are then approximated by the CLE
\begin{equation}\label{eq:CLE}
        d\mathbf{x} = \left( m V \mathbf{g}(\mathbf{x},V) - \ell \mathbf{x} \right)dt +  \sqrt{ m V \mathbf{g}(\mathbf{x},V) + \ell \mathbf{x}} \, d\mathbf{W},
\end{equation}
where \( \ell \) is the degradation rate of mRNA molecules, $m$ is the transcription rate, and the square root is taken entrywise. \rev{This corresponds to the following deterministic force and diffusion tensor in the probability flow ODE:
\begin{align}
    \mathbf{f(x)} &= m V \mathbf{g}(\mathbf{x},V) - \ell \mathbf{x},\\
    \mathbf{D(x)} &= \frac{1}{2}\text{diag}(mV g_1(\mathbf{x},V)+\ell x_1,\cdots,mV g_d(\mathbf{x},V)+\ell x_d).
\end{align}}
Although the CLE in Eq.~(\ref{eq:CLE}) is a simplified approximation to the underlying SRNs governing gene regulation (see Materials and Methods), it has been shown to quantitatively reproduce experimental steady-state single-cell transcriptomics profiles for known gene regulatory networks \cite{dibaeinia2020sergio}. By leveraging the CLE formulation in Eq.~(\ref{eq:CLE}), the PFI approach can be readily applied to infer parameters and forces in high-dimensional SRNs. To demonstrate this, we consider the  Mammalian Cortical Area Development (mCAD) gene regulatory network (see Fig.~\ref{Fig_CRN}, $d = 5$) and use the boolODE framework \cite{pratapa2020benchmarking} to compute $\mathbf{g}(\mathbf{x})$. To generate the marginal data, we conduct a detailed simulation of the mCAD stochastic reaction network using the Gillespie algorithm (Materials and Methods), producing \( K = 10 \) snapshots, each containing \( n = 6,000 \) samples. 

We begin the PFI procedure by training a score network $\mathbf{s}_\phi(\mathbf{x},t)$ to estimate the score function from the marginal data in concentration space $\mathbf{x}/V$ (\app{Fig.}{fig:score_mCAD}). Using the PFI approach, we fit the Eq.~(\ref{eq:CLE}) to the cross-sectional data by minimizing the loss function of Eq.~(\ref{eq:total_loss}). We parameterize the force using a feed-forward neural network consisting of four fully connected layers, each with 50 nodes and smooth ELU activation (see Materials and Methods). The explicit regularization parameter is set to $\lambda=10^{-4}$. To compare the predicted and measured distributions, we apply the Sinkhorn divergence with $\epsilon = 0.1$. To assess the usefulness of the CLE description, we compare the results with those obtained by using common models for the intrinsic noise: i) additive, ii) simple state-dependent ($\sqrt{\mathbf{x}}$), and iii) deterministic (Materials and Methods). \rev{The explicit forms of the diffusion tensor $\mathbf{D(x)}$ for different noise models are discussed in \app{Table}{PFI_terms}}. Fig.~\ref{Fig_CRN}A shows the errors in force fields under these noise models and for increasing reaction volumes $V$. As expected, because the system is well-mixed, at large volumes $V$ all the models become deterministic and perform equally well. However, across all reaction volumes, the CLE yields the most accurate force field estimates, outperforming both the state-dependent and additive noise models. Its performance degrades at small volumes when the CLE approximation breaks down. 

These differences in inferred dynamics between noise models are yet better quantified by examining the estimated Jacobian matrix ($\nabla \mathbf{\hat{g}(x)}$), which can be used to directly infer the gene regulatory network. In Fig.~\ref{Fig_CRN}D, we compare the average estimated Jacobian matrices for each diffusion model to the analytical Jacobian matrix (Fig.~\ref{Fig_CRN}C). Only the CLE model accurately identifies regulatory interactions, as indicated by AUC values approaching unity, based on the precision-recall curves (Fig.~\ref{Fig_CRN}E) \cite{pratapa2020benchmarking}. Conversely, non-CLE noise models yield an AUC close to $0.5$, indicating incorrect identification of interactions in the network. This result shows that while the non-CLE noise models can attain moderate errors of $\sim 10-20\%$ at intermediate volumes, they do so by learning a completely inaccurate regulatory network.

This observation should reflect a poor generalization performance of the non-CLE models. To test this, we ran the inferred process using initial conditions different from those in the training data (Materials and Methods). To quantify generalization, we calculated the Energy Distance (inset of Fig.~\ref{Fig_CRN}F) between the predicted marginals from various diffusion models and the ground-truth stochastic simulations. These results show that the inferred CLE process accurately tracks the true marginals, again outperforming the other noise models. This underscores the importance of the inferred force model's accuracy in determining generalization performance. 

Furthermore, we explored the dynamical behavior of these inferred force models by deterministically evolving and plotting the corresponding fixed points obtained for initial conditions sampled from the marginal distribution $\hat{p}_{t_0}(\mathbf{x})$. As molecular noise in the data increases, the fixed points for non-CLE force fields deviate from the ground truth, as shown in Fig.~\ref{Fig_CRN}G-I. These findings complement earlier one-dimensional studies \cite{coomer2022noise}, which showed that multiplicative noise can shift, create, or eliminate fixed points. Our results demonstrate this effect in a high-dimensional inverse setting, underscoring the importance of selecting appropriate noise priors to accurately capture the system's underlying dynamical properties.

Finally, in the scenario where the stoichiometry of the SRN is known, we can apply the PFI approach to estimate reaction rates in a high-dimensional setting. We consider a linear cyclic network consisting of \( d = 30 \) species and \( R = 30 \) reactions (Materials and Methods). In this constrained setting, the results are consistent with those observed for the mCAD network. The CLE approximation yields more accurate parameter estimates and better generalization performance compared to other noise models, as shown in \app{Fig.}{fig:SRN}.

\begin{figure*}[hbt!]
  \includegraphics[width=7.1in]{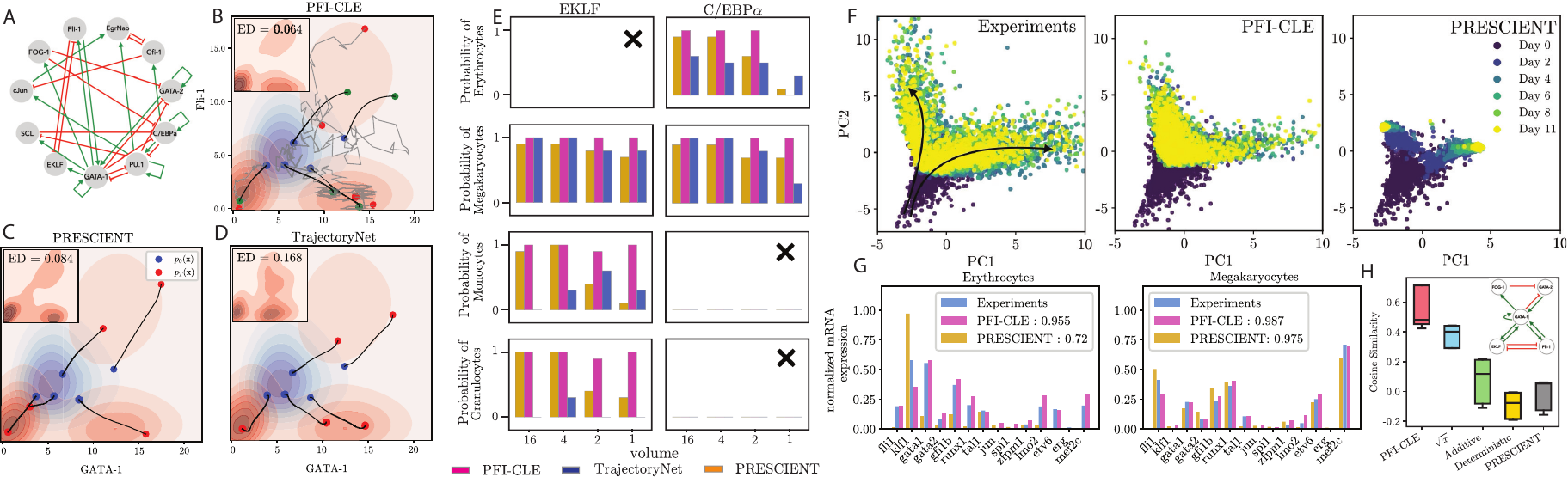}
  \caption{\textbf{Inferring HSC differentiation dynamics on simulated and experimental data.} {\bf{A.}} HSC gene regulatory network. Red lines denote inhibition and green arrows activation. 
  {\bf{B,C,D}.} Cell differentiation trajectories using different approaches (shown in title). The marginal densities at the initial state $p_0(\mathbf{x})$ (blue) and final state $p_T(\mathbf{x})$ (orange) are shown as contour plots. In the inset, we show the reconstructed marginal at time $T$ along with the reconstruction error (ED). {\bf{E.}} Probability of observing Erythrocytes, Megakaryocytes, Monocytes, and Granulocytes at steady state. Each panel displays the estimated probability of recovering the different cell types (rows) under specific perturbation conditions (columns). The averages are computed across varying initial conditions. The panels also show how these probabilities change with system size, $V$. Panels marked with a cross indicate the absence of the respective cell type under the corresponding perturbations. The results are color-coded by the approach used: TrajectoryNet (blue), PRESCIENT (gold), and Inferred Chemical Langevin model (purple). \rev{{\bf{F.}} (left panel) PCA projection of mRNA counts of 14 key TFs associated with \textit{ex vivo} hematopoiesis measured on days $\{0,2,4,6,8,11\}$. The two emergent branches (highlighted with black solid lines) lead to Erythrocytes and Megakaryocytes, the terminal cell types. Predicted differentiation trajectories following induction at day 0, using PFI-CLE (middle panel) and PRESCIENT (right panel). {\bf{G.}} Normalized expression levels of the key TFs from PFI-CLE and PRESCIENT compared with experimental observations on day 11 for each branch. The inset shows the cosine-similarity between the predicted and the measured gene profiles. {\bf{H.}} Cosine-similarity score between the inferred response matrix and known regulatory model (shown in inset) for different noise models, and also PRESCIENT.}}
  \label{perturbation}
\end{figure*}

\subsection*{PFI enables inference of cell differentiation dynamics with molecular noise} 

Cell differentiation dynamics arise from a complex interplay between deterministic mechanisms and stochastic fluctuations. However, many popular computational approaches for modeling differentiation are either purely deterministic or rely on oversimplified noise models. In this section, we examine the impact of biological noise on the predictive accuracy of widely used generative models of cellular differentiation and compare them with the CLE model inferred using the PFI approach (PFI-CLE). While various methods exist, we focus on two prominent techniques: the TrajectoryNet framework, based on dynamical optimal transport \cite{tong2020trajectorynet}, and the PRESCIENT technique \cite{yeo2021generative}, which models diffusion through a global potential function. TrajectoryNet fits a non-autonomous force field $\mathbf{f}(\mathbf{x}, t)$ to interpolate between the marginals data with minimal kinetic energy, while PRESCIENT models differentiation dynamics as a diffusion process, with the force field $\hat{\mathbf{f}}(\mathbf{x}) = -\nabla \hat{\phi}(\mathbf{x})$ and additive noise $\hat{\mathbf{D}}(\mathbf{x}) = \sigma^2 \mathbf{I}$.

We begin the comparison by generating longitudinal gene expression profiles using an expert-curated model of Myeloid Progenitor differentiation that incorporates realistic intrinsic stochasticity. This Hematopoietic Stem Cell (HSC) differentiation model includes 11 transcription factors and captures the differentiation of common myeloid progenitors (CMP) into Erythrocytes, Megakaryocytes, Monocytes, and Granulocytes. We simulate the HSC regulatory network (Fig.~\ref{perturbation}A) with the Gillespie algorithm (Materials and Methods) to generate the marginal data consisting of $K=8$ snapshots with $n=5000$ samples. The data is grouped into four clusters, with each cluster's average gene count profile shown in \app{Fig.}{fig:counts}, for varying reaction volumes. Each profile represents one of the four HSC differentiation cell types and aligns with experimentally measured mRNA data \cite{pronk2007elucidation, krumsiek2011hierarchical}. 

We train all three models on the marginal data with varying levels of intrinsic noise, corresponding to a reaction volume $V\in \{1,2,4,16\}$. The training procedure for all three methods is discussed in  Materials and Methods. To compare the inferred dynamics of the three models, we examine the individual cell trajectories generated by simulating the inferred diffusion processes. Starting from the same initial position in gene space (marked in blue), each approach predicts a different final state (marked in red) as illustrated in Fig.~\ref{perturbation}B-D. 
In other words, the predicted differentiated state of the same progenitor cell varies significantly between the three methods. However, despite the markedly distinct differentiation trajectories, all three methods accurately reconstruct the marginals, as shown by the low reconstruction error when measuring the distance between the predicted marginals and the data (inset Fig.~\ref{perturbation}B,C,D, and \rev{\app{Table}{leave_out_tp}}). For the PFI-CLE approach, we also illustrate the probability flow lines (black lines) to show how the deterministic phase velocity transports particles, and how it differs from the corresponding stochastic trajectories starting from the same initial condition. To further quantify the difference between the inferred process, we calculate the cosine similarity between the inferred probability flow lines and the CLE probability flow (PF) lines for each cell state. The probability flow of TrajectoryNet is taken to its non-autonomous force field $\mathbf{f}(\mathbf{x},t)$, while for PRESCIENT, it is estimated as $-\nabla \hat{\phi}(\mathbf{x}) - \sigma^2 \nabla \log p_t(\mathbf{x})$.  As intrinsic stochasticity increases, the flow lines corresponding to TrajectoryNet and PRESCIENT diverge significantly from the PFI-CLE flow lines in each cell state, as shown in Fig.~\ref{fig:HSC_add}A. These results illustrate that while all methods effectively interpolate between the marginal distributions, increasing intrinsic stochasticity in the data leads them to learn markedly different stochastic dynamics, as evidenced by the analysis of their probability flows and the observation of individual cell trajectories.

Given the distinct dynamics learned by each approach, we next explored whether good interpolation accuracy correlates with learning the correct dynamics, and how molecular noise affects this relationship. To evaluate this, we performed perturbation studies to certify the correctness of the learned dynamics under initial conditions different from the training data (Materials and Methods). Specifically, we conduct knockdown experiments by setting the expression levels of specific genes to zero. We focus on experimentally measured outcomes for knockdowns of the genes \textit{C/EBP$\alpha$} and \textit{EKLF}. Experimental studies have reported that disruption of \textit{C/EBP$\alpha$} blocks the transition from the common myeloid to the granulocyte/monocyte progenitor\cite{zhang2004enhancement}, and knock-down of the $EKLF$ gene leads to the absence of the Erythrocyte cell line \cite{nuez1995defective}. In Fig.~\ref{perturbation}E, we report the probability of observing each cell type under various perturbation conditions. To separate the effect of the force model from the influence of noise, we perform in-silico perturbation experiments across a range of noise levels. At low noise levels, both PRESCIENT and PFI-CLE, despite using different force models, accurately predict the perturbation outcomes. However, as molecular noise increases, differences in their inferred dynamics become more pronounced, with PRESCIENT being the most affected. For instance, under high noise conditions, PRESCIENT fails to recover the Monocyte/Granulocyte lineage in multiple knockdown scenarios, whereas PFI-CLE, which accounts for molecular noise, consistently reconstructs this lineage across all noise levels. In contrast, TrajectoryNet struggles to recover the Monocyte/Granulocyte lineage even at low noise, and as stochasticity increases, its ability to reconstruct the Megakaryocyte/Erythrocyte lineage also diminishes monotonically. These findings underscore the critical role of intrinsic stochasticity and illustrate how it can impede state-of-the-art methods from accurately capturing gene regulatory interactions during differentiation.

\rev{\indent Guided by insights from our simulation study of the HSC network, we apply the PFI framework to time-resolved mRNA count data collected during \textit{ex vivo} hematopoiesis \cite{georgolopoulos2021discrete}. This dataset comprises cells undergoing 12-day \textit{ex vivo} induced differentiation from CD34$^+$ hematopoietic stem and progenitor cells (HSPC) towards erythrocytes and megakaryocytes. Single-cell RNA-seq measurements are collected on days $2, 4, 6, 8,$ and $11$ following induction. In Figure~\ref{perturbation}F, the PCA projection of mRNA counts for 14 key transcription factors (TFs) reveals two branching trajectories over time, corresponding to erythropoiesis and megakaryopoiesis. Notably, the key regulatory genes identified in this experimental study overlap with those previously considered in the HSC differentiation (Fig.~\ref{perturbation}A), reinforcing the observation that hematopoietic differentiation is driven by a small subset of transcription factors among the approximately $10,000$ genes profiled.}

\rev{Using our two-step PFI approach, we directly fit the CLE given in Eq.~(\ref{eq:CLE}) to the measured mRNA count data.
The average degradation rate is set to $\ell = 1 \text{day}^{-1}$ consistent with typical measurements in mammalian cells \cite{shamir2016snapshot}. To assess the validity of the PFI-CLE formulation, we also infer force fields under other commonly used noise models. Figure~\ref{perturbation}F shows that, starting from day 0 initial conditions, both the PFI-CLE (middle panel) and the PRESCIENT method (right panel) effectively recover the branching trajectories associated with erythropoiesis and megakaryopoiesis. To quantify agreement with experimental data, we calculate the cosine similarity between predicted and measured normalized gene expression on day 11 along both branches (Fig.~\ref{perturbation}G), showing that both methods accurately recover the expected terminal cell types.}

\rev{To go further, we identify the causal relationships between the genes through an \textit{in silico} perturbation of the inferred force field $\mathbf{f(x)}$ (Materials and Methods). 
As highlighted in Fig.~\ref{perturbation}H, the PFI-CLE produced the highest similarity score when evaluated against known regulatory interactions involving the key genes driving erythrocyte-megakaryocyte progenitors to their respective terminal states. The performance degrades successively from the state-dependent $\sqrt{\mathbf{x}}$ model, to the additive noise model, and finally the deterministic model. This trend strikingly mirrors our earlier findings on the simulated mCAD network, where CLE accurately identifies regulatory interactions (Fig.~\ref{Fig_CRN}D,E), and non-CLE models lead to the identification of spurious interactions between the genes. Interestingly, even though PRESCIENT captures the branching differentiation (Fig.~\ref{perturbation}F, right panel), it does so by learning incorrect regulations between the genes as quantified in Fig.~\ref{perturbation}H. To evaluate the robustness of these findings, we also inferred force fields across different gene sets (\app{Table}{table:geneset}) and observed that the same trend consistently held across settings with varying dimensionality $d$. Consistent with our analysis on the simulated HSC network, the PFI-CLE formulation, by accounting for intrinsic stochasticity, enables a clearer separation of noise and regulatory signals and accurately captures the regulatory interactions among key genes.
}

\section*{Discussion}
In this paper, we propose PFI, an approach that transforms the problem of learning SDEs into inferring their corresponding phase-space probability flow. This is facilitated by recent advances in score-based generative modeling, which allow for efficient computation of the gradient log-probability, $\nabla \log p_t(\mathbf{x})$, from high-dimensional cross-sectional data of the time-evolving distribution $p_t(\mathbf{x})$. This reformulation dramatically simplifies optimization and, crucially, separates the impact of intrinsic noise from inferring the force field.

We first limited our analysis of the PFI problem to the analytically tractable class of Ornstein-Uhlenbeck processes. We proved that the regularization ensures a strongly convex loss with a unique global minimum in the limit of well-sampled distributions, both in time and in state-space. For an isotropic Ornstein-Uhlenbeck process, we minimized analytically this loss function, showing that the relative magnitudes of the regularization and of the time-dependent deformations of the marginals select the learnable non-equilibrium contributions to the force. While large deformations are used to learn non-conservative forces, smaller deformations are washed out by the regularization. Although this observation suggests to use a smaller regularization, we showed that the bias reduction obtained thereby trades-off with an increasing variance stemming from the finite sampling of distributions, both in time $\Delta t$ and in the number of samples $n$. This interplay between $n$, $\Delta t$, and the regularization calls for careful model selection when considering complex models.

Using the same analytical solution, we subsequently showed that an inaccurate estimation of the noise strength leads to a dramatic decrease in performance when the stochastic effects are of the same order of magnitude as the force. This is of paramount importance for the inference of gene regulatory networks for which intrinsic noise is strongly correlated with the amplitude of regulation. To further explore this role of the noise model, we numerically investigated more realistic models of stochastic reaction networks, using the PFI approach. Our study underscores the critical role of intrinsic noise in parameter estimation, regulatory network inference, and generalization to unseen data. An incorrect noise model can lead to spurious relationships between species, which affects the inferred probabilistic landscape \cite{coomer2022noise}. Consequently, force fields based on such models exhibit poor generalization performance when tested with initial conditions different from those used during training. The PFI approach, therefore, proves to be a valuable tool for analyzing single-cell omics data, even when the stoichiometry of the reaction network is unknown.

Finally, we applied our framework to learn data-driven models of cell differentiation. To account for molecular noise, we inferred a Chemical Langevin model using the PFI approach, and compared it with popular generative models in predicting the effects of gene knockdowns in the hematopoiesis system for increased noise strength. While PRESCIENT successfully predicts the effects of interventions when noise is negligible, it struggles when it becomes significant. On the other hand, the potential-based model TrajectoryNet fails to accurately capture regulatory interactions, particularly in high-dimensional settings, even with minimal molecular noise. In contrast, PFI-CLE consistently makes accurate predictions, emphasizing the importance of modeling intrinsic noise to infer accurate regulatory pathways. \rev{Applied to time-resolved mRNA count data collected during \textit{ex vivo} human hematopoiesis, the PFI-CLE model yields stable temporal dynamics and captures the branching trajectories corresponding to erythropoiesis and megakaryopoiesis. In addition, the PFI-CLE model interpolates the data by learning the regulations between genes. To our knowledge, this is the first demonstration on time-resolved single-cell RNAseq data of the necessity to account for intrinsic noise to predict accurate regulatory interactions.} Indeed, our results show that accurately interpolating the data is not \rev{sufficient}, and the interpolation needs to be guided by comprehensive biophysical priors of the latent stochastic processes, and special care should be given to modeling intrinsic noise. For this purpose, the PFI approach provides a very flexible solution to incorporate more realistic noise models in the inference of regulatory processes from single-cell omics data.  In this direction, future work should aim at extending the PFI approach to account for the unobserved protein dynamics between protein production and its regulatory effects on transcription \cite{josic2011stochastic, bratsun2005delay-induced,herbach2017inferring, herbach2021gene, ventre2023one}.

In this study, we primarily focus on intrinsic stochasticity in the form of molecular noise. However, extending the PFI approach to include other sources of stochasticity, like fluctuations in transcriptional rates \cite{ham2020extrinsic, gorin2022interpretable, herbach2017inferring}, would be a natural step forward. More importantly, future studies should build upon the PFI framework by integrating cell death and proliferation. We believe this is a necessary step to take to successfully and reliably apply the PFI approach to real data. Given the prevalence of noise in cellular processes, our approach marks a significant step toward developing biophysically accurate, data-driven models that incorporate non-trivial stochastic effects.

\section{Materials and Methods}
\subsection{Choice of matrices for the Ornstein-Uhlenbeck process}
To generate a random skew-symmetric matrix we first generate a matrix $\mathbf{U} \in \mathbb{R}^{d\times d}$ with i.i.d. entries drawn uniformly at random in $[0,1]$, and $\mathbf{e} \in \mathbb{R}^d$ a vector with i.i.d. entries drawn uniformly at random in $[0.9, 1]$. Denoting $\mathbf{W} \in \mathbb{C}^{d\times d}$ the eigenbasis of $\mathbf{U} - \mathbf{U}^T$, the matrix $\mathbf{A}$ is set to be $\mathbf{A} = \mathbf{W} i\mathbf{e} \mathbf{W}^{\dagger} / \mathrm{max}(\mathbf{e})$ where $i$ is the imaginary number and $\dagger$ denotes the hermitian conjugate. The qualitative behavior shown in Figs.~\ref{fig_ou_process} and \ref{fig_ou_error} is unchanged by repeatedly drawing new instances of $\mathbf{A}$.  In Figs.~\ref{fig_ou_process} and \ref{fig_ou_error} we used the same instance of the matrix $\mathbf{A}$. To generate a random symmetric matrix $\mathbf{S}$ we use a similar approach, with the matrix $\mathbf{W}$ being the eigenbasis of $\mathbf{U}+ \mathbf{U}^T$.

\subsection{Modeling gene regulation}\label{MM:celldiff}
Under the assumption that the system is well mixed, the regulation of gene $i$ by a set $R_i$ of regulators obeys the following set of stochastic reactions
\begin{equation}
X_i \xrightarrow{ m V \cdot g(R_i)} X_i + 1, \quad X_i \xrightarrow{\ell \cdot X_i} X_i - 1 \label{gill_mcad}
\end{equation}
Here, the term \(m V g(R_i)\) represents the propensity associated with the reaction that produces the mRNA for gene \(x_i\), where \(m\) denotes the transcription rate, and $R_i$ denotes the set of regulators of node $i$. The term \(\ell X_i\) corresponds to the propensity of the degradation reaction, with \(\ell\) being the degradation rate. These propensities denote the number of these reactions happening per unit of time. The non-linear function \(g(R_i)\) encapsulates the regulatory interactions governing the expression of gene \(i\), and is mediated by proteins. Given the volume of the reaction compartment, the functional form of $R_i$ as a function of the concentration of proteins is derived from equilibrium statistical mechanics considerations \cite{marbach2010revealing}. In this work, we adopt its Boolean network implementation as introduced in \cite{pratapa2020benchmarking}.

In the absence of joint measurements of mRNA molecules and proteins, we assume that protein and mRNA levels are strongly correlated, allowing us to replace the protein regulators $R_i$ with their mRNA counterparts. Although this assumption suggests that proteins are in quasi-steady-state with mRNA, which contrasts with observations \cite{abreu2009global,shahrezaei2008analytical} of faster mRNA turnover compared to proteins \cite{herbach2017inferring}, it still enables the recovery of single-cell RNA-seq profiles that are quantitatively comparable at steady state \cite{dibaeinia2020sergio}. In practice, most gene regulatory network inference approaches ignore protein dynamics, implicitly making a similar assumption.\\

\noindent \paragraph*{Simulation of linear stochastic reaction networks:} For the linear cyclic network $X_i \xrightarrow{k_i} X_{i+1}$, rate constants $k_i$ were generated using a logarithmic scale spanning from \(10^{0.1}\) to \(10^{1.5}\), distributed across $d=30$ values, and scaled by a factor of $0.1$. Gillespie simulations were then computed using the stoichiometric matrix and the rate constants.

\subsection{Comparison with existing methods} \label{MM:compare}
In the following, we describe the existing methods used to infer cell differentiation dynamics from empirical marginal distributions. We consider that $K+1$ distributions are observed at times $t_0 \leq ... \leq t_{K+1}$, each with $n$ samples. The PFI-CLE model is trained using the two-step PFI approach, with the force function \(\mathbf{f}_\theta: \mathbb{R}^d \rightarrow \mathbb{R}^d\) represented by a feed-forward neural network. The network consists of four fully connected layers, each with 30 nodes and smooth ELU activations.\\
\paragraph*{PRESCIENT:} Following the original idea by Hashimoto \cite{hashimoto2016learning}, popular generative models model cellular differentiation as a diffusion process \( \mathbf{x}(t) \) \cite{yeo2021generative, sisan2012predicting}, governed by the stochastic differential equation:
\begin{equation}
    d\mathbf{x}(t) = (-\nabla \Psi(\mathbf{x}) - \ell \mathbf{x}) dt + \sqrt{2\kappa^2} d\mathbf{W}(t), \label{eq:sde_prescient}
\end{equation}
where the drift is the gradient of a potential function $\Psi(\mathbf{x}):\mathbb{R}^d \rightarrow \mathbb{R}$. PRESCIENT proceeds by finding the function $\Psi$ that minimizes the loss function
\begin{equation}
    \mathcal{L}_{\mathrm{PRESCIENT}} = \sum_{i=1}^{K} \left[ \mathcal{W}_2^2 \left( \hat{p}_{t_i}(\mathbf{x}), p_{t_i}(\mathbf{x}) \right)^2 + \tau \sum_{j=1}^{n} \frac{\Psi (\mathbf{x}_j)}{\kappa^2} \right].
\end{equation}
The Wasserstein distance measures the difference between the observed distribution \( p_{t_i}(\mathbf{x}) \) and the predicted distribution \( \hat{p}_{t_i}(\mathbf{x}) \), and the parameter \( \tau \) controls the entropic regularization. The probability distributions are fitted to the observed data by simulating the stochastic differential equation Eq.~(\ref{eq:sde_prescient}). The potential function $\Psi(\mathbf{x})$ is parameterized with a fully connected neural network with ELU activation function
In the original study, the noise scale is set to be $\kappa = 0.1$ and hyperparameter $\tau = 10^{-6}$, and we use the same parameters in the training of the PRESCIENT model. The potential function $\Psi(\mathbf{x})$ is parameterized with a feed-forward neural network consisting of four fully connected layers, each with $30$ nodes and smooth ELU activation.\\

\paragraph*{TrajectoryNet:} Another popular approach for modeling cellular differentiation involves parameterizing the force field as a non-autonomous Neural ODE \cite{tong2020trajectorynet, sha2024reconstructing}, without any explicit noise model:
\begin{equation}
\frac{d \mathbf{x}}{dt} = \hat{\mathbf{f}}(\mathbf{x}, t).
\end{equation}
Using this framework, TrajectoryNet fits a continuous normalizing flow connecting the successive distributions, enforcing an analytically tractable reference distribution $p_{t_{-1}} \sim \mathcal{N}(0,1)$ at time $t_{-1}$. The loss function reads
\begin{equation}
\mathcal{L}_{\mathrm{TrajectoryNet}} = -\sum_{i = 0}^{K} \mathbb{E}_{p_{t_i}} \log \hat{p}_{t_i}(\mathbf{x}) + \mathrm{regularization}.
\end{equation}
The first term corresponds to the log-likelihood of the predicted distributions $\hat{p}_{t_i}(\mathbf{x})$ evaluated on the observed data  $p_{t_i}(\mathbf{x})$. TrajectoryNet includes different choices for the regularization on the force field, among which are the penalization of the curvature of the force field (used in PFI) and of its kinetic energy \cite{tong2020trajectorynet}. To train TrajectoryNet, we use the default parameters with the OT-inspired regularization on the kinetic energy suggested in the original study. The non-autonomous force field is modeled using a neural network with three fully connected layers, each containing 64 nodes and employing leaky ReLU activations.

\subsection{In-silico perturbations}\label{MM:insilico}
We conduct in-silico perturbations by simulating the inferred models from a given initial condition until a steady state is reached, with the concentration of the knocked-down gene set to zero.  The models are tested with varying initial conditions derived from the training data, given by \( p = (1 - c)p^* + c \, \textrm{U}[0.25, 0.5] \), to evaluate generalization. In this setup, \(c = 0\) corresponds to the first marginal of the training data $p^* = \hat{p}_{0}(\mathbf{x})$, while \(c = 1\) represents a uniform initialization of mRNA expression within the hypercube \([0.25, 0.5]^d\). We generate $10$ distinct initial conditions by selecting $c$ from a uniformly spaced grid between $0$ and $1$, with increments of $0.1$. Cells are assigned to a specific cell state if the cosine similarity between the cell state's expression profile and the predicted expression vector is $0.95$ or higher. 

\rev{Alternatively, to directly assess causal relationships between genes, we perform perturbation analysis by measuring the change in \( f_j \) resulting from a fold change in \( x_i \) \cite{shen2021finding}:
\begin{equation}
\Delta_{ij} \equiv f_j(\cdots, x_i) - f_j(\cdots, \xi x_i), \quad 0 < \xi < 1.
\end{equation}
This approach mimics biological knockdown experiments, with the parameter \( \xi \) controlling the perturbation strength. For example, \( \xi = 0 \) corresponds to a full knockout of the input, while \( \xi \approx 1 \) approximates a knockdown derivative. To quantify the regulatory influence of \( x_i \) on \( x_j \), we compute the average response matrix \( \langle \Delta_{ij} \rangle \) across all time points and samples, sweeping over perturbation strengths in the range \( 0.7 < \xi < 0.95 \). We quantify the agreement between the inferred response matrix \( \Delta \) and the known regulatory matrix \( M \) using the cosine similarity metric. The entries of \( M \) take values in \(\{1, -1, 0\}\), corresponding to activation, inhibition, and no interaction, respectively. The similarity score is defined as:
\begin{equation}
    \text{Cosine similarity} = \frac{\sum_{i,j} \Delta_{ij} M_{ij}}{\sqrt{\sum_{i,j} \Delta_{ij}^2} \, \sqrt{\sum_{i,j} M_{ij}^2}}.
\end{equation}
For hematopoiesis, the matrix $M$ is constructed based on the boolean network model discussed in \cite{krumsiek2011hierarchical}.
}

\bibliographystyle{arxiv}
\bibliography{references}

\appendix
\onecolumngrid

\newpage

\renewcommand{\thefigure}{S\arabic{figure}}
\setcounter{figure}{0}
\renewcommand{\thetable}{S\arabic{table}}
\setcounter{table}{0}

{\rev{\section{Solutions to the divergence equation and role of regularization}\label{sec:recovery}
We denote $S = \{ \mathbf{x} \text{ s.t. }p_t(\mathbf{x}) > 0 \text{ for } t \in [0,T]\}$. Any force field of the form $\hat{\mathbf{f}}(\mathbf{x}) = \mathbf{f}(\mathbf{x}) + \mathbf{h}(\mathbf{x})$ generates the same marginals distributions as $\mathbf{f}(\mathbf{x})$ if $\mathbf{h}(\mathbf{x})$ satisfies
\begin{equation}
\nabla \cdot (\mathbf{h}(\mathbf{x}) p_t(\mathbf{x})) = 0, \forall t \in [0,T], \;  \forall \mathbf{x} \in \mathbb{R}^d\label{div}
\end{equation}
We denote by $\mathbf{h}(\mathbf{x}) = \nabla \psi(\mathbf{x})$ a gradient solution to Eq.~(\ref{div}). With differential calculus we have the identity $\nabla \cdot (  \psi (\mathbf{x}) p_t(\mathbf{x})\nabla \psi(\mathbf{x})) = \psi (\mathbf{x}) \nabla \cdot (p_t(\mathbf{x}) \nabla\psi (\mathbf{x})) +  p_t (\mathbf{x})| \nabla \psi (\mathbf{x})|^2$. Because $\nabla \psi(\mathbf{x})$ satisfies Eq.~(\ref{div}), we find $\nabla \cdot (  \psi (\mathbf{x}) p_t(\mathbf{x})\nabla \psi(\mathbf{x})) = p_t (\mathbf{x})| \nabla \psi (\mathbf{x})|^2$. Using the divergence theorem, and with the assumption that $p_t(\mathbf{x})$ decays sufficiently fast at infinity, the integral of $p_t (\mathbf{x})| \nabla \psi (\mathbf{x})|^2$ vanishes. As a result, the only gradient solution to Eq.~(\ref{div}) on $\mathcal{S}$ is the trivial solution $\nabla \psi = 0$.} 

We can now argue that, if there exists a gradient force $\nabla \phi(\mathbf{x})$ which conserves the evolution of $p_t(\mathbf{x})$, we uniquely recover it on $\mathcal{S}$ in the limit $\lambda \rightarrow 0$.  We can make this argument rigorous in the case of the regularization on the kinetic energy. Let $\hat{\mathbf{f}}(\mathbf{x})$ be the force of the inferred process. The loss then reads:
\begin{equation}
\mathcal{L}_{\Delta t, K} =  \sum_{i=1}^{K} \Bigg[\mathcal{W}_2^2 (\hat{p}_{t_i}( \mathbf{x}), p_{t_i}( \mathbf{x}) ) \notag + \lambda \Delta t \int_{t_{i-1}}^{t_i}  \int \!\!\|  \hat{\mathbf{f}} (\mathbf{x}) \|^2 {p}_{t}(\mathbf{x}) d\mathbf{x} dt \Bigg]. 
\end{equation}
Taking $\Delta t \rightarrow 0$ with $K = \lfloor T / \Delta t \rfloor$ and $T$ fixed, we have that
\begin{equation}
\lim\limits_{\Delta t \rightarrow 0}\frac{\mathcal{L}_{\Delta t, K}}{\Delta t} = \mathcal{L} = \lim_{\Delta t \rightarrow 0} \frac{1}{\Delta t} \sum_{i=1}^{K} \mathcal{W}_2^2 (\hat{p}_{t_i}( \mathbf{x}),  p_{t_i}( \mathbf{x}) ) + \lambda \int_0^T \int\!\!\|  \hat{\mathbf{f}} (\mathbf{x}) \|^2 p_t(\mathbf{x}) d\mathbf{x} dt. 
\end{equation}
The first term of the loss compares the distributions $\hat{p}_t$ and $p_t$ at all times, and is zero when they are equal. \rev{We denote by $\mathcal{F}$ the set of functions $\hat{\mathbf{f}}(\mathbf{x}) = \nabla \phi (\mathbf{x}) + \mathbf{h}(\mathbf{x})$ where $\mathbf{h}(\mathbf{x})$ satisfies Eq.~(\ref{div}). Any force in $\mathcal{F}$ conserves the evolution of marginals $p_t(\mathbf{x})$, and the Wasserstein term in the loss is zero. As a result, we have
\begin{equation}
\min_{\hat{\mathbf{f}} \in \mathcal{F}} \mathcal{L}  = \min_{\hat{\mathbf{f}} \in \mathcal{F}} \int_0^T \int\!\!\|  \hat{\mathbf{f}} (\mathbf{x}) \|^2 p_t(\mathbf{x}) d\mathbf{x} dt.
\end{equation}
Expanding the square, we have
\begin{equation}
\int_0^T \int\!\!\|  \hat{\mathbf{f}} (\mathbf{x}) \|^2 p_t(\mathbf{x}) d\mathbf{x} dt = \int_0^T \int \left[ \|\nabla \phi(\mathbf{x})\|^2 + 2 \mathbf{h}(\mathbf{x}) \cdot \nabla \phi(\mathbf{x}) + \|\mathbf{h}(\mathbf{x})\|^2\right] p_t(\mathbf{x}) d\mathbf{x} dt.
\end{equation}
Since $\nabla \cdot (\mathbf{h}(\mathbf{x}) \phi(\mathbf{x}) p_t(\mathbf{x})) =\phi(\mathbf{x}) \nabla \cdot (\mathbf{h}(\mathbf{x})p_t( \mathbf{x})) + p_t( \mathbf{x}) \mathbf{h}(\mathbf{x}) \cdot \nabla \phi(\mathbf{x}) $, using the divergence theorem and with the assumption that $p_t$ decreases sufficiently fast at infinity, we find
\begin{equation}
\int_0^T \int  \mathbf{h}(\mathbf{x}) \cdot \nabla \phi(\mathbf{x}) p_t(\mathbf{x})d \mathbf{x} dt = -\int_0^T \int \phi(\mathbf{x}) \nabla \cdot (\mathbf{h}(\mathbf{x}) p_t(\mathbf{x}))d\mathbf{x} dt = 0.
\end{equation}
Therefore, 
\begin{equation}
\min_{\hat{\mathbf{f}} \in \mathcal{F}} \mathcal{L}  = \min_{\mathbf{h}} \int_0^T \int \left[ \|\nabla \phi(\mathbf{x})\|^2 + \|\mathbf{h}(\mathbf{x})\|^2\right] p_t(\mathbf{x}) d\mathbf{x} dt.
\end{equation}
This implies that within the function class \( \mathcal{F} \), the loss attains a unique minimum at \( \mathbf{h} = 0 \), corresponding to the gradient solution \( \hat{\mathbf{f}}(\mathbf{x}) = \nabla \phi(\mathbf{x}) \). However, when \( \lambda > 0 \), the minimum may lie outside the set \( \mathcal{F} \) due to regularization. As \( \lambda \to 0 \), the contribution from the regularization term vanishes, and the solution converges to the unique gradient field \( \hat{\mathbf{f}}(\mathbf{x}) = \nabla \phi(\mathbf{x}) \).}

\section{Analytical results for the Ornstein-Uhlenbeck process}
\label{sec:SM_ouprocess}
\subsection{Notation and preliminaries}\label{sec:SM_notations}
We focus on the $d$-dimensional Ornstein-Uhlenbeck process. The true process reads as follows:
\begin{equation}
d\mathbf{x} = \mathbf{\Omega} \mathbf{x} dt + \sqrt{2\mathbf{D}} d\mathbf{W}, \: \text{with } \mathbf{x}_0 \sim p_0(\mathbf{x}) .
\end{equation}
When $p_0 = \mathcal{N}(\mathbf{m}_0, \mathbf{\Sigma}_0)$, the solution of this stochastic differential equation is Gaussian at all times with
\begin{gather}
\mathbf{x}_t \sim \mathcal{N} ( \mathbf{m}_t, \mathbf{\Sigma}_t ), \\
\text{where } \mathbf{m}_t = e^{\mathbf{\Omega} t} \mathbf{m}_0, \: \mathbf{\Sigma}_t = e^{\mathbf{\Omega} t} \mathbf{\Sigma}_0 e^{\mathbf{\Omega}^T t} + \int_0^t e^{\mathbf{\Omega} (t - s)} 2 \mathbf{D} e^{\mathbf{\Omega}^T (t - s)} ds . \label{cov_gen}
\end{gather}
We require that the eigenvalues of $\mathbf{\Omega}$ have a strictly negative real part. We can see that if $\mathbf{\Omega}$ is a normal matrix (i.e. it commutes with its transpose), the diffusion matrix $\mathbf{D}$ and the initial condition $\mathbf{\Sigma}_0$ are isotropic (proportional to identity) and the initial condition $\mathbf{m}_0 = 0$, then the evolution of the marginals of the Ornstein-Uhlenbeck process is independent of the skew-symmetric part of $\mathbf{\Omega}$ with
\begin{equation}
\mathbf{m}_t = 0, \: \mathbf{\Sigma}_t = \Sigma_0 e^{(\mathbf{\Omega} + \mathbf{\Omega}^T) t} + 2D \int_0^t e^{(\mathbf{\Omega} + \mathbf{\Omega}^T)(t-s)} ds. 
\end{equation}
In that case, since neither the first or second moment depend on the skew-symmetric part, it is impossible to infer it by only exploiting time deformations of the marginal distributions. We see here that in such a scenario the marginal distributions are not at steady-state but the true skew-symmetric part of the interaction matrix can't be uniquely identified. However, as soon as $\mathbf{m}_0 \neq 0$ this degeneracy is lifted and the skew-symmetric terms can be identified. To illustrate this we further simplify the true process by taking the symmetric part of $\mathbf{\Omega}$ also isotropic. With $\mathbf{\Sigma}_0 = \Sigma_0\mathbf{I}$, isotropic diffusion $D$ and interaction matrix $\mathbf{\Omega} = \Omega_s \mathbf{I} + \mathbf{\Omega}_a$ with $\mathbf{\Omega}_a$ skew-symmetric, the covariance matrix at time $t$ is always isotropic and reads
\begin{gather}
\Sigma_{t} = \Sigma_0 e^{2\Omega_s t} + D \frac{e^{2 \Omega_s t}}{\Omega_s} \left[ 1 - e^{-2\Omega_s t} \right] .\label{cov_iso}
\end{gather}
The distribution associated with the true process is denoted $p_t(\mathbf{x})$. \rev{The inferred process is the solution of the probability flow ODE on the interval $]t_k, t_{k+1}]$ with initial condition at time $t_k$ drawn from the true distribution $p_{t_k}(\mathbf{x})$. This equation reads
\begin{equation}
\frac{d \hat{\mathbf{x}}}{dt} = \hat{\mathbf{\Omega}} \hat{\mathbf{x}} + \hat{\mathbf{D}} \mathbf{\Sigma}_t^{-1} \mathbf{x}.
\end{equation}
In practice, given a set of samples observed from the true process $\{\mathbf{x}_{i,t_k}, 1 \leq i < n_k, 0 \leq k \leq K\}$, we only perform one integration step to generate the inferred samples, and the update equation reads
\begin{equation}
\hat{\mathbf{x}}_{i, \Delta t} = \left(\mathbf{I} + \Delta t \hat{\mathbf{\Omega}} \right) \mathbf{x}_{i,0} + \Delta t\hat{\mathbf{D}} \mathbf{\Sigma_0}^{-1} \left(\mathbf{x}_{i,0} - \mathbf{m}_0 \right). \label{discrete_PF}
\end{equation}
We will investigate the limit $\Delta t \rightarrow 0$. Whenever it is mentioned, this limit needs to be interpreted in terms of the non-dimensional number $\Omega_{\mathrm{max}} \Delta t$ where $\Omega_{\mathrm{max}}$ is the largest mode of $\mathbf{\Omega}$.}

\subsection{Proof of Theorem 1}\label{sec:SM_loss}
\begin{theorem*} \label{theo_loss}
With $K = \lfloor T/ \Delta t \rfloor$, when $n\rightarrow \infty$ and $\Delta t \rightarrow 0$, the loss function $\mathcal{L}_{\Delta t, K} /\Delta t \rightarrow$  $\mathcal{L}$ with
\begin{align} \label{wass_loss}
\mathcal{L} &= \mathrm{tr} \left( (\hat{\mathbf{\Omega}} - \mathbf{\Omega})^T(\hat{\mathbf{\Omega}} - \mathbf{\Omega} ) \mathbf{P} + \lambda T \hat{\mathbf{\Omega}} \hat{\mathbf{\Omega}}^T\right) \\
&+ \int_0^T  \sum_{i,p} \frac{\sigma_{i,t}^2}{\left( \sigma_{i,t}^2 + \sigma_{p,t}^2\right)^2} \bigg(\mathbf{w}_{i,t}^T \bigg(\sigma_{p,t}^2 (\hat{\mathbf{\Omega}} - \mathbf{\Omega})\notag + \sigma_{i,t}^2 (\hat{\mathbf{\Omega}}^T - \mathbf{\Omega}^T) + 2(\hat{\mathbf{D}} - \mathbf{D})\bigg) \mathbf{w}_{p,t} \bigg)^2 dt, 
\end{align}
where 
\begin{equation}
\mathbf{P} = \int_0^T \mathbf{m}_t \mathbf{m}_t^T dt.
\end{equation}
For $\lambda > 0$ this loss function is strongly convex, so it has a unique minimum on $\mathbb{R}^d$.
\end{theorem*}

\begin{proof}
We consider that $n \rightarrow \infty$ such that $\hat{p}_{t_i}$ and $p_{t_i}$ are Gaussian distributions. The loss function reads as follows
\begin{gather}
\mathcal{L}_{\Delta t, K} = \sum_{i = 1}^{K} \mathcal{W}_{2}^2 (\hat{p}_{t_i}, p_{t_i}) + \lambda K \Delta t^2 \| \hat{\mathbf{\Omega}}\|_F^2 = \sum_{i = 0}^{K-1} \mathcal{W}_{2}^2 (\hat{p}_{t_i + \Delta t}, p_{t_i + \Delta t}) + \lambda K \Delta t^2 \| \hat{\mathbf{\Omega}}\|_F^2. \label{loss}
\end{gather}
Let's consider the first term $t_0 = 0$ in the above sum. We have $p_{\Delta t} \sim \mathcal{N}(\mathbf{m}_{\Delta t}, \mathbf{\Sigma}_{\Delta t})$ and $\hat{p}_{\Delta t} \sim \mathcal{N}(\hat{\mathbf{m}}_{\Delta t}, \hat{\mathbf{\Sigma}}_{\Delta t})$, and the Wasserstein distance between both distribution reads
\begin{equation}
\mathcal{W}_{2}^2 (\hat{p}_{\Delta t}, p_{\Delta t}) = \|\hat{\mathbf{m}}_{\Delta t} - \mathbf{m}_{\Delta t}\|^2 + \mathrm{tr} (\hat{\mathbf{\Sigma}}_{\Delta t}) + \mathrm{tr} (\mathbf{\Sigma}_{\Delta t}) - 2 \mathrm{tr} \left( \left( \mathbf{\Sigma}_{\Delta t}^{1/2} \hat{\mathbf{\Sigma}}_{\Delta t} \mathbf{\Sigma}_{\Delta t}^{1/2} \right)^{1/2} \right).
\end{equation}
We denote the small time step expansion of the covariance matrices as
\begin{gather}
\mathbf{\Sigma}_{\Delta t} = \mathbf{\Sigma}_0 + \Delta t \mathbf{B} + \Delta t^2 \mathbf{C} + o(\Delta t^2), \\
\hat{\mathbf{\Sigma}}_{\Delta t} = \mathbf{\Sigma}_0 + \Delta t \hat{\mathbf{B}} + \Delta t^2 \hat{\mathbf{C}} + o(\Delta t^2), \label{dt_cov}
\end{gather}
where at all orders the matrices are symmetric. We then search for $\mathbf{U}, \mathbf{V}$ and $\mathbf{Z}$ such that $\mathbf{\Sigma}_{\Delta t}^{1/2} = \mathbf{U} + \Delta t \mathbf{V} + \Delta t^2 \mathbf{Z} + o(\Delta t^2)$. By squaring this guess and using the uniqueness of the Taylor expansion we find the conditions
\begin{equation}
\mathbf{U} = \mathbf{\Sigma}_0^{1/2}, \mathbf{U} \mathbf{V} + \mathbf{V} \mathbf{U} = \mathbf{B}, \mathbf{U} \mathbf{Z} + \mathbf{Z} \mathbf{U} + \mathbf{V}^2 = \mathbf{C}.
\end{equation}
$\mathbf{V}$ is solution of a Lyapunov equation, for which the solution can be written as an integral. Denoting $\mathbf{\Sigma}_0^{1/2} = \sum_i \sigma_i \mathbf{w}_i \mathbf{w}_i^T = \sum_i \sigma_i \mathbf{z}_i $ with $\mathbf{z}_i = \mathbf{w}_i \mathbf{w}_i^T$ the solution reads
\begin{align}
\mathbf{V} &= \int_0^{\infty} e^{-\mathbf{\Sigma}_0^{1/2} t} \mathbf{B} e^{-\mathbf{\Sigma}_0^{1/2} t} dt = \sum_{i,j} \mathbf{z}_i \int_0^{\infty} e^{-\sigma_i t} \mathbf{B} e^{-\sigma_j t} \mathbf{z}_j dt = \sum_{i,j} \frac{1}{\sigma_i + \sigma_j} \mathbf{z}_i \mathbf{B} \mathbf{z}_j .
\end{align}
We then have
\begin{align}
\mathbf{V}^2 &= \sum_{i,j,p} \frac{1}{(\sigma_i + \sigma_p)(\sigma_p + \sigma_j)}\mathbf{z}_i\mathbf{B} \mathbf{z}_p \mathbf{B} \mathbf{z}_j,
\end{align}
which allows us to compute
\begin{equation}
\mathbf{Z} = \sum_{i,j}  \frac{1}{\sigma_i + \sigma_j} \mathbf{z}_i \mathbf{C}\mathbf{z}_j - \sum_{i,j,p} \frac{1}{(\sigma_i + \sigma_j)(\sigma_p + \sigma_j)(\sigma_i + \sigma_p)} \mathbf{z}_i \mathbf{B} \mathbf{z}_p \mathbf{B} \mathbf{z}_j .
\end{equation}
We introduce the notation `$\mathrm{t.}$` for transpose, and we can compute the expansion in powers of $\Delta t$ of $\mathbf{\Sigma}_{\Delta t}^{1/2} \hat{\mathbf{\Sigma}}_{\Delta t} \mathbf{\Sigma}_{\Delta t}^{1/2}$
\begin{align}
\mathbf{\Sigma}_{\Delta t}^{1/2} \hat{\mathbf{\Sigma}}_{\Delta t}\mathbf{\Sigma}_{\Delta t}^{1/2} &= \mathbf{\Sigma}_0^{2} + \Delta t \mathbf{\Sigma}_0^{1/2} \hat{\mathbf{B}} \mathbf{\Sigma}_0^{1/2} + \Delta t \left(\mathbf{\Sigma}_0^{3/2} \sum_{i,j} \frac{1}{\sigma_i + \sigma_j} \mathbf{z}_i \mathbf{B}\mathbf{z}_j + \mathrm{t.} \right) \notag \\
&+ \Delta t^2 \left( \mathbf{\Sigma}_0^{3/2} \left[ \sum_{i,j}  \frac{1}{\sigma_i + \sigma_j} \mathbf{z}_i \mathbf{C}\mathbf{z}_j - \sum_{i,j,p} \frac{1}{(\sigma_i + \sigma_j)(\sigma_p + \sigma_j)(\sigma_i + \sigma_p)} \mathbf{z}_i \mathbf{B} \mathbf{z}_p \mathbf{B} \mathbf{z}_j  \right] + \mathrm{t.} \right) \notag \\
&+ \Delta t^2\left( \mathbf{\Sigma}_0^{1/2} \hat{\mathbf{B}} \sum_{i,j}  \frac{1}{\sigma_i + \sigma_j} \mathbf{z}_i\mathbf{B}\mathbf{z}_j + \mathrm{t.} \right) + \Delta t^2 \mathbf{\Sigma}_0^{1/2} \hat{\mathbf{C}} \mathbf{\Sigma}_0^{1/2} + \Delta t^2 \sum_{i,j,p,l} \frac{1}{(\sigma_p + \sigma_l)(\sigma_i + \sigma_j)} \mathbf{z}_i \mathbf{B} \mathbf{z}_j \mathbf{\Sigma}_0 \mathbf{z}_p \mathbf{B} \mathbf{z}_l.
\end{align}
Following the same approach, we now perform the expansion in powers of $\Delta t$ of $\left(\mathbf{\Sigma}_{\Delta t}^{1/2} \hat{\mathbf{\Sigma}}_{\Delta t}\mathbf{\Sigma}_{\Delta t}^{1/2}\right)^{1/2}$
\begin{align}
\left(\mathbf{\Sigma}_{\Delta t}^{1/2} \hat{\mathbf{\Sigma}}_{\Delta t}\mathbf{\Sigma}_{\Delta t}^{1/2}\right)^{1/2} &= \mathbf{\Sigma}_0 + \Delta t \sum_{i,j} \frac{\sigma_i \sigma_j}{\sigma_i^2 + \sigma_j^2} \mathbf{z}_i \hat{\mathbf{B}} \mathbf{z}_j + \Delta t \left( \sum_{i,j}  \frac{\sigma_i^3 }{(\sigma_i + \sigma_j)(\sigma_i^2 + \sigma_j^2)}\mathbf{z}_i\mathbf{B}\mathbf{z}_j + \mathrm{t.}\right) \notag \\
&+ \Delta t^2 \left( \sum_{i,j} \frac{\sigma_i^3}{(\sigma_i + \sigma_j)(\sigma_i^2 + \sigma_j^2)} \mathbf{z}_i \mathbf{C} \mathbf{z}_j + \mathrm{t.}\right) \\
&- \Delta t^2 \left( \sum_{i,j} \frac{\sigma_i^3}{(\sigma_i + \sigma_j)(\sigma_i + \sigma_p)(\sigma_p + \sigma_j)(\sigma_i^2 + \sigma_j^2)} \mathbf{z}_i \mathbf{B} \mathbf{z}_p  \mathbf{B} \mathbf{z}_j + \mathrm{t.}\right) \notag \\
&+ \Delta t^2 \sum_{i,j} \frac{\sigma_i \sigma_j}{\sigma_i^2 + \sigma_j^2} \mathbf{z}_i \hat{\mathbf{C}} \mathbf{z}_j + \Delta t^2 \left( \sum_{i,j,p} \frac{\sigma_i}{(\sigma_j + \sigma_p)(\sigma_i^2 + \sigma_j^2)} \mathbf{z}_i \hat{\mathbf{B}} \mathbf{z}_p \mathbf{B} \mathbf{z}_j  + \mathrm{t.}\right) \notag \\
&+ \Delta t^2 \sum_{i,j,p} \frac{\sigma_p^2}{(\sigma_i + \sigma_p)(\sigma_p + \sigma_i)(\sigma_i^2 + \sigma_j^2)} \mathbf{z}_i \mathbf{B} \mathbf{z}_p \mathbf{B} \mathbf{z}_j - \Delta t^2 \mathbf{E}, \label{res_1}
\end{align}
where $\mathbf{E}$ is given by
\begin{align}
\mathbf{E} &= \sum_{i,j} \frac{1}{\sigma_i^2 + \sigma_j^2} \mathbf{z}_i  \left(\sum_{p,l} \frac{\sigma_p \sigma_l}{\sigma_p^2 + \sigma_l^2} \mathbf{z}_p \hat{\mathbf{B}} \mathbf{z}_l + \left( \sum_{p,l}  \frac{\sigma_p^3 }{(\sigma_p + \sigma_l)(\sigma_p^2 + \sigma_l^2)}\mathbf{z}_p\mathbf{B}\mathbf{z}_l + \mathrm{t.}\right) \right)^2 \mathbf{z}_j \notag \\
&= \sum_{i,j,p} \frac{\sigma_i \sigma_j \sigma_p^2}{(\sigma_i^2 + \sigma_j^2)(\sigma_i^2 + \sigma_p^2)(\sigma_p^2 + \sigma_j^2)} \mathbf{z}_i \hat{\mathbf{B}} \mathbf{z}_p \hat{\mathbf{B}} \mathbf{z}_j + \sum_{i,j,p} \frac{(\sigma_i^3 + \sigma_p^3)(\sigma_p^3 + \sigma_j^3)}{(\sigma_i^2 + \sigma_j^2)(\sigma_i^2 + \sigma_p^2)(\sigma_p^2 + \sigma_j^2)(\sigma_i + \sigma_p)(\sigma_j + \sigma_p)} \mathbf{z}_i \mathbf{B} \mathbf{z}_p \mathbf{B} \mathbf{z}_j \notag \\
&+ \left( \sum_{i,j,p} \frac{\sigma_i \sigma_p (\sigma_j^3 + \sigma_p^3)}{(\sigma_i^2 + \sigma_j^2)(\sigma_i^2 + \sigma_p^2)(\sigma_p^2 + \sigma_j^2)(\sigma_p + \sigma_j)} \mathbf{z}_i \hat{\mathbf{B}} \mathbf{z}_p \mathbf{B} \mathbf{z}_j + \mathrm{t.} \right)
\end{align}
We then take the trace of Eq.~(\ref{res_1}), and using the circular property of the trace and $\sum_i \mathbf{z}_i = \mathbf{I}$ we are left with
\begin{align}
\mathrm{tr} \left(\left(\mathbf{\Sigma}_{\Delta t}^{1/2} \hat{\mathbf{\Sigma}}_{\Delta t}\mathbf{\Sigma}_{\Delta t}^{1/2}\right)^{1/2}\right) &= \mathrm{tr}(\mathbf{\Sigma}_0) + \frac{\Delta t}{2} \left( \mathrm{tr}(\mathbf{B} +\hat{\mathbf{B}}) \right)+\frac{\Delta t^2}{2} \left( \mathrm{tr}(\mathbf{C} + \hat{\mathbf{C}}) \right) \notag \\
&+ \frac{\Delta t^2}{2}  \sum_{i,p}\left( \frac{1}{(\sigma_i + \sigma_p)\sigma_i} \mathrm{tr} (\hat{\mathbf{B}} \mathbf{z}_p \mathbf{B} \mathbf{z}_i + \mathrm{t.}) +  \frac{\sigma_p^2}{\sigma_i^2(\sigma_i + \sigma_p)^2} \mathrm{tr} (\mathbf{B} \mathbf{z}_p \mathbf{B} \mathbf{z}_i) -  \frac{1}{(\sigma_i + \sigma_p)^2} \mathrm{tr} (\mathbf{B} \mathbf{z}_p \mathbf{B} \mathbf{z}_i) \right)  \notag \\
&- \Delta t^2 \mathrm{tr} (\mathbf{E}), \label{trace_exp}
\end{align}
where we also have
\begin{equation}
\mathrm{tr}(\mathbf{E}) = \frac{1}{2} \sum_{i,p} \left( \frac{(\sigma_i^3 + \sigma_p^3)^2}{\sigma_i^2(\sigma_i^2 + \sigma_p^2)^2(\sigma_i + \sigma_p)^2} \mathrm{tr}(\mathbf{B} \mathbf{z}_p \mathbf{B} \mathbf{z}_i) + \frac{\sigma_p^2}{(\sigma_i^2 + \sigma_p^2)^2} \mathrm{tr}(\hat{\mathbf{B}} \mathbf{z}_p \hat{\mathbf{B}} \mathbf{z}_i) + \frac{\sigma_p(\sigma_i^3 + \sigma_p^3)}{\sigma_i(\sigma_i^2 + \sigma_p^2)^2(\sigma_i + \sigma_p)} \mathrm{tr}(\hat{\mathbf{B}} \mathbf{z}_p \mathbf{B} \mathbf{z}_i + \mathrm{t.})  \right)
\end{equation}
For $\mathbf{M}$ and $\mathbf{N}$ two symmetric positive definite matrices, we have that $\mathbf{M} \mathbf{N}$ is similar to $\mathbf{N}^{1/2} \left(\mathbf{M} \mathbf{N} \right)\mathbf{N}^{-1/2} = \mathbf{N}^{1/2} \mathbf{M} \mathbf{N}^{1/2}$. With the same argument, $\mathbf{M} \mathbf{N}$ is similar to $\mathbf{M}^{1/2} \mathbf{N} \mathbf{M}^{1/2}$, such that $\mathbf{M}^{1/2} \mathbf{N} \mathbf{M}^{1/2}$ and $\mathbf{N}^{1/2} \mathbf{M} \mathbf{N}^{1/2}$ are similar. As a result, $\mathbf{\Sigma}_{\Delta t}^{1/2} \hat{\mathbf{\Sigma}}_{\Delta t}\mathbf{\Sigma}_{\Delta t}^{1/2}$ and $\hat{\mathbf{\Sigma}}_{\Delta t}^{1/2} \mathbf{\Sigma}_{\Delta t}\hat{\mathbf{\Sigma}}_{\Delta t}^{1/2}$ are similar, and we can write
\begin{equation}
\mathrm{tr}\left(\left(\mathbf{\Sigma}_{\Delta t}^{1/2} \hat{\mathbf{\Sigma}}_{\Delta t}\mathbf{\Sigma}_{\Delta t}^{1/2}\right)^{1/2}\right) = \mathrm{tr} \left(\left(\hat{\mathbf{\Sigma}}_{\Delta t}^{1/2} \mathbf{\Sigma}_{\Delta t}\hat{\mathbf{\Sigma}}_{\Delta t}^{1/2}\right)^{1/2}\right).
\end{equation}
At all order in $\Delta t$ this equality holds, such that the terms $\hat{\mathbf{B}}$ and $\mathbf{B}$ as well as the terms $\hat{\mathbf{C}}$ and $\mathbf{C}$ should be exchangeable in Eq.~(\ref{trace_exp}). This symmetry is clearly respected at the order $\Delta t^2$ for $\mathbf{C}$ and $\hat{\mathbf{C}}$, but it is not striking for $\hat{\mathbf{B}}$ and $\mathbf{B}$. Let's illustrate that this symmetry holds: using the circular property of the trace we see that the trace terms are symmetric in the indices $i$ and $p$, and we can reorganize the sums
\begin{align}
\sum_{i,p} \frac{\sigma_i^2 \sigma_p^4}{\sigma_i^2(\sigma_i^2 + \sigma_p^2)^2(\sigma_i + \sigma_p)^2} \mathrm{tr}(\mathbf{B} \mathbf{z}_p \mathbf{B} \mathbf{z}_i) = \sum_{i,p} \frac{\sigma_i^6}{\sigma_i^2(\sigma_i^2 + \sigma_p^2)^2(\sigma_i + \sigma_p)^2} \mathrm{tr}(\mathbf{B} \mathbf{z}_p\mathbf{B} \mathbf{z}_i),
\end{align}
which allows us to find that
\begin{align}
\sum_{i,p} \left( \frac{\sigma_p^2}{\sigma_i^2(\sigma_i + \sigma_p)^2} - \frac{1}{(\sigma_i + \sigma_p)^2} -\frac{(\sigma_i^3 + \sigma_p^3)^2}{\sigma_i^2(\sigma_i^2 + \sigma_p^2)^2(\sigma_i + \sigma_p)^2} \right) \mathrm{tr}(\mathbf{B} \mathbf{z}_p \mathbf{B} \mathbf{z}_i) = - \sum_{i,p} \frac{\sigma_p^2}{(\sigma_i^2 + \sigma_p^2)^2} \mathrm{tr}(\mathbf{B} \mathbf{z}_p \mathbf{B} \mathbf{z}_i).
\end{align}
Finally, using the following identity
\begin{align}
\sum_{i,p} \frac{\sigma_p\sigma_i^3}{\sigma_i(\sigma_i^2 + \sigma_p^2)^2(\sigma_i + \sigma_p)} \mathrm{tr}(\hat{\mathbf{B}} \mathbf{z}_p \mathbf{B} \mathbf{z}_i + \mathrm{t.}) = \sum_{i,p} \frac{\sigma_p^2\sigma_i^2}{\sigma_i(\sigma_i^2 + \sigma_p^2)^2(\sigma_i + \sigma_p)} \mathrm{tr}(\hat{\mathbf{B}} \mathbf{z}_p \mathbf{B} \mathbf{z}_i + \mathrm{t.}),
\end{align}
we can rewrite Eq.~(\ref{trace_exp}) as
\begin{align}
\mathrm{tr} \left(\left(\mathbf{\Sigma}_{\Delta t}^{1/2} \hat{\mathbf{\Sigma}}_{\Delta t}\mathbf{\Sigma}_{\Delta t}^{1/2}\right)^{1/2}\right) &= \mathrm{tr}(\mathbf{\Sigma}_0) + \frac{\Delta t}{2} \left( \mathrm{tr}(\mathbf{B} +\hat{\mathbf{B}}) \right)+\frac{\Delta t^2}{2} \left( \mathrm{tr}(\mathbf{C} + \hat{\mathbf{C}}) \right) \notag \\
&+ \frac{\Delta t^2}{2} \sum_{i,p}\left( \frac{\sigma_i}{(\sigma_i + \sigma_p)(\sigma_i^2 + \sigma_p^2)} \mathrm{tr} (\hat{\mathbf{B}} \mathbf{z}_p \mathbf{B} \mathbf{z}_i + \mathrm{t.}) -  \frac{\sigma_p^2}{(\sigma_i^2 + \sigma_p^2)^2} \mathrm{tr} (\mathbf{B} \mathbf{z}_p \mathbf{B} \mathbf{z}_i + \hat{\mathbf{B}}\mathbf{z}_p \hat{\mathbf{B}} \mathbf{z}_i) \right).
\end{align}
With this equality we see that the symmetry for the role of $\hat{\mathbf{B}}$ and $\mathbf{B}$ is also respected. We can perform one last simplification since we have
\begin{equation}
\sum_{i,p} \frac{\sigma_i}{(\sigma_i + \sigma_p)(\sigma_i^2 + \sigma_p^2)} \mathrm{tr} (\hat{\mathbf{B}} \mathbf{z}_p \mathbf{B} \mathbf{z}_i + \mathrm{t.}) = \sum_{i,p} \frac{\sigma_p^2}{(\sigma_i^2 + \sigma_p^2)^2} \mathrm{tr} (\hat{\mathbf{B}} \mathbf{z}_p \mathbf{B} \mathbf{z}_i + \mathrm{t.}),
\end{equation}
such that the trace reads
\begin{align}
\mathrm{tr} \left(\left(\mathbf{\Sigma}_{\Delta t}^{1/2} \hat{\mathbf{\Sigma}}_{\Delta t}\mathbf{\Sigma}_{\Delta t}^{1/2}\right)^{1/2}\right) &= \mathrm{tr}(\mathbf{\Sigma}_0) + \frac{\Delta t}{2} \left( \mathrm{tr}(\mathbf{B} +\hat{\mathbf{B}}) \right)+\frac{\Delta t^2}{2} \left( \mathrm{tr}(\mathbf{C} + \hat{\mathbf{C}}) \right) \notag \\
&+ \frac{\Delta t^2}{2} \sum_{i,p} \frac{\sigma_p^2}{(\sigma_i^2 + \sigma_p^2)^2} \mathrm{tr} (\hat{\mathbf{B}} \mathbf{z}_p \mathbf{B} \mathbf{z}_i + \mathbf{B} \mathbf{z}_p \hat{\mathbf{B}} \mathbf{z}_i - \mathbf{B} \mathbf{z}_p \mathbf{B} \mathbf{z}_i - \hat{\mathbf{B}} \mathbf{z}_p \hat{\mathbf{B}} \mathbf{z}_i),
\end{align}
which leaves us with
\begin{align}
\mathrm{tr} \left(\left(\mathbf{\Sigma}_{\Delta t}^{1/2} \hat{\mathbf{\Sigma}}_{\Delta t}\mathbf{\Sigma}_{\Delta t}^{1/2}\right)^{1/2}\right) &= \mathrm{tr}(\mathbf{\Sigma}_0) + \frac{\Delta t}{2} \left( \mathrm{tr}(\mathbf{B} +\hat{\mathbf{B}}) \right)+\frac{\Delta t^2}{2} \left( \mathrm{tr}(\mathbf{C} + \hat{\mathbf{C}}) \right) - \frac{\Delta t^2}{2} \sum_{i,p} \frac{\sigma_p^2}{(\sigma_i^2 + \sigma_p^2)^2} \mathrm{tr} \left((\hat{\mathbf{B}} - \mathbf{B}) \mathbf{z}_p (\hat{\mathbf{B}} - \mathbf{B})\mathbf{z}_i\right) \label{trsqroot}
\end{align}
The covariance part of the Wasserstein loss is $\mathrm{tr}(\tilde{\mathbf{\Sigma}}_{\Delta t}) + \mathrm{tr}(\mathbf{\Sigma}_{\Delta t}) - 2\mathrm{tr} ((\mathbf{\Sigma}_{\Delta t}^{1/2} \hat{\mathbf{\Sigma}}_{\Delta t}\mathbf{\Sigma}_{\Delta t}^{1/2})^{1/2})$. Using the small time-step expansion of the covariance matrices Eq.~(\ref{dt_cov}), only the last term Eq.~(\ref{trsqroot}) remains and the Wasserstein loss now reads
\begin{equation}
\mathcal{W}_2^2(\hat{p}_{\Delta t}, p_{\Delta t}) = \Delta t^2 \|(\hat{\mathbf{\Omega}} - \mathbf{\Omega})\mathbf{m}_0\|^2 + \Delta t^2 \sum_{i,p} \frac{\sigma_p^2}{(\sigma_i^2 + \sigma_p^2)^2} \mathrm{tr} \left((\hat{\mathbf{B}} - \mathbf{B}) \mathbf{z}_p (\hat{\mathbf{B}} - \mathbf{B})\mathbf{z}_i\right) + o(\Delta t^2). \label{wass_exp}
\end{equation}
Using Eq.~(\ref{cov_gen}) we find that $\mathbf{B} = \mathbf{\Omega} \mathbf{\Sigma}_0 + \mathbf{\Sigma}_0 \mathbf{\Omega}^T + 2 \mathbf{D}$ and $\mathbf{B} = \hat{\mathbf{\Omega}} \mathbf{\Sigma}_0 + \mathbf{\Sigma}_0 \hat{\mathbf{\Omega}}^T + 2 \hat{\mathbf{D}}$ such that
\begin{equation}
\mathcal{W}_2^2(\hat{p}_{\Delta t}, p_{\Delta t}) = \Delta t^2 \|(\hat{\mathbf{\Omega}} - \mathbf{\Omega})\mathbf{m}_0\|^2 + \Delta t^2 \sum_{i,p} \frac{\sigma_p^2}{(\sigma_i^2 + \sigma_p^2)^2} \left( \mathbf{w}_i^T \left((\hat{\mathbf{\Omega}} - \mathbf{\Omega})\mathbf{\Sigma}_0 + \mathbf{\Sigma}_0 (\hat{\mathbf{\Omega}} - \mathbf{\Omega})^T + 2 (\hat{\mathbf{D}} - \mathbf{D})\right) \mathbf{w}_p \right)^2 + o(\Delta t^2).
\end{equation}
Doing this for every term of the sum in Eq.~(\ref{loss}) we recognize a Riemann sum such that in the limit $\Delta t \rightarrow 0$ the loss $\mathcal{L}_{\Delta t, K_{\Delta t}}/ \Delta t \rightarrow \mathcal{L}$ were $\mathcal{L}$ is given by Eq.~(\ref{wass_loss}). The function $\mathbf{X} \longmapsto \mathrm{tr}(\mathbf{X} \mathbf{A} \mathbf{X}^T \mathbf{B})$ is convex for any $\mathbf{A}, \mathbf{B}$ positive semi-definite matrices. Since $\mathbf{m}_i\mathbf{m}_i^T, \mathbf{z}_i$ for $1 \leq i \leq d$ are positive semi-definite, $\hat{\mathbf{\Omega}} \longmapsto \hat{\mathbf{\Omega}} - \mathbf{\Omega}$ and $\hat{\mathbf{\Omega}} \longmapsto (\hat{\mathbf{\Omega}} - \mathbf{\Omega})\mathbf{\Sigma}_0 + \mathbf{\Sigma}_0 (\hat{\mathbf{\Omega}} - \mathbf{\Omega})^T + 2 (\hat{\mathbf{D}} - \mathbf{D})$ are affine functions of $\hat{\mathbf{\Omega}}$, we can conclude that the mean and covariance part of the loss are convex functions of $\hat{\mathbf{\Omega}}$. The function $\hat{\mathbf{\Omega}} \longmapsto \|\hat{\mathbf{\Omega}}\|^2_F$ being strongly convex and the sum of convex and strongly convex functions being strongly convex, so is the loss function.
\end{proof}

\subsection{Generalization to the Sinkhorn divergence}\label{sec:SM_sinkhorn}
We can extend this approach to a more general version of the loss replacing the Wasserstein distance by the Sinkhorn divergence. The Sinkhorn divergence between the inferred and true processes with entropic regularization $\epsilon \geq  0$ reads as follows
\begin{equation}
\mathcal{S}_{2,\epsilon} (\hat{p}_{\Delta t},p_{\Delta t}) = \mathcal{W}_{2,\epsilon}^2 (\hat{p}_{\Delta t},p_{\Delta t}) - \frac{1}{2} \left( \mathcal{W}_{2,\epsilon}^2 (p_{\Delta t},p_{\Delta t}) + \mathcal{W}_{2,\epsilon}^2 (\hat{p}_{\Delta t}, \hat{p}_{\Delta t}) \right).
\end{equation}
where $\mathcal{W}_{2,\epsilon}^2$ denotes the entropy-regularized Wasserstein distance, which reads for Gaussian distributions
\begin{gather}
\mathcal{W}_{2,\epsilon} (\hat{p}_{\Delta t},p_{\Delta t}) = \| \hat{\mathbf{m}}_{\Delta t} - \mathbf{m}_{\Delta t} \|^2 + \mathrm{tr} (\hat{\mathbf{\Sigma}}_{\Delta t}) + \mathrm{tr} (\mathbf{\Sigma}_{\Delta t}) + \frac{\epsilon}{2} \log \mathrm{det} \left( \mathbf{I} + \frac{1}{2} \mathbf{M}_{\epsilon}(\hat{\mathbf{\Sigma}}_{\Delta t}, \mathbf{\Sigma}_{\Delta t}) \right) - \frac{\epsilon}{2} \mathrm{tr} \left( \mathbf{M}_{\epsilon} (\hat{\mathbf{\Sigma}}_{\Delta t}, \mathbf{\Sigma}_{\Delta t})\right), \\
\text{where } \mathbf{M}_{\epsilon} (\hat{\mathbf{\Sigma}}_{\Delta t}, \mathbf{\Sigma}_{\Delta t}) = -\mathbf{I} + \left( \mathbf{I} + \frac{16}{\epsilon^2} \mathbf{\Sigma}_{\Delta t}^{1/2} \hat{\mathbf{\Sigma}}_{\Delta t} \mathbf{\Sigma}_{\Delta t}^{1/2}\right)^{1/2}.
\end{gather}
We can perform a similar expansion in powers of $\Delta t$ for this loss, using the matrix square root expansion illustrated above, along with the matrix logarithm expansion. This latter expansion around a matrix $\mathbf{X}$ reads as follows, with $u \rightarrow 0$
\begin{align}
\log (\mathbf{X} + u\mathbf{Y}) = \log \mathbf{X} + \int_0^{\infty} dt \Bigg( u(\mathbf{X} + t\mathbf{I})^{-1} \mathbf{Y} (\mathbf{X} + t\mathbf{I})^{-1} \notag -  u^2(\mathbf{X} + t\mathbf{I})^{-1}\mathbf{Y} (\mathbf{X} + t\mathbf{I})^{-1} \mathbf{Y} (\mathbf{X} + t\mathbf{I})^{-1}\Bigg) + O(u^3).
\end{align}
Thanks to the similarity of the matrices $\mathbf{\Sigma}_{\Delta t}^{1/2} \hat{\mathbf{\Sigma}}_{\Delta t}\mathbf{\Sigma}_{\Delta t}^{1/2}$ and $\hat{\mathbf{\Sigma}}_{\Delta t}^{1/2} \mathbf{\Sigma}_{\Delta t}\hat{\mathbf{\Sigma}}_{\Delta t}^{1/2}$ we can directly simplify many terms in the expansion to account for the symmetry between inferred and true matrices. The calculations are nonetheless much lengthier than for the Wasserstein distance, and we only provide the end result for the continuous-time loss
\begin{align}
\mathcal{L}_{\epsilon}& = \mathrm{tr} \left( (\hat{\mathbf{\Omega}} - \mathbf{\Omega})^T(\hat{\mathbf{\Omega}} - \mathbf{\Omega} ) \mathbf{P} + \lambda T \hat{\mathbf{\Omega}} \hat{\mathbf{\Omega}}^T\right)  \\
&+ \int_0^T  \sum_{i,p} \frac{\xi_{i,t}}{\left( \xi_{i,t} + \xi_{p,t}\right)^2} \left(\mathbf{w}_{i,t}^T \left(\sigma_{p,t}^2 (\hat{\mathbf{\Omega}} - \mathbf{\Omega}) + \sigma_{i,t}^2 (\hat{\mathbf{\Omega}}^T - \mathbf{\Omega}^T) + 2(\hat{\mathbf{D}} - \mathbf{D})\right) \mathbf{w}_{p,t} \right)^2 dt,\notag
\end{align}
where 
\begin{equation}
\xi_{i,t} = \sqrt{\epsilon^2/16 + \sigma_{i,t}^4}, \; \mathbf{P} = \int_0^T \mathbf{m}_t \mathbf{m}_t^T dt.
\end{equation}
Similarly to the case of the Wasserstein loss, this loss function is strongly convex. We verify that when $\epsilon \rightarrow 0$ we recover the continuous-time loss for the Wasserstein distance Eq.~(\ref{loss}).

\subsection{Solution in the isotropic case}\label{sec:SM_isotropic}
The matrix $\mathbf{P}$ is symmetric positive semi-definite since 
\begin{equation}
\mathbf{x}^T \mathbf{P} \mathbf{x} = \int_0^T \mathbf{x}^T e^{\mathbf{\Omega} t} \mathbf{m}_0 \mathbf{m}_0^T e^{\mathbf{\Omega}^T t} \mathbf{x} dt = \int_0^T \| \mathbf{x}^T e^{\mathbf{\Omega} t} \mathbf{m}_0 \|^2 dt \geq 0.
\end{equation}
This matrix is therefore diagonalizable in an orthogonal basis of vectors $\mathbf{U} = (\mathbf{u}_1, ..., \mathbf{u}_d)$ of $\mathbb{R}^d$, such that $\mathbf{P} = \mathbf{U} \bm{\Gamma} \mathbf{U}^T$ where $\bm{\Gamma} = \mathrm{diag}(\gamma_1,...,\gamma_d)$, and $\gamma_1 \geq \gamma_ 2 \geq ... \geq \gamma_d$.

\begin{proposition*}
For an isotropic process $\sigma_{i,t} = \sigma_{p,t}$ for all $i,p$ and $t \geq 0$, and up to terms independent of $\mathbf{\Omega}$ the loss reads
\begin{align}
\mathcal{L} = \mathrm{tr}\bigg( (\hat{\mathbf{\Omega}} - \mathbf{\Omega})^T(\hat{\mathbf{\Omega}} - \mathbf{\Omega})\mathbf{P} + \frac{q}{4}  (\hat{\mathbf{\Omega}} +\hat{\mathbf{\Omega}}^T - 2 \Omega_s \mathbf{I} )^2  + (\hat{\mathbf{\Omega}} + \hat{\mathbf{\Omega}}^T - 2 \Omega_s \mathbf{I} ) (\hat{\mathbf{D}} - D \mathbf{I})  + \lambda T \hat{\mathbf{\Omega}} \hat{\mathbf{\Omega}}^T\bigg) \
\end{align}
where $q = \int_0^T \sigma_t^2dt$. For $\lambda > 0$, the minimum value of this loss is attained for
\begin{equation}
\hat{\mathbf{\Omega}} = \mathbf{\Omega} - \sum_{i,j} \frac{\tilde{\lambda} \tilde{q}^{-1} \eta_{ij} (1 + \gamma_i \tilde{\lambda}^{-1}) + \omega_{ij} (1 + \gamma_i \tilde{q}^{-1})}{\tilde{q}^{-1} \tilde{\lambda}^{-1} \gamma_i \gamma_j + \Gamma_{+}(\gamma_i + \gamma_j) + 1} \mathbf{u}_i \mathbf{u}_j^T. \label{iso_sol}
\end{equation}
with $\tilde{\lambda} = \lambda T, \; \tilde{q} = q  + \tilde{\lambda}, \; \Gamma_{+} = \frac{1}{2}(\tilde{\lambda}^{-1} + \tilde{q}^{-1})$, $\omega_{ij} = \mathbf{u}_i^T \mathbf{\Omega}_a \mathbf{u}_j$ and $\eta_{ij} =  \mathbf{u}_i^T(\Omega_s \mathbf{I} + (\hat{\mathbf{D}} - D \mathbf{I})/\tilde{\lambda}) \mathbf{u}_j$.
\end{proposition*}

\begin{proof}
The loss function $\mathcal{L}$ is a polynomial of the coefficients of $\mathbf{\Omega}$ and is therefore infinitely differentiable over $\mathbb{R}^d$. We can find the minimum by writing the first order optimality condition
\begin{equation}
2 (\hat{\mathbf{\Omega}} - \mathbf{\Omega}) \mathbf{P} + q (\hat{\mathbf{\Omega}}^T + \hat{\mathbf{\Omega}} - 2\Omega_s \mathbf{I}) + 2 (\hat{\mathbf{D}} - D \mathbf{I}) + 2\lambda T \hat{\mathbf{\Omega}} = 0. \label{crit}
\end{equation}
Let's for now assume that $\lambda > 0$. We introduce $\mathbf{C} = \hat{\mathbf{\Omega}} - \mathbf{\Omega}$, and we would like to rewrite this equation in the form of a linear system displaying only $\mathbf{C}$. To do that we separate symmetric and skew-symmetric parts. We denote $\mathbf{Y} =  \Omega_s\mathbf{I} + (\hat{\mathbf{D}} - D \mathbf{I})/(\lambda T)$ such that
\begin{gather}
\mathbf{C}\mathbf{P} + \mathbf{P}\mathbf{C}^T + q (\mathbf{C} + \mathbf{C}^T) + \lambda T(\mathbf{C} + \mathbf{C}^T) + 2 \lambda T \mathbf{Y} = 0, \label{sym}\\
\mathbf{C}\mathbf{P} - \mathbf{P}\mathbf{C}^T  + \lambda T(\mathbf{C} - \mathbf{C}^T) + 2\lambda T\mathbf{\Omega}_a = 0. \label{skew_1}
\end{gather}
We denote $\tilde{q} = q + \lambda T$ and $\tilde{\lambda} = \lambda T$, and we can solve for $\mathbf{C}^T$ in the first equation and then replace it in the second equation. We use the fact that the matrix $(\tilde{q} \mathbf{I} + \mathbf{P})$ is symmetric positive definite and hence invertible
\begin{equation}
\mathbf{C}^T = - (\tilde{q} \mathbf{I} + \mathbf{P})^{-1} \left( \mathbf{C}(\tilde{q} \mathbf{I} + \mathbf{P}) + 2 \tilde{\lambda} \mathbf{Y}\right). \label{ct_1}
\end{equation}
Using the eigendecomposition of $\mathbf{P}$ and using the Woodbury formula we find
\begin{align}
(\tilde{q}\mathbf{I} + \mathbf{P})^{-1} &= \tilde{q}^{-1}\mathbf{I} - \tilde{q}^{-2} \mathbf{U} (\bm{\Gamma}^{-1} + \mathbf{U}^T \mathbf{U} \tilde{q}^{-1})^{-1} \mathbf{U}^T.
\end{align}
Since $\mathbf{\Gamma}$ is diagonal and $\mathbf{U}^T \mathbf{U} =  \mathbf{I}$, $\bm{\Gamma}^{-1} + \mathbf{U}^T \mathbf{U} \tilde{q}^{-1}$ is diagonal and invertible and $(\bm{\Gamma}^{-1} + \mathbf{U}^T \mathbf{U} \tilde{q}^{-1})_{jj} = 1 + \gamma_j\tilde{q}^{-1}/\gamma_j$ for all $1 \leq j \leq d$, which gives
\begin{equation}
(\tilde{q}\mathbf{I} + \mathbf{P})^{-1} = \tilde{q}^{-1} \mathbf{I} - \sum_{j=1}^{d} \frac{\gamma_j \tilde{q}^{-2}}{1 + \gamma_j \tilde{q}^{-1}} \mathbf{u}_j \mathbf{u}_j^T = \tilde{q}^{-1} (\mathbf{I} - \tilde{q}^{-1}\tilde{\mathbf{P}}), \label{id_1}
\end{equation}
where we have introduced the matrix $\tilde{\mathbf{P}}$ that reads $\tilde{\mathbf{P}} = \sum_{i=1}^{d} \frac{\gamma_i}{1 + \gamma_i \tilde{q}^{-1}} \mathbf{u}_i \mathbf{u}_i^T$. Using Eq.~(\ref{id_1}) we have $\mathbf{P} (\tilde{q} \mathbf{I}+ \mathbf{P})^{-1} = \tilde{q}^{-1}\tilde{\mathbf{P}}$ such that 
\begin{align}
- \mathbf{P} \mathbf{C}^T - \tilde{\lambda} \mathbf{C}^T = (1 - \tilde{\lambda} \tilde{q}^{-1}) \tilde{q}^{-1} \tilde{\mathbf{P}} \left( \mathbf{C}(\tilde{q} \mathbf{I} + \mathbf{P}) + 2 \tilde{\lambda} \mathbf{Y}\right) + \tilde{\lambda} \mathbf{C} + \tilde{\lambda} \tilde{q}^{-1} \mathbf{CP} + 2\tilde{\lambda} \tilde{q}^{-1} \tilde{\lambda} \mathbf{Y}.
\end{align}
Injecting it in Eq.~(\ref{skew_1}) we are left with
\begin{align}
(1 &+ \tilde{\lambda} \tilde{q}^{-1}) \mathbf{CP} + 2 \tilde{\lambda} \mathbf{C} + (1 - \tilde{\lambda} \tilde{q}^{-1}) \tilde{\mathbf{P}} \mathbf{C} + \tilde{q}^{-1}(1 - \tilde{\lambda}\tilde{q}^{-1}) \tilde{\mathbf{P}} \mathbf{CP} + 2 \tilde{\lambda} \tilde{q}^{-1}(1 - \tilde{\lambda}\tilde{q}^{-1}) \tilde{\mathbf{P}} \mathbf{Y} + 2\tilde{\lambda} \tilde{q}^{-1} \tilde{\lambda}\mathbf{Y} + 2\tilde{\lambda} \mathbf{\Omega}_a = 0.
\label{eq:full_eq}
\end{align}
We introduce $\Gamma_{-} = (\tilde{\lambda}^{-1} - \tilde{q}^{-1})/2$ and $\Gamma_{+} = (\tilde{\lambda}^{-1} + \tilde{q}^{-1})/2$, and we divide the full equation Eq.~(\ref{eq:full_eq}) by $2\tilde{\lambda}$ to find
\begin{align}
\Gamma_{+} \mathbf{CP} + \mathbf{C} + \Gamma_{-} \tilde{\mathbf{P}} \mathbf{C} + \tilde{q}^{-1}\Gamma_{-} \tilde{\mathbf{P}} \mathbf{CP} + 2 \tilde{\lambda} \tilde{q}^{-1}\Gamma_{-} \tilde{\mathbf{P}} \mathbf{Y} + \tilde{\lambda} \tilde{q}^{-1} \mathbf{Y} + \mathbf{\Omega}_a = 0 . 
\end{align}
We can now compute the projections $\mathbf{u}_i^T \mathbf{C} \mathbf{u}_j$. We multiply this equation by $\mathbf{u}_i^T$ and $\mathbf{u}_j$ respectively on the left and on the right, and we denote $c_{ij} = \mathbf{u}_i^T \mathbf{C} \mathbf{u}_j$ and $\omega_{ij} = \mathbf{u}_i^T \mathbf{\Omega}_a \mathbf{u}_j$ and $\eta_{ij} = \mathbf{u}_i^T \mathbf{Y} \mathbf{u}_j$
\begin{align}
c_{ij} \left[\Gamma_{+} \gamma_j + 1  + \Gamma_{-} \frac{\gamma_i}{1 + \gamma_i\tilde{q}^{-1}} + \tilde{q}^{-1}\Gamma_{-} \frac{\gamma_i \gamma_j}{1 + \gamma_i\tilde{q}^{-1}}\right] &= - \tilde{\lambda} \tilde{q}^{-1} \eta_{ij} -  2\tilde{\lambda} \tilde{q}^{-1} \Gamma_{-} \frac{\gamma_i}{1 + \gamma_i\tilde{q}^{-1}} \eta_{ij} - \omega_{ij} \notag \\
&= - \tilde{\lambda} \tilde{q}^{-1} \frac{1 + \gamma_i \tilde{\lambda}^{-1}}{1 + \gamma_i \tilde{q}^{-1}} \eta_{ij} - \omega_{ij}.
\end{align}
The term in brackets reads is always positive and reads
\begin{align}
\Gamma_{+} \gamma_j + 1  + \Gamma_{-} \frac{\gamma_i}{1 + \gamma_i\tilde{q}^{-1}} + \tilde{q}^{-1}\Gamma_{-} \frac{\gamma_i \gamma_j}{1 + \gamma_i\tilde{q}^{-1}} = \frac{\tilde{q}^{-1} \tilde{\lambda}^{-1} \gamma_i \gamma_j + (\gamma_i + \gamma_j)\Gamma_{+} + 1}{1 + \gamma_i \tilde{q}^{-1}},
\end{align}
leading us to
\begin{equation}
c_{ij} = - \frac{\tilde{\lambda} \tilde{q}^{-1} \eta_{ij} (1 + \gamma_i \tilde{\lambda}^{-1}) + \omega_{ij} (1 + \gamma_i \tilde{q}^{-1})}{\tilde{q}^{-1} \tilde{\lambda}^{-1} \gamma_i \gamma_j + \Gamma_{+}(\gamma_i + \gamma_j) + 1}.
\end{equation}
Knowing the matrix elements of $\mathbf{C}$ in the basis $\mathbf{U}$, we can write $\mathbf{C}$ as a sum of outer products
\begin{equation}
\mathbf{C} = \sum_{i,j} c_{ij} \mathbf{u}_i \mathbf{u}_j^T,
\end{equation}
such that the solution reads
\begin{equation}
\mathbf{C} =- \sum_{i,j} \frac{\tilde{\lambda} \tilde{q}^{-1} \eta_{ij} (1 + \gamma_i \tilde{\lambda}^{-1}) + \omega_{ij} (1 + \gamma_i \tilde{q}^{-1})}{\tilde{q}^{-1} \tilde{\lambda}^{-1} \gamma_i \gamma_j + \Gamma_{+}(\gamma_i + \gamma_j) + 1} \mathbf{u}_i \mathbf{u}_j^T. \label{bias}
\end{equation}
\end{proof}
In the case $\mathbf{P} = 0$ only the symmetric part can be inferred and the solution reduces to $\hat{\mathbf{\Omega}} = \Omega_s (1 - \tilde{\lambda} \tilde{q}^{-1}) \mathbf{I}$. The skew symmetric part is only observable through rigid-body rotations of the distribution, which are not observable when the mean is zero. In order to work out the $\lambda \rightarrow 0$ limit, we introduce $l = \mathrm{rank(\mathbf{P})}$ which defines the ambient dimension of the inferred process. It is also useful to separate the sum in four quadrants $(i \leq l, j\leq l)$, $(i > l, j\leq l)$, $(i \leq l, j > l)$ and $(i > l, j > l)$. By doing so, in the limit $\lambda \rightarrow 0$ we find that the minimum tends to
\begin{equation}
\hat{\mathbf{\Omega}} = \mathbf{\Omega} - (\mathbf{I} - \mathbf{Q}) (\mathbf{\Omega}_a + q^{-1}(\hat{\mathbf{D}} - D \mathbf{I}))(\mathbf{I} - \mathbf{Q}) - 2\sum_{\substack{j \leq d \\  i \leq l}} \frac{\gamma_i  q^{-1} \mu_{ij} }{2q^{-1} \gamma_i \gamma_j + \gamma_i + \gamma_j} \mathbf{u}_i \mathbf{u}_j^T, \label{lim_zero_bias}
\end{equation}
where $\mathbf{Q}$ is the orthogonal projector on $\mathrm{range}(\mathbf{P})$ and $\mu_{ij} = \mathbf{u}_i^T (\hat{\mathbf{D}} - D \mathbf{I}) \mathbf{u}_j$. If $\mathrm{rank}(\mathbf{P})  \geq d - 1$ and $\hat{\mathbf{D}} = \mathbf{D}$ we recover the true matrix $\hat{\mathbf{\Omega}} = \mathbf{\Omega}$. Assuming $\hat{\mathbf{D}} = \mathbf{D}$, we can provide an approximation to Eq.~(\ref{bias}) when $|\tilde{\lambda}^{-1} \gamma_i - 1| \gg 1$ and $\tilde{\lambda} \ll q$. To derive it we need to separate the sum in four quadrants $(\gamma_i \ll \tilde{\lambda}, \gamma_j \ll \tilde{\lambda})$, $(\gamma_i \ll \tilde{\lambda}, \gamma_j \gg \tilde{\lambda})$, $(\gamma_i \gg \tilde{\lambda}, \gamma_j \ll \tilde{\lambda})$, $(\gamma_i \gg \tilde{\lambda}, \gamma_j \gg \tilde{\lambda})$. In the first quadrant the denominator reduces to $1$ and the numerator to $\omega_{ij}$. The sum then reads
\begin{equation}
\sum_{\substack{\gamma_i \ll \tilde{\lambda} \\ \gamma_j \ll \tilde{\lambda}}} \frac{\tilde{\lambda} \tilde{q}^{-1} \eta_{ij} (1 + \gamma_i \tilde{\lambda}^{-1}) + \omega_{ij} (1 + \gamma_i \tilde{q}^{-1})}{\tilde{q}^{-1} \tilde{\lambda}^{-1} \gamma_i \gamma_j + \Gamma_{+}(\gamma_i + \gamma_j) + 1} \mathbf{u}_i \mathbf{u}_j^T = \sum_{\substack{\gamma_i \ll \tilde{\lambda} \\ \gamma_j \ll \tilde{\lambda}}} \omega_{ij} \mathbf{u}_i \mathbf{u}_j^T = \mathbf{Q}_{\tilde{\lambda}} \mathbf{\Omega}_a \mathbf{Q}_{\lambda},
\end{equation}
where $\mathbf{Q}_{\tilde{\lambda}}$ is the orthogonal projector on this first quadrant. In the second quadrant, the denominator will, at worst be of order $\gamma_j \tilde{\lambda}^{-1} \gg 1$, and the numerator of order $1$, such that the sum vanishes. In the third quadrant the denominator will be, at worst, of order $\gamma_i \tilde{\lambda}^{-1} \gg 1$, and the numerator reduces to $\tilde{\lambda} q^{-1} \gamma_i \tilde{\lambda}^{-1} \eta_{ij} + \omega_{ij}$, such that the sum vanishes since $\tilde{\lambda} q^{-1} \ll 1$. In the last quadrant, the sum vanishes for the same reasons, and under the aforementioned conditions on $\tilde{\lambda}$ and the $\gamma_i$'s we have $\hat{\mathbf{\Omega}} \approx \mathbf{\Omega} - \mathbf{Q}_{\tilde{\lambda}} \mathbf{\Omega}_a  \mathbf{Q}_{\tilde{\lambda}}$.

\rev{\subsection{First order finite sample size correction to the loss function}
\label{sec:SM_variance}
\paragraph{One-dimensional example}
To illustrate the effect of finite sample size, we first consider that the true process is a one-dimensional Ornstein-Uhlenbeck process with steady-state mean value $\mu$
\begin{equation}
dx = w(x - \mu)dt + \sqrt{2D} dW,
\end{equation}
with $w < 0$, $D$ the diffusion constant, $W$ a Wiener process and $x_{t_0} \sim \mathcal{N}(m_{t_0}, \Sigma_{t_0})$. At all times $x_t \sim \mathcal{N}(m_t, \Sigma_t)$ with $m_{t+u} = e^{w u} m_{t} + \mu(1 - e^{w u})$ for all $0<u$. The inferred process is the solution of the PF ODE which reads
\begin{equation}
\frac{d\hat{x}}{dt} = \hat{w}(\hat{x} - \mu) + D \Sigma_t^{-1}(\hat{x} - m_t),
\end{equation}
and we are left with estimating the coefficient $\hat{w}$. In principle, we could also estimate $\mu$, but for sake of simplicity of the argument we consider it is known. Finally, we also consider that the covariance is already set at its steady-state value $\Sigma_t = \Sigma_{\infty}$ such that it doesn't change in time. With this last assumption, the loss at time $t_i$ reduces to
\begin{equation}
L(\mathcal{N} (m_{t_i}, \Sigma_{\infty}), \mathcal{N} (\hat{m}_{t_i}, \Sigma_{\infty})) = \left( m_{t_i} - \hat{m}_{t_i} \right)^2 + \lambda \Delta t^2 \hat{w}^2 .
\end{equation}
Because the true process is sampled once and for all at the time of optimization, at each time $t_i$ we don't have access to $m_{t_i}$, but only to its estimate obtained with $n$ samples,
\begin{equation}
m_{t_i,n} = m_{t_i} + \frac{1}{\sqrt{n}} \Sigma_{\infty}^{1/2} h_{t_i},
\end{equation}
with $h_{t_i}$ a standard normal random variable. For this reason, we minimize $L(\mathcal{N} (m_{t_i,n}, \Sigma_{\infty}), \mathcal{N} (\hat{m}_{t_i,n}, \Sigma_{\infty}))$ rather than $L(\mathcal{N} (m_{t_i}, \Sigma_{\infty}), \mathcal{N} (\hat{m}_{t_i}, \Sigma_{\infty}))$. For the true process, the marginals evolve and are sampled independently at each time, such that we only need to write the evolution of the mean of a OU process plus the finite sampling error. This reads
\begin{equation}
m_{t_i,n} = m_{t_i} + \frac{1}{\sqrt{n}} \Sigma_{\infty}^{1/2} h_{t_i} = e^{w \Delta t} m_{t_{i-1}} + \mu(1 - e^{w \Delta t}) + \frac{1}{\sqrt{n}} \Sigma_{\infty}^{1/2} h_{t_i}.
\end{equation}
On the other hand, the inferred process is solution of the PF ODE, with the observed samples as initial condition. The mean of the inferred process $\hat{m}_{t_i,n}$ then reads, at lowest order in $\Delta t$:
\begin{equation}
\hat{m}_{t_i,n} = m_{t_{i-1},n} + \hat{w} \Delta t( m_{t_{i-1},n} - \mu) + D \Delta t\Sigma_{\infty}^{-1} (m_{t_{i-1},n} - m_{t_{i-1}}).
\end{equation}
This gives us the difference between the true and inferred estimated means at the lowest order in $\Delta t$:
\begin{equation}
(\hat{m}_{t_i,n} - m_{t_i,n}) = \Delta t(\hat{w} - w)(m_{t_{i-1}} - \mu) + \frac{1}{\sqrt{n}}\Sigma_{\infty}^{1/2} h_{t_{i-1}} + \hat{w} \frac{\Delta t}{\sqrt{n}} \Sigma_{\infty}^{1/2} h_{t_{i-1}} + \Delta t \frac{D\Sigma_{\infty}^{-1/2}}{\sqrt{n}} h_{t_{i-1}} - \frac{1}{\sqrt{n}} \Sigma_{\infty}^{1/2} h_{t_i}.
\end{equation}
We additionally assume that $\sqrt{n} \Delta t\rightarrow \text{constant}$ when $\Delta t \rightarrow 0$, which also implies that $1/\sqrt{n} \ll 1$. The $1/\sqrt{n}$ noise terms are therefore negligible, and we have telescopic sums for the remaining noise terms. This leaves us with
\begin{equation}
\sum_{i=1}^{K} L(m_{t_i, n},\hat{m}_{t_i,n})/\Delta t \rightarrow \int_0^T \left( (\hat{w} - w) (m_{t} - \mu) + \frac{\Sigma_{\infty}^{1/2} }{\sqrt{n} \Delta t}(h_T - h_0) \right)^2 dt + \lambda T \hat{w}^2
\end{equation}
Minimizing this loss function as a function of $\hat{w}$, we find, with $\tilde{\lambda}$ defined as in Eq.~(\ref{iso_sol}),
\begin{equation}
\int_0^T (m_t - \mu) \left( (\hat{w} - w) (m_t - \mu) + \frac{\Sigma_{\infty}^{1/2} }{\sqrt{n} \Delta t}(h_T - h_0) \right) dt + \tilde{\lambda} \hat{w}= 0.
\end{equation}
This gives us the following error on $w^{*}$, the optimal value when $n \rightarrow\infty$:
\begin{equation}
| \hat{w} - w^{*} | = \bigg|\frac{\Sigma_{\infty}^{1/2} (h_T -  h_0)}{\sqrt{n} \Delta t} \frac{\int_0^T (m_t - \mu) dt}{(\tilde{\lambda} + \int_0^T(m_t - \mu)^2dt)} \bigg|.
\end{equation}
We see here that the error diverges not only if $\sqrt{n} \Delta t \ll 1$, but also if the non-dimensional prefactor diverges, i.e. if
\begin{equation}
\Sigma_{\infty}^{1/2}\bigg|(h_T - h_0) \frac{\int_0^T (m_t - \mu) dt}{(\tilde{\lambda} + \int_0^T(m_t - \mu)^2dt)}\bigg|\gg 1.
\end{equation}
When the data approaches steady-state, this factor is controlled by the inverse of the regularization, and the smaller the regularization the larger the error.}

\paragraph{In arbitrary dimensions}
\rev{Following the approach of the previous one-dimensional example, we derive the first order finite sample size correction to the loss function in arbitrary dimensions. We first need to state the central limit theorem for the mean and covariance of normally distributed i.i.d random variables. Let $\mathbf{x}_1, ..., \mathbf{x}_n$ be independent samples drawn from a probability law $\mathbb{P}$. We will denote the sample covariance and sample mean of this law as
\begin{equation}
\mathbf{m}_n = \frac{1}{n} \sum_{i=1}^n \mathbf{x}_i \text { and } \mathbf{\Sigma}_n = \frac{1}{n-1} \sum_{i=1}^n (\mathbf{x}_i - \mathbf{m}_n)(\mathbf{x}_i - \mathbf{m}_n)^T.
\end{equation}
Then, if $\mathbb{P} = \mathcal{N}(\mathbf{m}, \mathbf{\Sigma})$, we have the following asymptotic results (see \cite{rippl2016limit} Lemma 4.2.)
\begin{equation}
\sqrt{n}(\mathbf{m}_n - \mathbf{m}) \xrightarrow[n \rightarrow \infty]{d} \mathbf{\Sigma}^{1/2}\mathbf{h} \text{ and } \sqrt{n}(\mathbf{\Sigma}_n - \mathbf{\Sigma}) \xrightarrow[n \rightarrow \infty]{d} \mathbf{\Sigma}^{1/2}\mathbf{H}\mathbf{\Sigma}^{1/2}, \label{clt}
\end{equation}
where $\xrightarrow[]{d}$ denotes convergence in distribution, $\mathbf{h}$ and $\mathbf{H}$ are independent random variables such that $\mathbf{h} \sim \mathcal{N}(0, \mathbf{I})$ and $\mathbf{H}$ is drawn from the Gaussian Orthogonal Ensemble (GOE), i.e. $\mathbf{H}$ is a real random symmetric matrix with entries satisfying
\begin{equation}
H_{ij}=\biggl\{
  \begin{array}{@{}ll@{}}
    \mathcal{N}(0,1), & \text{if}\ i > j \\
    \mathcal{N}(0,2), & \text{if}\ i = j
  \end{array}\;.
\end{equation}
This allows us to derive the following result
\begin{proposition*}
When $\Delta t \rightarrow 0$, with $\sqrt{n} \Delta t$ fixed, the loss function $\mathcal{L}_{\Delta t, K,n}/\Delta t$ tends to $\mathcal{L}_{\sqrt{n} \Delta t}$ which reads
\begin{align} \label{wass_loss}
\mathcal{L}_{\sqrt{n} \Delta t} &= \mathcal{L} + \frac{2}{\sqrt{n}\Delta t}\mathrm{tr} \left( (\hat{\mathbf{\Omega}} - \mathbf{\Omega})^T \int_0^T \mathbf{\Sigma}_t^{1/2} (\mathbf{h}_t - \mathbf{h}_{t + \Delta t}) \mathbf{m}_t^T dt \right) \label{loss_n} \\
&+  \int_0^T  \sum_{i,p} \frac{\sigma_{i,t}^2}{\left( \sigma_{i,t}^2 + \sigma_{p,t}^2\right)^2} \frac{2 \sigma_{i,t} \sigma_{p,t}}{\sqrt{n} \Delta t} \mathbf{w}_{i,t}^T \bigg(\sigma_{p,t}^2 (\hat{\mathbf{\Omega}} - \mathbf{\Omega})\notag + \sigma_{i,t}^2 (\hat{\mathbf{\Omega}}^T - \mathbf{\Omega}^T) + 2(\hat{\mathbf{D}} - \mathbf{D})\bigg) \mathbf{w}_{p,t} \mathbf{w}_{i,t}^T \bigg(  (\mathbf{H}_t - \mathbf{H}_{t + \Delta t})\bigg) \mathbf{w}_{p,t} \bigg] dt, 
\end{align}
plus constant terms independent of $\hat{\mathbf{\Omega}}$. Additionally, the noise terms in this loss are telescoping at this lowest order in $\Delta t$, and only the most extreme contributions $t = 0$ and $t = T$ to the integrals remain.
\end{proposition*}
\begin{proof}
Using the central limit theorem, we have that the estimated mean and covariance of the true process read, at time $t = 0$ and time $t = \Delta t$,
\begin{gather}
\mathbf{\Sigma}_{n,0} = \mathbf{\Sigma}_{0} + \frac{1}{\sqrt{n}}\mathbf{\Sigma}_{0}^{1/2} \mathbf{H}_{0} \mathbf{\Sigma}_{0}^{1/2}, \; \mathbf{m}_{n,0} = \mathbf{m}_{0} + \frac{1}{\sqrt{n}}\mathbf{\Sigma}_{0}^{1/2} \mathbf{h}_{0},\\
\mathbf{\Sigma}_{n,\Delta t} = \mathbf{\Sigma}_{\Delta t} + \frac{1}{\sqrt{n}}\mathbf{\Sigma}_{\Delta t}^{1/2} \mathbf{H}_{\Delta t} \mathbf{\Sigma}_{\Delta t}^{1/2}, \; \mathbf{m}_{n,\Delta t} = \mathbf{m}_{\Delta t} + \frac{1}{\sqrt{n}}\mathbf{\Sigma}_{\Delta t}^{1/2} \mathbf{h}_{\Delta t}.
\end{gather}
Keeping the lowest order in $\Delta t$ with $\sqrt{n}\Delta t$ constant for the mean and covariance of the true process at time $\Delta t$, we have
\begin{align}
\mathbf{\Sigma}_{n,\Delta t} &= \mathbf{\Sigma}_{0} + \Delta t( \mathbf{\Omega} \mathbf{\Sigma}_0 + \mathbf{\Sigma}_0 \mathbf{\Omega}^T + \frac{1}{\sqrt{n} \Delta t}\mathbf{\Sigma}_{0}^{1/2} \mathbf{H}_{\Delta t} \mathbf{\Sigma}_{0}^{1/2}),
\\
\mathbf{m}_{n,\Delta t} &= \mathbf{m}_{0} + \Delta t(\mathbf{\Omega} \mathbf{m}_0 + \frac{1}{\sqrt{n} \Delta t}\mathbf{\Sigma}_{0}^{1/2} \mathbf{h}_{\Delta t}).
\end{align}
On the other hand, the mean and covariance of the inferred process read
\begin{align}
\hat{\mathbf{m}}_{n,\Delta t} &= \frac{1}{n} \sum_k \hat{\mathbf{x}}_{k,\Delta t} = \left(\mathbf{I} + \Delta t \hat{\mathbf{\Omega}} \right) \mathbf{m}_{n,0} + \hat{\mathbf{D}} \mathbf{\Sigma}_0^{-1} (\mathbf{m}_{n,0} - \mathbf{m}_0), \\
\hat{\mathbf{\Sigma}}_{n,\Delta t} &= \frac{1}{n} \sum_k (\hat{\mathbf{x}}_{k,\Delta t} - \hat{\mathbf{m}}_{n,\Delta t})(\hat{\mathbf{x}}_{k,\Delta t} - \hat{\mathbf{m}}_{n,\Delta t})^T = (\mathbf{I} + \Delta t (\hat{\mathbf{\Omega}} + \hat{\mathbf{D}} \mathbf{\Sigma}_0^{-1}) ) \mathbf{\Sigma}_{0,n} (\mathbf{I} + \Delta t (\hat{\mathbf{\Omega}} + \hat{\mathbf{D}} \mathbf{\Sigma}_0^{-1}))^T.
\end{align}
We only keep the lowest order in $\Delta t$ with $\sqrt{n}\Delta t$ constant, which gives us
\begin{align}
\hat{\mathbf{m}}_{n,\Delta t} &= \mathbf{m}_0 + \Delta t (\hat{\mathbf{\Omega}} \mathbf{m}_0 + \frac{1}{\sqrt{n} \Delta t} \mathbf{\Sigma}_0^{1/2} \mathbf{h}_0), \\
\hat{\mathbf{\Sigma}}_{n,\Delta t} &= \mathbf{\Sigma}_0 + \Delta t (\hat{\mathbf{\Omega}} \mathbf{\Sigma}_0 + \mathbf{\Sigma}_0 \hat{\mathbf{\Omega}} + 2 \hat{\mathbf{D}} + \frac{1}{\sqrt{n} \Delta t} \mathbf{\Sigma}_0^{1/2} \mathbf{H}_0 \mathbf{\Sigma}_0^{1/2}).
\end{align}
We now have small $\Delta t$ expansions for empirical inferred and true means and covariances. We only need to plug the first order terms in these expansion in Eq.~(\ref{wass_exp}). This gives us the expected result.
\end{proof}
\begin{proposition*}
For an isotropic process $\sigma_{i,t} = \sigma_{p,t}$ for all $i,p$ and $t >0$, the loss function $\mathcal{L}_{\sqrt{n} \Delta t}$ reads, up to terms independent of $\hat{\mathbf{\Omega}}$:
\begin{equation}
\mathcal{L}_{\sqrt{n}\Delta t} = \mathcal{L} + \frac{1}{\sqrt{n}\Delta t} \left( 2 \mathrm{tr} \left((\hat{\mathbf{\Omega}} - \mathbf{\Omega})^T \mathbf{Y} \right) + \frac{1}{2} \mathrm{tr}\left( (\hat{\mathbf{\Omega}} + \hat{\mathbf{\Omega}}^T - 2\Omega_s) \mathbf{W} \right)\right)
\end{equation}
where $\mathbf{Y} = (\sigma_0 \mathbf{h}_0 \mathbf{m}_0^T- \sigma_T \mathbf{h}_T \mathbf{m}_T^T)$ and $\mathbf{W} = (\sigma_0^2 \mathbf{H}_0 - \sigma_T^2 \mathbf{H}_T)$. This loss attains its unique minimum in $\hat{\mathbf{\Omega}}$ defined as 
\begin{equation}
\hat{\mathbf{\Omega}} = \mathbf{\Omega} - \sum_{i,j} \frac{\tilde{\lambda} \tilde{q}^{-1} \eta_{ij} (1 + \gamma_i \tilde{\lambda}^{-1}) + \omega_{ij} (1 + \gamma_i \tilde{q}^{-1})}{\tilde{q}^{-1} \tilde{\lambda}^{-1} \gamma_i \gamma_j + \Gamma_{+}(\gamma_i + \gamma_j) + 1} \mathbf{u}_i \mathbf{u}_j^T,
\end{equation}
$\omega_{ij} = \mathbf{u}_i^T (\mathbf{\Omega}_a + \mathbf{Z}_a) \mathbf{u}_j$ and $\eta_{ij} =  \mathbf{u}_i^T(\Omega_s \mathbf{I} + (\hat{\mathbf{D}} - D \mathbf{I})/\tilde{\lambda} + \mathbf{Z}_s) \mathbf{u}_j$, where $\mathbf{Z}_a$ and $\mathbf{Z}_s$ are respectively the skew-symmetric and symmetric parts of $\mathbf{Z} = \mathbf{Z} = (\mathbf{Y} + \mathbf{W}/2)/(\tilde{\lambda} n^{-1/2} \Delta t)$.
\end{proposition*}
\begin{proof}
We take the isotropic limit of Eq.~(\ref{loss_n}), and we then use the same proof as for Eq.~(\ref{iso_sol}).
\end{proof}}

\newpage\section{Generating gene expression data} \label{sec:SM_boolode}
\paragraph{BooldODE formalism.}
The Chemical Master Equation (CME) provides a framework for modeling stochastic gene transcription dynamics \cite{gillespie1992rigorous}. Gillespie’s stochastic simulation algorithm (SSA) \cite{gillespie1976general}, \cite{gillespie1977exact} allows for the computation of reaction trajectories governed by the CME. For gene transcription and translation, the key reactions are:
\begin{align}
    x_i \xrightarrow{ m g_i(\mathbf{R})} x_i + 1, \quad & x_i \xrightarrow{\ell_x x_i } x_i - 1, \\
    p_i \xrightarrow{r x_i} p_i + 1, \quad & p_i \xrightarrow{\ell_p p_i} p_i - 1.
\end{align}
Here, \(x_i\) represents mRNA molecules, and \(p_i\) denotes protein molecules. The propensities for these reactions incorporate the regulatory interactions controlling the expression of gene (or node) \(i\). These interactions can be expressed as:
\begin{equation}
    \Pr(S) = \frac{\prod_{p \in S} \left( \frac{p}{k} \right)^n}{1 + \sum_{S \in 2^{R_i - 1}} \prod_{p \in S} \left( \frac{p}{k} \right)^n},
\end{equation}
and the activation function is defined as:
\begin{equation}
    g(R_i) = \sum_{S \in 2^{R_i}} \alpha_S \Pr(S).
\end{equation}
In this formulation, the product in the numerator accounts for all bound regulators in a given configuration \(S\), while the sum in the denominator includes all possible configurations in the powerset of regulators \(R_i\), excluding the empty set. The parameters are defined as \(m = 20\), \(\ell_x = 5\), \(r = 1\), \(\ell_p = 1\), \(k = 10\), and \(n = 10\), following \cite{pratapa2020benchmarking}. To simplify the system and reduce the number of variables, we assume that protein dynamics equilibrate faster than mRNA dynamics. Under this approximation, the protein abundance can be expressed in terms of mRNA levels, i.e., \(\bar{p}_i = \left(\frac{r}{\ell_p}\right) x_i\). This assumption, while commonly used, may or may not hold depending on the biological system in question. In some cases, the reverse limit—where mRNA equilibrates faster than protein (known as the quasi-steady-state approximation, QSSA)—is equally valid and more widely adopted.\\

\paragraph{mCAD model.} The Boolean rules associated with the mCAD network are adapted from \cite{giacomantonio2010boolean},
\begin{equation}
P \leftarrow \neg C \wedge \neg E \wedge S; \quad S \leftarrow F \wedge \neg E; \quad F \leftarrow F \wedge S \wedge \neg E; \quad E \leftarrow \neg F \wedge \neg P \wedge \neg S \wedge C; \quad C \leftarrow \neg S \wedge \neg F.
\end{equation}
For instance, the propensity function $f_P$ for $k=n=1$, is
\begin{equation}
 g_p(R) = \left(\frac{ [S] }{1+ [C] + [E] + [S] + [C][E] + [E][S] + [C][S] + [C][S][E]}\right).
 \end{equation}

\paragraph{HSC model.} The Boolean rules associated with the HSC model are taken from \cite{krumsiek2011hierarchical},
 \begin{align}
& G1 \leftarrow (G1 \vee  G2 \vee Fli) \wedge \neg P; \quad 
G2 \leftarrow G2 \wedge \neg (G1 \wedge Fg) \wedge \neg P; \quad 
Fg \leftarrow G1; \quad 
E \leftarrow G1 \wedge \neg Fli; \\
& Fli \leftarrow G1 \wedge \neg E; \quad 
S \leftarrow G1 \wedge \neg P; \quad 
Ceb \leftarrow Ceb \wedge \neg( G1 \wedge Fg \wedge S ); \quad 
P \leftarrow (Ceb \vee  P ) \wedge \neg  ( G1 \vee G2 ); \\
& cJ \leftarrow (P \wedge \neg  G ); \quad 
Eg \leftarrow (P \wedge cJ) \wedge \neg G; \quad 
G \leftarrow (Ceb \wedge \neg Eg) .
\end{align}

\setlength{\tabcolsep}{0.5em} 
{\renewcommand{\arraystretch}{1.5}
\rev{
\begin{table}[hbt!]
  \centering
\begin{tabular}{ |c| c| c| } 
 \hline
 &Stochastic Differential Equation  &   $\mathbf{D(x)}:\mathbb{R}^d \rightarrow \mathbb{R}^{d\times d}$\\ 
  \hline
  CLE     &$ d\mathbf{x} = \left( m V \mathbf{g}(\mathbf{x},V) - \ell \mathbf{x} \right)dt +  \sqrt{ m V \mathbf{g}(\mathbf{x},V) + \ell \mathbf{x}} \, d\mathbf{W}$   & $\frac{1}{2}\text{diag}\left(m V g_1(\mathbf{x},V) - \ell x_1,\cdots,m V g_d(\mathbf{x},V) - \ell x_d\right)$ \\ 
\hline
  Multiplicative $\sqrt{x}$    &$ d\mathbf{x} = \left( m V \mathbf{g}(\mathbf{x},V) - \ell \mathbf{x} \right)dt +  \sqrt{ \mathbf{x} + \ell \mathbf{x}} \, d\mathbf{W}$ &  $\frac{1}{2}\text{diag}\left(x_1 - \ell x_1,\cdots,x_d - \ell x_d\right)$ \\ 
\hline
  Additive noise     &$ d\mathbf{x} = \left( m V \mathbf{g}(\mathbf{x},V) - \ell \mathbf{x} \right)dt + d\mathbf{W}$ &  $\frac{1}{2} \mathbf{I}_d$ \\ 
\hline
    Deterministic   &$ d\mathbf{x} = \left( m V \mathbf{g}(\mathbf{x},V) - \ell \mathbf{x} \right)dt$  & $0$ \\ 
  \hline
\end{tabular}
\vspace{0.1em}
\caption{\rev{Summary of the explicit form of the diffusion tensor $\mathbf{D(x)}$ used in the probability flow ODE under different noise models. The deterministic force remains fixed across all models and is given by \( \mathbf{f}(\mathbf{x}) = mV\mathbf{g}(\mathbf{x}, V) - \ell \mathbf{x} \).}} \label{PFI_terms}
\end{table}}

\section{Score estimation and validation}
\label{sec:SM_score}
\rev{We evaluate two strategies for training noise-conditional score networks (NCSNs): \textit{denoising score matching} (DSM) and \textit{sliced score matching} (SSM). Both approaches are effective for estimating the score function from a high-dimensional empirical distribution \cite{song2019generative}. Below, we outline the two methods and discuss their training objectives and computational performance.}

\noindent \rev{\paragraph*{Denoising Score Matching (DSM):} For the Gaussian noise model \( q_\sigma(\tilde{\mathbf{x}} \mid \mathbf{x}) = \mathcal{N}(\tilde{\mathbf{x}} \mid \mathbf{x}, \sigma^2 \mathbf{I}_d) \), the denoising score matching (DSM) objective aggregated over multiple noise levels is given by:
\begin{equation}
\mathcal{\ell}(\phi, t_k) = 
\frac{1}{2} \sum_{i=1}^{L} \sigma_i^2 \, \mathbb{E}_{\mathbf{x}(t_k)} \, \mathbb{E}_{\tilde{\mathbf{x}} \sim \mathcal{N}(\mathbf{x}(t_k), \sigma_i^2 I)} 
\left[ \left\| \mathbf{s}_\phi(\tilde{\mathbf{x}}, \sigma_i, t_k) + \frac{\tilde{\mathbf{x}} - \mathbf{x}(t_k)}{\sigma_i^2} \right\|_2^2 \right].\label{eq:DSM}
\end{equation}
where the score function \( \mathbf{s}_\phi(\mathbf{x}, \sigma, t): \mathbb{R}^{d+1} \times [0,T] \to \mathbb{R}^d \). For time-resolved cross-sectional data, we define the overall training objective by summing the losses across all time points:
\begin{equation}
\min_\phi \sum_{k=1}^K \lambda_k \, \mathcal{\ell}(\phi, t_k),
\end{equation}
where \( \lambda_k \) are adaptive weights and \( t \sim \mathcal{U}(0, T) \). In all experiments, the set of noise levels \( \{ \sigma_i \}_{i=1}^L \) is chosen to follow a geometric progression, with \( L = 10 \), \( \sigma_1 = 10 \), and \( \sigma_{10} = 0.01 \) \cite{song2019generative}.}\\

\noindent \paragraph*{Sliced Score Matching (SSM):}  Sliced score matching \cite{song2020sliced} estimates the score without corrupting the input data, instead projecting gradients onto random directions \( \mathbf{v} \sim \mathcal{N}(0, I) \). The objective is:
\begin{align}
\min_{\phi} \sum_{k=1}^K \lambda_k \, \mathbb{E}_{\mathbf{x}(t_k)} \mathbb{E}_{\mathbf{v} \sim p_\mathbf{v}} \left[
\frac{1}{2} \left( \mathbf{v}^\top \mathbf{s}_\phi(\mathbf{x}(t_k), t_k) \right)^2 + 
\mathbf{v}^\top \nabla_{\mathbf{x}} \left( \mathbf{v}^\top \mathbf{s}_\phi(\mathbf{x}(t_k), t_k) \right)
\right]. \label{eq:sliced_score}
\end{align}
For both DSM and SSM, we parameterize the score function \( \mathbf{s}_\phi(\mathbf{x}, t): \mathbb{R}^{d} \times [0,T] \to \mathbb{R}^d \) using a feedforward neural network. Adaptive weighting \( \lambda_k \) is tuned using the variance normalization strategy proposed in \cite{maddu2022inverse}. Optimization is performed using the Adam optimizer with an initial learning rate \( \eta = 10^{-3} \).  Hyperparameter details of the score network architecture are provided in Table~\ref{network_architecture}.

\paragraph*{Score validation:} After training a neural network to approximate the score function  \( \mathbf{s}_\theta(\mathbf{x}) \approx \nabla_{\mathbf{x}} \log p_{\text{data}}(\mathbf{x}) \), Langevin dynamics can be used to sample from the target distribution \( p_{\text{data}}(\mathbf{x}) \). Starting with a fixed step size \( \epsilon > 0 \) and an initial value \( \tilde{\mathbf{x}}_0 \sim \pi(\mathbf{x}) \), where \( \pi \) is a prior distribution, the Langevin method iteratively updates the samples using the equation:
\begin{equation}
    \tilde{\mathbf{x}}_t = \tilde{\mathbf{x}}_{t-1} + \frac{\epsilon}{2} \nabla_{\mathbf{x}} \log p(\tilde{\mathbf{x}}_{t-1}) + \sqrt{\epsilon} \mathbf{z}_t,
\end{equation}
where \( \mathbf{z}_t \sim \mathcal{N}(0, \mathbf{I}_d) \). Under certain conditions, as \( \epsilon \to 0 \) and \( T \to \infty \), the distribution of \( \tilde{\mathbf{x}}_T \) converges to \( p(\mathbf{x}) \), resulting in exact samples from \( p(\mathbf{x}) \) \cite{welling2011bayesian}. For finite \( \epsilon \) and \( T \), a Metropolis-Hastings update is often used to correct the approximation, though this correction is typically negligible for small \( \epsilon \) and large \( T \) \cite{song2019generative}. This sampling approach relies only on the score function \( \nabla_{\mathbf{x}} \log p(\mathbf{x}) \). We validate the score model by running Langevin dynamics to generate samples and evaluate how accurately they match the cross-sectional data.

\setlength{\tabcolsep}{0.5em} 
{\renewcommand{\arraystretch}{1.2}
\begin{table}[htbp]
  \centering
\begin{tabular}{ |c|c|c|c|c|c| } 
 \hline
  & Dimension (d)  &  $\#$ hidden layers & $\#$ nodes & $\#$ snapshots (K)/ $\Delta t$ & $\#$ samples ($n$)\\ 
  \hline
Ornstein-Ulhenbeck &10 & 3 & 50 & 10/0.05 & 8000\\ 
 Cyclic SRN &30 & 4 & 100 & 10/0.04 & 10000\\ 
 mCAD   &5 & 4 & 50 & 10/0.04 & 6000\\ 
 simulated HSC   &11 & 6 & 100 & 8/0.04 & 5000\\ 
  \textit{ex vivo} HSC   & 10,15,20,30 & 4 & 100 & 6/2 days & $6000-10000$\\ 
  \hline
\end{tabular}
\vspace{0.1em}
\caption{Hyper-parameters of the different network architectures used for training the score network $\mathbf{s}_\phi: \mathbb{R}^{d}\times [0,T] \rightarrow \mathbb{R}^d$. } \label{network_architecture}
\end{table}

\begin{figure}[hbt!]
\centering
  \includegraphics[width=16cm]{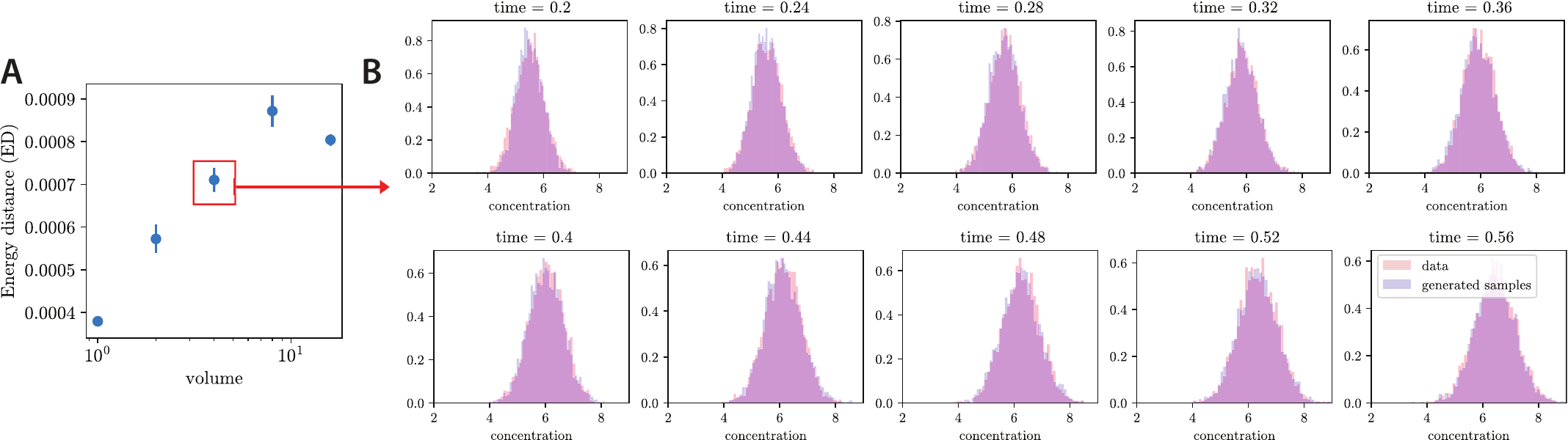}
  \caption{\textbf{Score validation for linear cyclic network $(d=30)$:} {\bf{A.}} Energy distance between the generated samples from Langevin dynamics and the marginal data for different system size, $(V)$. {\bf{B.}} In panel {\bf{A.}}, we show the histogram of concentration values over time for both data (red) and generated samples (blue) for the system size $V=4$. The times range from $0.2$ to $0.56$, showing the evolution of the distribution of concentration values.}\label{fig:score_linear}
\end{figure}

\begin{figure}[hbt!]
\centering
  \includegraphics[width=16cm]{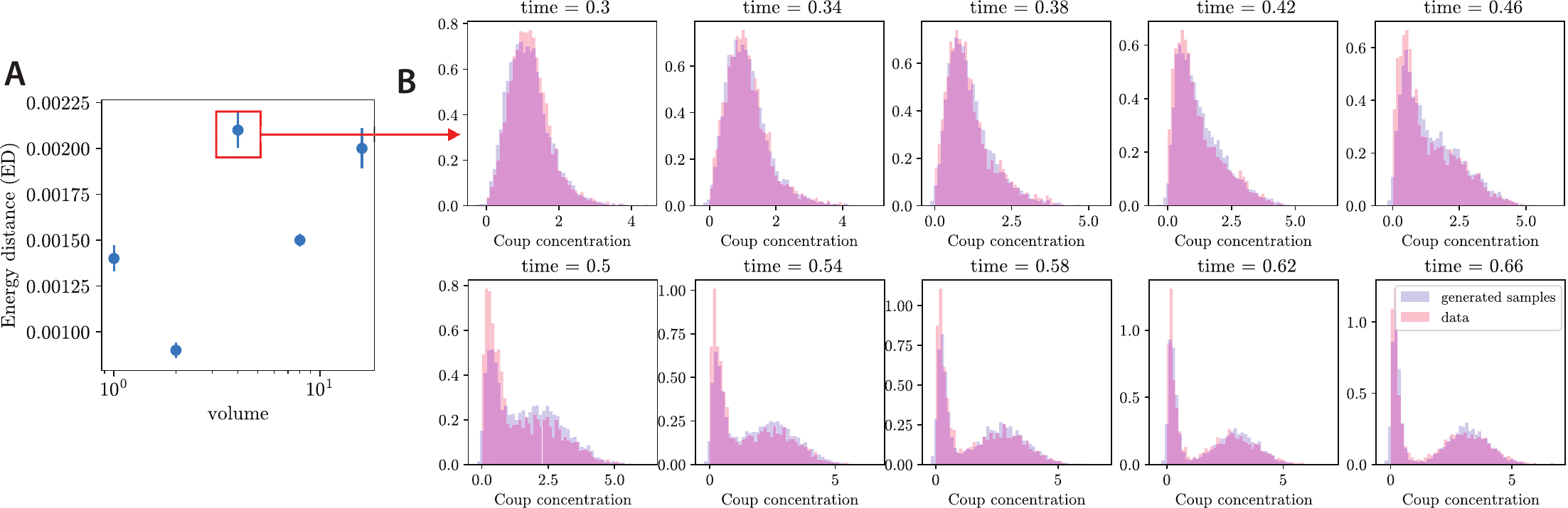}
  \caption{\textbf{Score validation for mCAD network $(d=5)$:} {\bf{A.}} Energy distance between the generated samples from Langevin dynamics and the marginal data for different system size, $(V)$. {\bf{B.}} 
  In panel {\bf{A.}}, we show the histogram of mRNA concentration values for the \textit{Coup} gene, comparing data (red) and generated samples (blue) for the system size $V=4$. }\label{fig:score_mCAD}
\end{figure}

\begin{figure}[hbt!]
\centering
  \includegraphics[width=16cm]{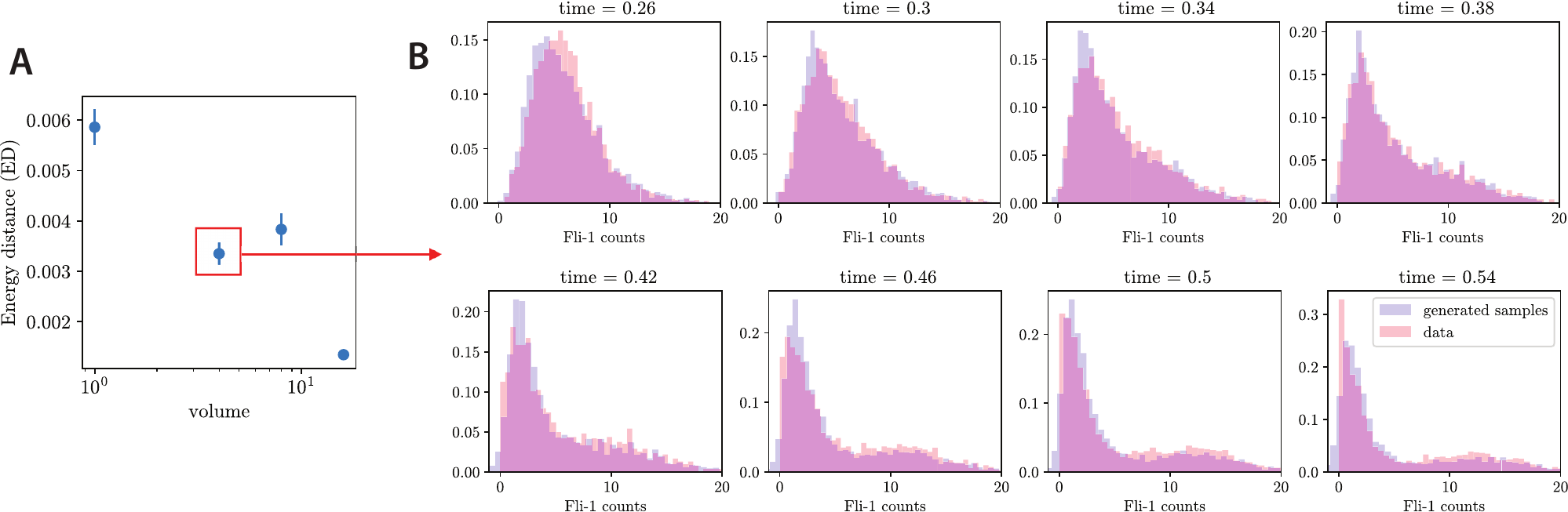}
  \caption{\textbf{Score validation for HSC network $(d=11)$:} {\bf{A.}} Energy distance between the generated samples from Langevin dynamics and the marginal data for different system size, $(V)$. {\bf{B.}} {\bf{B.}} 
   Histograms showing the time evolution of mRNA concentration values for the \textit{Fli-1} gene over evolving, comparing data (red) and generated samples (blue), are shown for system size $V=4$, as highlighted in panel {\bf{A.}}}\label{fig:score_HSC}
\end{figure}


\newpage

\section{Simulation-free training based on Conditional Flow Matching (CFM) with quadratic spline interpolation}
\label{sec:algorithm}

\rev{Flow Matching (FM) is a simulation-free framework for generative modeling based on learning continuous-time vector fields that transform samples from a source distribution \( p_0 \) to a target distribution \( p_T \) \cite{lipman2022flow}. The transformation is modeled via a time-dependent velocity field \( \mathbf{u}_t(\mathbf{x}) \) such that the solution of the ODE \( \dot{\mathbf{x}} = \mathbf{u}_t(\mathbf{x}) \), initialized at \( \mathbf{x}(0) \sim p_{0} \), results in \( \mathbf{x}(T) \sim p_{T} \). The learning objective is to minimize the squared error between a learned field \( \mathbf{v}_t(\mathbf{x}; \theta) \) and the true velocity field:
\[
\mathcal{L}_{\mathrm{FM}}(\theta) = \mathbb{E}_{t, \mathbf{x} \sim p_t(\mathbf{x})} \left\| \mathbf{v}_t(\mathbf{x}; \theta) - \mathbf{u}_t(\mathbf{x}) \right\|_2^2.
\]
with $t$ sampled from the uniform distribution $t \sim U([0, T])$. In practice, the intermediate distributions \( p_t(\mathbf{x}) \) and the true velocity field \( \mathbf{u}_t(\mathbf{x}) \) are unknown. To address this, Conditional Flow Matching (CFM) introduces auxiliary variables \( \mathbf{z} \), and defines conditional paths \( p_t(\mathbf{x} \mid \mathbf{z}) \), leading to the tractable objective:
\begin{equation}\label{eq:CFM}
    \mathcal{L}_{\mathrm{CFM}}(\theta) = \mathbb{E}_{t, \mathbf{z} \sim q(\mathbf{z}), \mathbf{x} \sim p_t(\mathbf{x} \mid \mathbf{z})} \left\| \mathbf{v}_t(\mathbf{x}; \theta) - \mathbf{u}_t(\mathbf{x} \mid \mathbf{z}) \right\|_2^2.
\end{equation}
\noindent The conditioning variable $\mathbf{z}$ and conditional probability paths $p_t(\mathbf{x} | \mathbf{z})$ are chosen such that the marginals match the boundary distributions $p_0$ and $p_T$. A very common form for conditional probability paths is given by:
\begin{equation}
p_t(\mathbf{x} \mid \mathbf{z}) = \mathcal{N}(\mathbf{x} \mid \bm{\mu}_t(\mathbf{z}), \sigma_t(\mathbf{z})^2 I),
\end{equation}
where $\bm{\mu}: [0, T] \times \mathcal{Z} \to \mathcal{X}$ and $\sigma: [0,T] \times \mathcal{Z} \to \mathbb{R}^+$ define the time-dependent mean and standard deviation of a Gaussian distribution. The above choice of Gaussian probability path leads to a unique inducing vector field (Theorem 3 in \cite{lipman2022flow}), given by:
\begin{equation}\label{eq:conditional_vel}
\mathbf{u}_t(\mathbf{x} \mid \mathbf{z}) = \frac{\sigma_t'(\mathbf{z})}{\sigma_t(\mathbf{z})} \left( \mathbf{x}_t - \bm{\mu}_t(\mathbf{z}) \right) + \bm{\mu}_t'(\mathbf{z}).
\end{equation}
The CFM can also be used to bridge multiple marginals if the parameterizations of $\bm{\mu}_t(\mathbf{z})$ and $\sigma_t(\mathbf{z})$ are designed to satisfy the marginal boundary conditions $\int p_{t_k}(\mathbf{x} \mid \mathbf{z}) q(\mathbf{z}) d\mathbf{z} \approx p_{t_k}(\mathbf{x}), \: \forall k \in \{0,...,K \}$. We use quadratic spline interpolation to represent the conditional paths $\bm{\mu}_t(\mathbf{z})$ and set $q(\mathbf{z}) = p_{t_{k-1}}(\mathbf{x}_{k-1})p_{t_{k}}(\mathbf{x}_k) p_{t_{k+1}}(\mathbf{x}_{k+1})$ to be independent sampling from each distribution. However, this approach may become unstable in the presence of noise. A potential improvement would be to use sampling strategies based on optimal transport \cite{tong2023improving}, though this lies beyond the scope of the present work. To define the conditional mean path \( \bm{\mu}_t(\mathbf{z}) \) analytically, we parameterize it using a piecewise quadratic spline over three consecutive time points \( t_{k-1} \leq t_k \leq t_{k+1}, k \in \{1,...,K-1\} \), with corresponding positions \( \mathbf{x}_{k-1} = \bm{\mu}_{t_{k-1}}(\mathbf{z}) \), \( \mathbf{x}_{k} = \bm{\mu}_{t_{k}}(\mathbf{z}) \), and \( \mathbf{x}_{k+1} = \bm{\mu}_{t_{k+1}}(\mathbf{z}) \). 
The conditional mean path is then defined as:
\[
\bm{\mu}_t(\mathbf{z}) =
\begin{cases}
a_1 t^2 + b_1 t + c_1, & t \in [t_{k-1}, t_k) \\
a_2 t^2 + b_2 t + c_2, & t \in [t_{k}, t_{k+1}]
\end{cases}
\]
The coefficients of the spline are computed to ensure position continuity at \( t_{k-1}, t_k, t_{k+1} \), and velocity continuity at \( t_k \). To close the system, we impose a natural spline boundary condition by setting the initial velocity to zero: \( \bm{\mu}_t'(\mathbf{z})\big|_{t = t_{k-1}} = 0 \). For constant variance $\sigma^2$, the conditional velocity is the time derivative of the conditional path $\bm{\mu}_t$ as per Eq.~(\ref{eq:conditional_vel}), i.e.,
\[
\mathbf{u}_t(\mathbf{x} \mid \mathbf{z}) =
\begin{cases}
2 a_1 t + b_1, & t \in [t_{k-1}, t_{k}) \\
2 a_2 t + b_2, & t \in [t_{k}, t_{k+1}]
\end{cases}
\]
By construction, this velocity field is continuous and satisfies \( \mathbf{u}_{t_{k-1}}(\mathbf{x} \mid \mathbf{z}) = 0 \). These resulting \( (\mathbf{x}, t) \) and \( \mathbf{u}_t(\mathbf{x} \mid \mathbf{z}) \) pairs are used to optimize the parameters $\theta$ of the force field by minimizing the CFM objective as described in the algorithm~2.}

\begin{figure}[hbt!]
\centering
  \includegraphics[width=16cm]{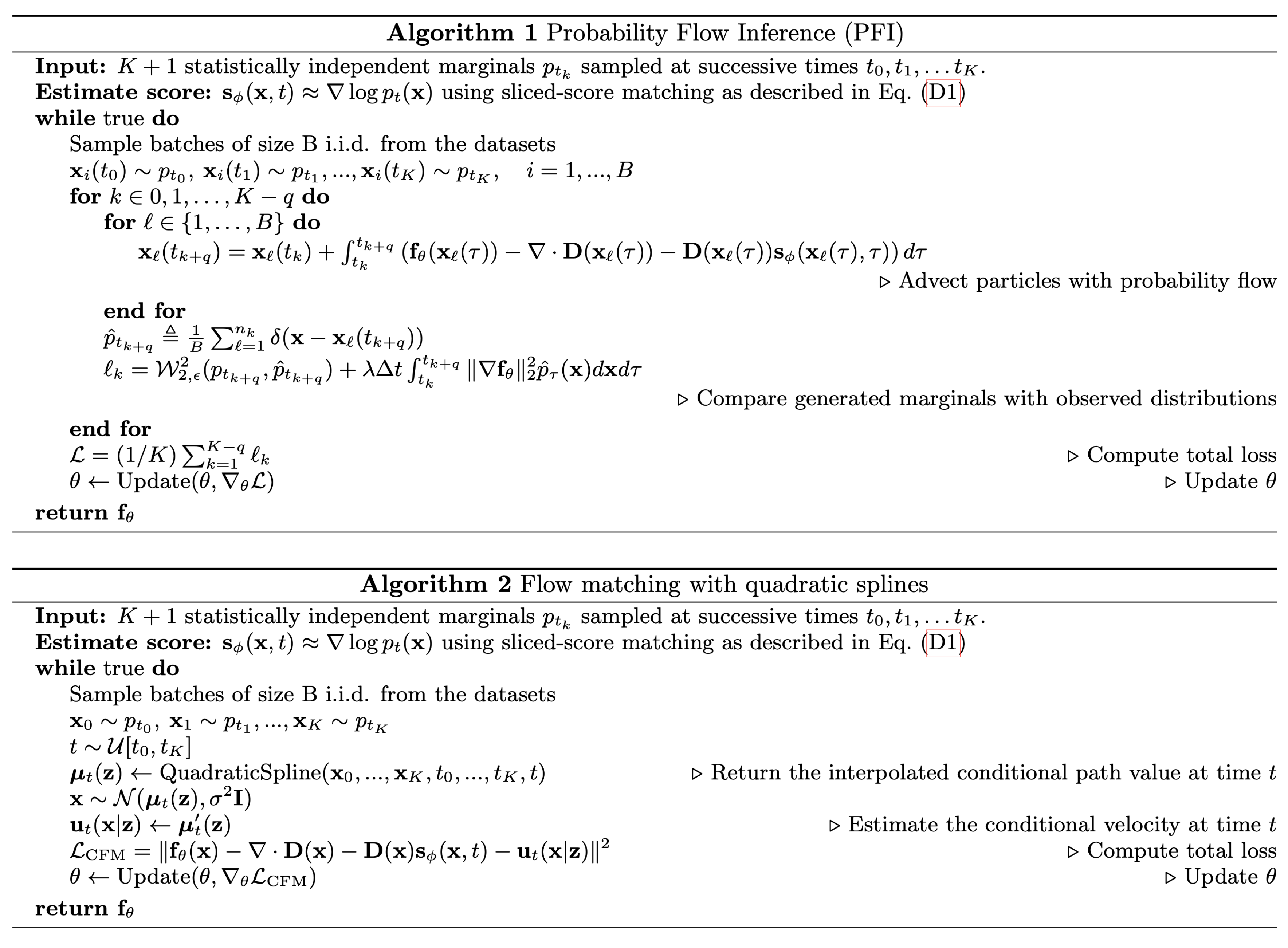}
\end{figure}

\newpage
\section{Computational aspects of Probability Flow Inference }

\rev{\noindent \paragraph*{Score estimation:} In general, the memory requirement and time per gradient update for Denoising Score Matching (DSM) scale as $\mathcal{O}(B \cdot d)$ and $\mathcal{O}(B)$, respectively, where $B$ denotes the batch size and $d$ is the input dimension. In contrast, for Sliced Score Marching (SSM), both the memory and time per gradient update scale as $\mathcal{O}(B \cdot d^2)$, assuming the number of slicing directions is proportional to the input dimension $d$. This is partially illustrated on the linear cyclic network problem to infer the kinetic rate parameters. In terms of time taken per gradient update, DSM scales more favorably with both batch size (Fig.~\ref{fig:scaling}B) and dimensionality (Fig.~\ref{fig:scaling}D), incurring only a modest memory overhead compared to SSM (Fig.~\ref{fig:scaling}A,C). In terms of accuracy, DSM and SSM are both viable, and the choice may depend on dataset structure and downstream inference requirements (see Fig.~\ref{fig:scaling}E). Importantly, DSM remains robust in high-noise regimes, especially when many gene counts are close to zero, as is often the case in biological datasets. In these settings, DSM provides improved numerical stability and computational efficiency due to its simpler optimization objective and faster convergence. Additionally, by injecting large Gaussian noise, DSM populates low-density regions of the data distribution, providing a richer training signal and improving score estimation in areas that would otherwise be poorly represented \cite{song2019generative}. Furthermore, both the SSM and DSM objectives can be evaluated in a batch-wise manner, allowing for straightforward integration with time-resolved marginal data, even when the number of samples varies across time points.}

\rev{\noindent \paragraph*{Force inference:} In general, PFI exhibits quadratic scaling with batch size ($B$) (Fig.~\ref{fig:scaling}F,G) and potentially with input dimension ($d$) (Fig.~\ref{fig:scaling}H,I), primarily due to the use of Sinkhorn distance and the need to compute the divergence of the diffusion tensor, \( \nabla \cdot \mathbf{D}(\mathbf{x}) \), which in turn requires evaluating the Jacobian of the force field. In contrast, Flow Matching (FM) with quadratic splines scales linearly with batch size, making it significantly more efficient in large sample settings. While FM avoids pairwise distance computations and explicit integration, it still requires evaluating \( \nabla \cdot \mathbf{D}(\mathbf{x}) \), which involves computing the Jacobian of the force field and can lead to quadratic scaling with input dimension. However, in both PFI and FM, this cost can be substantially reduced in practice by using techniques such as Hutchinson’s trace estimator \cite{grathwohl2018ffjord}, which approximates the divergence using random projections and scales linearly with input dimension. Overall, improving the scalability and accuracy of flow matching, both in terms of divergence estimation and conditional velocity construction, remains an important direction for future work.}

\begin{figure}[hbt!]
\centering
  \includegraphics[width=16cm]{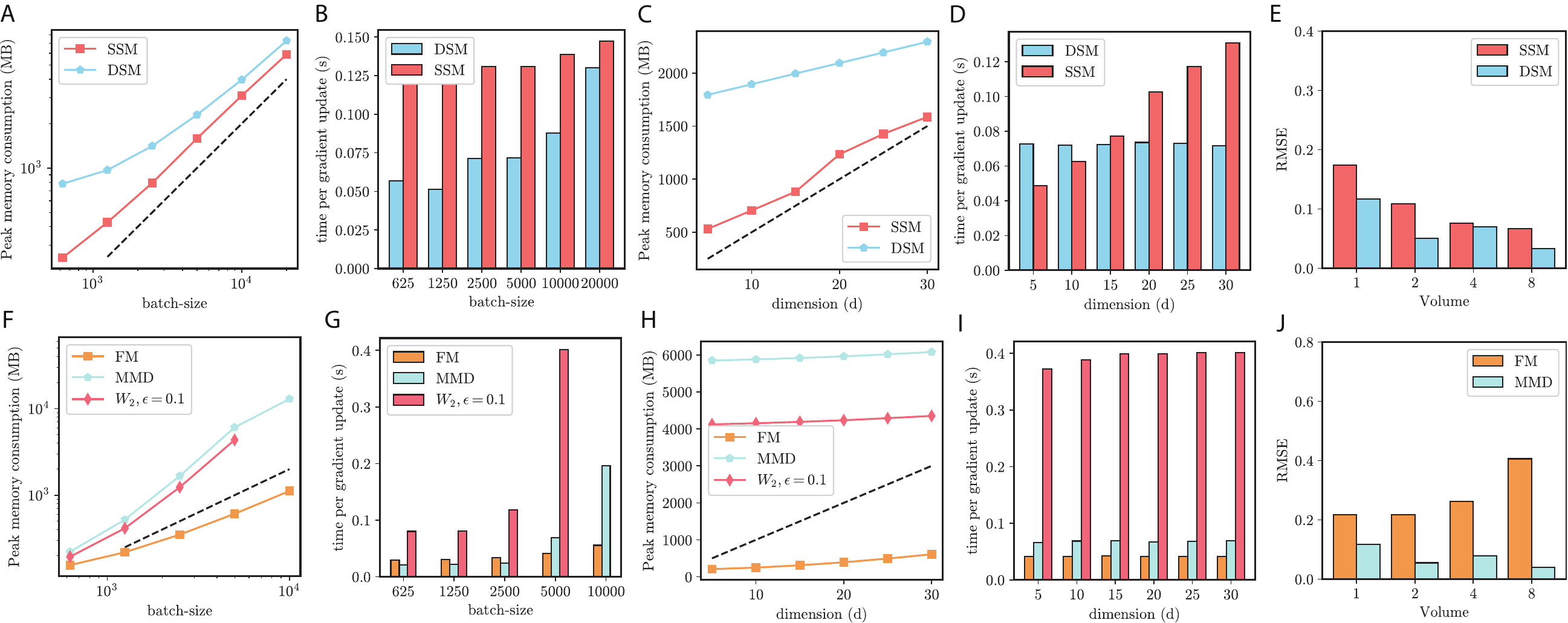}
  \caption{\rev{\textbf{Computational aspects.} We demonstrate the computational performance on the linear cyclic network example to infer the kinetic rates. \textbf{Score estimation step: }Peak memory usage and time per gradient update as a function of batch size (A, B) for \( d = 30 \), and as a function of dimension (C, D) with a fixed batch size of 5000. Comparisons are shown for sliced score matching (SSM) and denoising score matching (DSM) methods. (E) Accuracy of the inferred kinetic rate parameters using score estimates from DSM and SSM.
  \textbf{Force inference step:} Peak memory usage and time per gradient update as a function of batch size (F, G) for $d=30$, and as a function of dimension (H, I) at a fixed batch size of 5000. Comparisons include simulation-based methods using different distance metrics (\( \mathcal{W}_{2,\epsilon} \), MMD) and simulation-free flow-matching (FM) strategies. (J) Accuracy of the inferred parameters across different volumes for both simulation-based and simulation-free approaches. All memory and timing measurements were performed on an NVIDIA A100-SXM4-80GB GPU. The dashed black line denotes linear scaling.}}\label{fig:scaling}
\end{figure}

\newpage 

\section{Additional results}

\begin{figure}[hbt!]
 \centering
  \includegraphics[width=16cm]{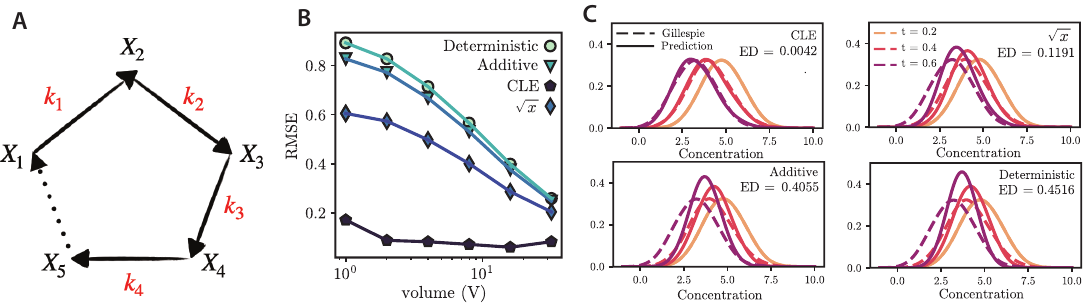}
  \caption{\textbf{Parameter estimation in linear cyclic stochastic reaction network:} {\bf{A.}} Schematic of the linear cyclic network. {\bf{B.}} The RMSE $ \Vert \hat{\mathbf{k}} - \mathbf{k}  \Vert_2^2/\Vert \mathbf{k}  \Vert_2^2 $ of the inferred rate constants shown for different compartment volumes $V$. {\bf{C.}} Comparison of empirical marginals generated from the inferred diffusion process under various noise forms (inset) with marginals generated from Gillespie's simulation with $V=4$.}\label{fig:SRN}
\end{figure}

\begin{figure}[hbt!]
  \includegraphics[width=\textwidth]{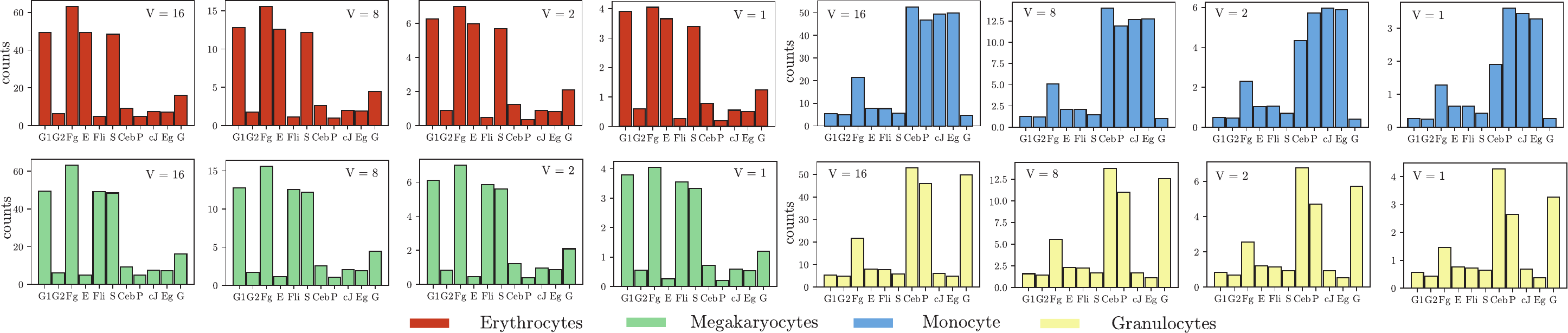}
  \caption{\textbf{HSC differentiation.} Average molecular counts as a function of the reaction volume (V) in each cell type represented by different colors. The y-axis represents the average molecular counts, highlighting the effect of varying reaction volumes on the observed counts. }\label{fig:counts}
\end{figure}

\begin{table}[hbt!]
  \centering
\begin{tabular}{ |c|c|c|c|c| } 
 \hline
 Algorithm  & $V=16$  &  $V=4$ & $V=2$ & $V=1$ \\ 
  \hline
TrajectoryNet & $0.037 \pm 0.014$ & $0.009 \pm 0.003$ & $\textbf{0.013} \pm 0.002$ & $ 0.037 \pm 0.008$ \\ 
 PFI-CLE & $  \textbf{0.004} \pm 0.001 $  & $\textbf{0.005} \pm 0.002$ & $ 0.021 \pm 0.012$ & $ 0.052 \pm 0.028$ \\ 
 PRESCIENT  & $0.009 \pm 0.004$ & $0.019 \pm 0.01$ & $0.019 \pm 0.011$ & $ \textbf{0.03} \pm 0.012$ \\ 
  \hline
\end{tabular}
\vspace{1em}
\caption{\rev{Comparison between prediction accuracy of different methods on the simulated HSC dataset for varying system size, $V$. We report the energy distance between the predicted and ground truth distributions at the left-out time point.}} \label{leave_out_tp}
\end{table}

\begin{table}[hbt!] 
  \centering
\begin{tabular}{|l|p{12cm}|}
\hline
\textbf{Gene Set} & \textbf{Genes} \\
\hline
Set 1 $(d=10)$ & fli1, klf1, gata1, gata2, gfi1b, runx1, scl, cjun, pu1, fog1 \\
\hline
Set 2 $(d=14)$ & fli1, klf1, gata1, gata2, gfi1b, runx1, scl, cjun, pu1, fog1, lmo2, etv6, erg, mef2c \\
\hline
Set 3$(d=20)$  & fli1, klf1, gata1, gata2, gfi1b, runx1, scl, cjun, pu1, fog1, lmo2, etv6, erg, mef2c, cebpa, nfe2, myc, stat3, nanog, meis1 \\
\hline
Set 4 $(d=30)$ & fli1, klf1, gata1, gata2, gfi1b, runx1, scl, cjun, pu1, fog1, lmo2, etv6, erg, mef2c, cebpa, nfe2, myc, stat3, nanog, meis1, foxo3, hoxa9, xbp1, tcf4, ets2, ctcf, mllt10, nfib, myb, mybl2 \\
\hline
\end{tabular}
\caption{\rev{Transcription factor gene sets used to analyze \textit{ex vivo} hematopoeisis dataset.}} \label{table:geneset}
\end{table}

\begin{figure}[hbt!]
\centering
  \includegraphics[width=14cm]{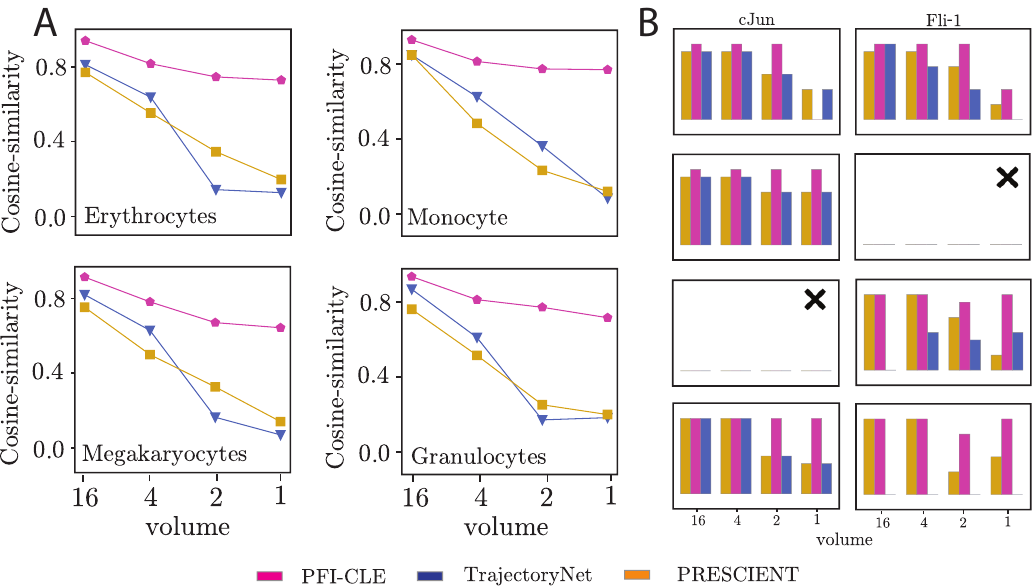}
  \caption{\rev{ \textbf{Inferring HSC differentiation dynamics on simulated data.}  {\bf{A.}} Cosine similarity between inferred probability flow lines and the true CLE flow lines for four cell types (Erythrocytes, Megakaryocytes, Monocytes, and Granulocytes) under varying noise conditions (represented by volume) shown for three different approaches (PRESCIENT, TrajectoryNet, PFI).
  {\bf{B.}} Probability of observing Erythrocytes, Megakaryocytes, Monocytes, and Granulocytes at steady state. Each panel displays the estimated probability of recovering the different cell types (rows) under specific perturbation conditions (columns). The averages are computed across varying initial conditions. The panels also show how these probabilities change with system size, $V$. Panels marked with a cross indicate the absence of the respective cell type under the corresponding perturbations. The results are color-coded by the approach used: TrajectoryNet (blue), PRESCIENT (gold), and Inferred Chemical Langevin model (purple).}}\label{fig:HSC_add}
\end{figure}

\begin{figure}[hbt!] 
\centering
  \includegraphics[width=\textwidth]{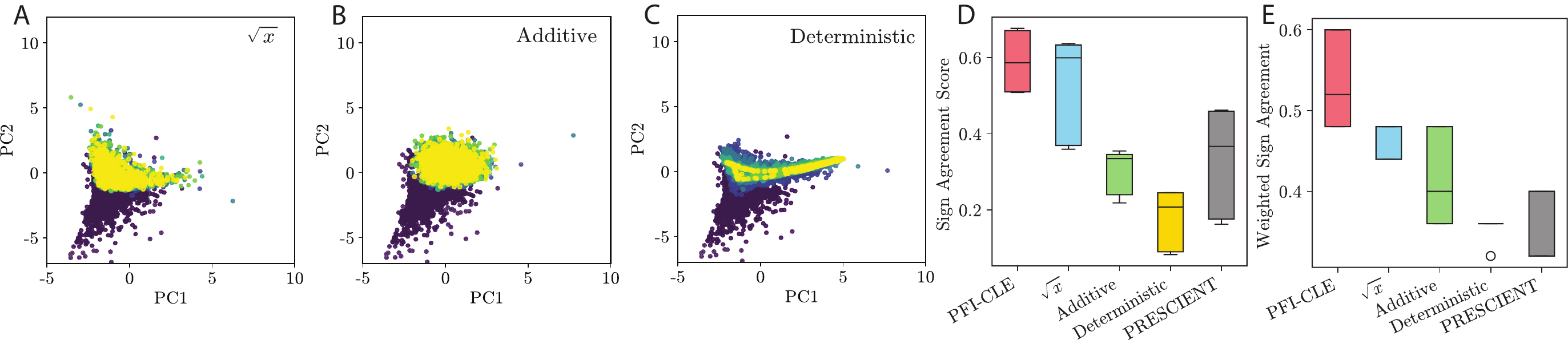}
  \caption{\rev{ \textbf{Inferring HSC differentiation dynamics on \textit{ex vivo} human hematapoeisis data.}  (\textbf{A,B,C}). Predicted differentiation trajectories following induction on day 0, using force fields inferred under different noise priors (shown in inset). (\textbf{D}) Sign Agreement Score, defined as $ = \sum_{ij} \mathbf{1} \left( \text{sign}(\Delta_{ij}) = M_{ij}\right)/d^2 $ and (\textbf{E}) Weighted Sign Agreement $ = \sum_{ij} \vert \Delta_{ij} \vert \mathbf{1} \left( \text{sign}(\Delta_{ij}) = M_{ij}\right)/\sum_{ij}  \vert \Delta_{ij} \vert $ quantify the agreement between the inferred response matrix $\Delta$ and the known regulatory matrix $M$, across different noise models and the PRESCIENT method.}}\label{fig:HSPC_different_noise}
\end{figure}

\newpage



\end{document}